\documentclass[sigconf,authorversion=true]{acmart}

\usepackage[linesnumbered, vlined, ruled]{algorithm2e}
\SetArgSty{textup}

\def\vector#1{\mbox{\boldmath $#1$}}

\usepackage{booktabs} 
\usepackage{graphicx}
\usepackage{rotating}
\usepackage{tabularx}
\usepackage{xspace}  
\usepackage{xstring} 
\usepackage{wasysym} 
\usepackage{MnSymbol} 
\usepackage{float}
\usepackage{ifthen}
\usepackage{xcolor}

\definecolor{Gray}{gray}{0.6}
\definecolor{NavyBlue}{rgb}{0.0, 0.0, 0.5}
\definecolor{Magenta}{rgb}{1.0, 0.0, 1.0}
\definecolor{Black}{rgb}{0.0, 0.0, 0.0}

\newcommand{\ERT}{\ensuremath{\mathrm{ERT}}}

\newcommand{\Df}{\ensuremath{\Delta f}}

\newcommand{\fopt}{\ensuremath{f_\mathrm{opt}}}
\newcommand{\ftarget}{\ensuremath{f_\mathrm{t}}}

\usepackage{xcolor}
\usepackage{colortbl}

\setcopyright{acmcopyright}

\copyrightyear{2022}
\acmYear{2022}
\setcopyright{acmcopyright}\acmConference[GECCO '22 Companion]{Genetic and Evolutionary Computation Conference Companion}{July 9--13, 2022}{Boston, MA, USA}
\acmBooktitle{Genetic and Evolutionary Computation Conference Companion (GECCO '22 Companion), July 9--13, 2022, Boston, MA, USA}
\acmPrice{15.00}
\acmDOI{10.1145/3520304.3533951}
\acmISBN{978-1-4503-9268-6/22/07}

\setcopyright{acmcopyright}

\copyrightyear{2022}
\acmYear{2022}
\setcopyright{acmcopyright}\acmConference[GECCO '22 Companion]{Genetic and Evolutionary Computation Conference Companion}{July 9--13, 2022}{Boston, MA, USA}
\acmBooktitle{Genetic and Evolutionary Computation Conference Companion (GECCO '22 Companion), July 9--13, 2022, Boston, MA, USA}
\acmPrice{15.00}
\acmDOI{10.1145/3520304.3533951}
\acmISBN{978-1-4503-9268-6/22/07}

% ...
\newcommand{\bbobdatapath}{ppdata/} 

\providecommand{\bbobecdfcaptionsinglefunctionssingledim}[1]{
Empirical cumulative distribution of simulated (bootstrapped)
             runtimes, measured in number of $f$-evaluations,
             divided by dimension (FEvals/DIM) for the $51$ 
             targets $10^{[-8..2]}$ in dimension #1.
}
\providecommand{\cocoversion}{{\scriptsize\sffamily{}\color{Gray}Data produced with COCO v2.6}}
\providecommand{\numofalgs}{6}
\providecommand{\bbobECDFslegend}[1]{
Bootstrapped empirical cumulative distribution of the number of $f$-evaluations divided by dimension (FEvals/DIM) for $51$ targets with target precision in $10^{[-8..2]}$ for all functions and subgroups in #1-D. 
}
\providecommand{\bbobppfigslegend}[1]{
Expected running time (\ERT\ in number of $f$-evaluations
                    as $\log_{10}$ value), divided by dimension for target function value $10^{-8}$
                    versus dimension. Slanted grid lines indicate quadratic scaling with the dimension. Different symbols correspond to different algorithms given in the legend of #1. Light symbols give the maximum number of evaluations from the longest trial divided by dimension. Black stars indicate a statistically better result compared to all other algorithms with $p<0.01$ and Bonferroni correction number of dimensions (six).  
Legend: 
{\color{NavyBlue}$\circ$}: \algorithmA
, {\color{Magenta}$\diamondsuit$}: \algorithmB
, {\color{Orange}$\star$}: \algorithmC
, {\color{CornflowerBlue}$\triangledown$}: \algorithmD
, {\color{red}$\varhexagon$}: \algorithmE
, {\color{YellowGreen}$\triangle$}: \algorithmF
}
\definecolor{NavyBlue}{HTML}{000080}
\definecolor{Magenta}{HTML}{FF00FF}
\definecolor{Orange}{HTML}{FFA500}
\definecolor{CornflowerBlue}{HTML}{6495ED}
\definecolor{YellowGreen}{HTML}{9ACD32}
\definecolor{Gray}{HTML}{BEBEBE}
\definecolor{Yellow}{HTML}{FFFF00}
\definecolor{GreenYellow}{HTML}{ADFF2F}
\definecolor{ForestGreen}{HTML}{228B22}
\definecolor{Lavender}{HTML}{FFC0CB}
\definecolor{SkyBlue}{HTML}{87CEEB}
\definecolor{NavyBlue}{HTML}{000080}
\definecolor{Goldenrod}{HTML}{DDF700}
\definecolor{VioletRed}{HTML}{D02090}
\definecolor{CornflowerBlue}{HTML}{6495ED}
\definecolor{LimeGreen}{HTML}{32CD32}

\providecommand{\ntables}{7}
\providecommand{\ntables}{7}
\providecommand{\ntables}{7}
\providecommand{\algFtables}{\StrLeft{LBFGS}{\ntables}}
\providecommand{\algEtables}{\StrLeft{MTSLS1-9}{\ntables}}
\providecommand{\algDtables}{\StrLeft{MTSLS1-5}{\ntables}}
\providecommand{\algCtables}{\StrLeft{HJ-9}{\ntables}}
\providecommand{\algBtables}{\StrLeft{HJ-5}{\ntables}}
\providecommand{\algAtables}{\StrLeft{BSrr}{\ntables}}
\providecommand{\ntables}{7}
\providecommand{\ntables}{7}
\providecommand{\pptablesfooter}{
\end{tabularx}
}
\providecommand{\pptablesheader}{
\begin{tabularx}{1.0\textwidth}{@{}c@{}|*{7}{@{}r@{}X@{}}|@{}r@{}@{}l@{}}
$\Delta f_\mathrm{opt}$ & \multicolumn{2}{@{\,}l@{\,}}{1e1} & \multicolumn{2}{@{\,}l@{\,}}{1e0} & \multicolumn{2}{@{\,}l@{\,}}{1e-1} & \multicolumn{2}{@{\,}l@{\,}}{1e-2} & \multicolumn{2}{@{\,}l@{\,}}{1e-3} & \multicolumn{2}{@{\,}l@{\,}}{1e-5} & \multicolumn{2}{@{\,}l@{\,}}{1e-7} & \multicolumn{2}{|@{}l@{}}{\#succ}\\\hline
}
\providecommand{\ntables}{7}
\providecommand{\bbobpptablesmanylegend}[1]{%
        Expected runtime (\ERT) to reach given targets, measured
        in number of $f$-evaluations, in #1. For each function, the \ERT\ 
        and, in braces as dispersion measure, the half difference between 10 and 
        90\%-tile of (bootstrapped) runtimes is shown for the different
        target \Df-values as shown in the top row. 
        \#succ is the number of trials that reached the last target
        $\fopt + 10^{-8}$.
        The median number of conducted function evaluations is additionally given in 
        \textit{italics}, if the target in the last column was never reached.
        Entries, succeeded by a star, are statistically significantly better (according to
        the rank-sum test) when compared to all other algorithms of the table, with
        $p = 0.05$ or $p = 10^{-k}$ when the number $k$ following the star is larger
        than 1, with Bonferroni correction by the number of functions (24). Best results are printed in bold.
        }
\providecommand{\algsfolder}{BSrr_HJ-5_HJ-9_MTSLS_MTSLS_LBFGS/}
\providecommand{\algorithmA}{BSrr}
\providecommand{\algorithmB}{HJ-5}
\providecommand{\algorithmC}{HJ-9}
\providecommand{\algorithmD}{LBFGS}
\providecommand{\algorithmE}{MTSLS1-5}
\providecommand{\algorithmF}{MTSLS1-9}

\ifthenelse{\equal{\numofalgs}{1}}{
   \graphicspath{{\bbobdatapath\algfolder}}}{
	 \graphicspath{{\bbobdatapath\algsfolder}}
}

\ifthenelse{\isundefined{\algorithmA}}{\newcommand{\algorithmA}{\algname}}{}

\begin{document}

\title{Benchmarking the Hooke-Jeeves Method, MTS-LS1, and BSrr on the Large-scale BBOB Function Set}
\renewcommand{\shorttitle}{Benchmarking the Hooke-Jeeves Method, MTS-LS1, and BSrr}

\author{Ryoji Tanabe}
\affiliation{%
  \institution{Yokohama National University \& RIKEN AIP}
  \city{Yokohama}
  \state{Kanagawa}
  \country{Japan}}
  \email{rt.ryoji.tanabe@gmail.com}

\begin{abstract}

  
This paper investigates the performance of three black-box optimizers exploiting separability on the 24 large-scale BBOB functions, including the Hooke-Jeeves method, MTS-LS1, and BSrr. 
Although BSrr was not specially designed for large-scale optimization, the results show that BSrr has a state-of-the-art performance on the five separable large-scale BBOB functions.
The results show that the asymmetry significantly influences the performance of MTS-LS1.
The results also show that the Hooke-Jeeves method performs better than MTS-LS1 on unimodal separable BBOB functions.

\end{abstract}

%
%




\maketitle

\section{Introduction}


The Hooke-Jeeves pattern search method \cite{HookeJ61} proposed in 1961 is one of the most classical black-box optimizers.
Throughout this paper, HJ denotes the Hooke-Jeeves method.
HJ generates a new search point $\vector{x}^{\mathrm{new}}$ by perturbing only one variable in the current one $\vector{x}$.
This operation is called a pattern move.
After all variables in $\vector{x}$ have been perturbed, the so-called exploratory move is performed.
If the pattern move was successful for at least one variable, HJ generates a new search point $\vector{x}^{\mathrm{new}}$ by taking the difference from the previous one $\vector{x}^{\mathrm{prev}}$ to the current one $\vector{x}$, i.e., $\vector{x}^{\mathrm{new}} := \vector{x} + (\vector{x} - \vector{x}^{\mathrm{prev}})$.


Multiple trajectory search (MTS) \cite{TsengC08} was designed for large-scale black-box optimization.
MTS was the winner of the IEEE CEC2008 competition on large-scale global optimization (LSGO).
MTS adaptively uses three local search methods: MTS-LS1, MTS-LS2, and MTS-LS3.
In this paper, we focus only on MTS-LS1 rather than MTS.
Some hybrid optimizers use MTS-LS1 as a local search method, including iCMAES-ILS \cite{LiaoS13}, MOS \cite{LaTorreMP11,LaTorreMP13}, and SHADE-ILS \cite{MolinaLH18}.
Here, they were the winners of the IEEE CEC2013, CEC2013 LSGO, and CEC2018 LSGO competitions, respectively.
%
MTS-LS1 is similar to HJ.
Their two essential differences are as follows.
MTS-LS1 does not adopt the exploratory move.
MTS-LS1 also reinitializes the step-size $\sigma$ for the perturbation operator when $\sigma$ is smaller than a pre-defined threshold.

BSrr (also known as NDsqistep and BrentSTEPrr) \cite{PosikB15,BaudisP15} is one of the state-of-the-art optimizers for the five separable noiseless BBOB functions ($f_1, \dots, f_5$) \cite{hansen2012fun}.
Some previous studies on feature-based algorithm selection \cite{KerschkeT19,DerbelLVAT19,JankovicPED21} incorporated BSrr into algorithm portfolios.
The Brent-STEP algorithm \cite{BaudisP15} is a hybrid algorithm of two line search methods: the Brent method \cite{Brent73} and STEP \cite{LangermanSB94}, which perform particularly well for unimodal and multimodal functions, respectively.
Roughly speaking, the Brent-STEP algorithm first applies the Brent method to a partition of the one-dimensional space.
If the search fails, the Brent-STEP algorithm then runs STEP.
Note that the Brent-STEP algorithm is available only for univariate optimization.
BSrr is an extended version of the Brent-STEP algorithm for multivariate optimization.
BSrr applies the Brent-STEP algorithm to each variable in a round-robin manner.




%

We believe that the separability rarely appears in real-world applications.
However, just in case, it is better for an algorithm portfolio and a hybrid method to contain an optimizer that can exploit the separability, e.g., BSrr.
This paper investigates the performance of HJ, MTS-LS1, and BSrr on the large-scale BBOB function set \cite{varelas2020benchmarking} to address the following three research questions:


\noindent \textbf{RQ1} \textit{How does the performance of BSrr scale to large dimensions?}
%
BSrr was not designed for large-scale optimization.
BSrr has been benchmarked on problems with up to 20 dimensions.
Thus, we are interested whether BSrr works well even for more than 20 dimensions.

\noindent \textbf{RQ2} \textit{Does MTS-LS1 perform well on the large-scale BBOB function set?}
%
MTS-LS1 has not been benchmarked on any BBOB function set.
We are interested whether MTS-LS1 can handle various transformations (e.g., asymmetry breaking) in the BBOB functions.
We are also interested in comparing MTS-LS1 with BSrr, which has shown its effectiveness on the five separable noiseless BBOB functions.



\noindent \textbf{RQ3} \textit{Do HJ and MTS-LS1 perform similar?}
Although HJ and MTS-LS1 are very similar, their performance has not been compared.
It is interesting to compare MTS-LS1 proposed in 2008 and HJ proposed in 1961 in a systematic manner.


 \section{Experimental setup}

We implemented HJ and MTS-LS1 in C.
Their code is available at \url{https://github.com/ryojitanabe/largebbob2022}.
We believe that there are no ``\textit{the} HJ'' and ``\textit{the} MTS-LS1''.
We referred to the description in \cite{Caraffini14} to implement HJ.
We also referred to the code of MTS-LS1 in MOS (\url{https://sci2s.ugr.es/EAMHCO}) to implement MTS-LS1.
For the sake of clarity, Algorithms \ref{alg:hj} and \ref{alg:mtsls1} show \textit{the} HJ and \textit{the} MTS-LS1 which we understood.
Here, $\vector{x} = (x_1, ..., x_D)^{\top}$ is the current $D$-dimensional search point, $\vector{x}^{\mathrm{prev}}$ is the previous one, $\vector{x}^{\mathrm{new}}$ is the new one, $\vector{x}^{\mathrm{upper}}$ and $\vector{x}^{\mathrm{lower}}$ are the upper and lower bounds of a problem.
In addition, $\sigma$ is the step-size, $\sigma^{\mathrm{init}}$ is the initial value of $\sigma$, and $c$ is the learning rate for $\sigma$.
If $x^{\mathrm{new}}_i$ is out of the range $[x^{\mathrm{lower}}_i, x^{\mathrm{upper}}_i]$, $x^{\mathrm{new}}_i$ is replaced with the closest bound.
We used the Python implementation of BSrr provided by the authors of \cite{BaudisP15} (\url{https://github.com/pasky/step}).

For HJ and MTS-LS1, we set $\sigma^{\mathrm{init}} = 0.4$ and $c=0.5$ as in \cite{TsengC08}.
We also investigate the performance of HJ and MTS-LS1 with $c=0.9$.
We denote HJ and MTS-LS1 with $c=0.5$ and $c=0.9$ as ``HJ-5'', ``HJ-9'', ``MTS-LS1-5'', and ``MTS-LS1-9'', respectively.
We used the default parameter setting for BSrr.
We set the maximum number of function evaluations (\texttt{max\_budget}) for HJ and MTS-LS1 to $10^4 \times D$.
Because BSrr implemented in Python is time-consuming for large dimensions, we set \texttt{max\_budget} for BSrr to $10^3 \times D$.



\IncMargin{0.5em}
\begin{algorithm}[t]
\small
\SetSideCommentRight
Initialize $\vector{x}$, $\sigma \leftarrow \sigma^{\mathrm{init}}$\;
\While{not happy}{
  $\vector{x}^{\mathrm{prev}} \leftarrow \vector{x}$\;  
  \For{$i \in \{1, \dots, D\}$}{
    $\vector{x}^{\mathrm{new}} \leftarrow \vector{x}$, $x^{\mathrm{new}}_i \leftarrow x_i + \sigma (x^{\mathrm{upper}}_i - x^{\mathrm{lower}}_i)$\;    
    \lIf{$f(\vector{x}^{\mathrm{new}}) < f(\vector{x})$}{
      $\vector{x} \leftarrow \vector{x}^{\mathrm{new}}$
    }
    \Else{
      $\vector{x}^{\mathrm{new}} \leftarrow \vector{x}$, $x^{\mathrm{new}}_i \leftarrow x_i - \sigma (x^{\mathrm{upper}}_i - x^{\mathrm{lower}}_i)$\;       
      \lIf{$f(\vector{x}^{\mathrm{new}}) < f(\vector{x})$}{
        $\vector{x} \leftarrow \vector{x}^{\mathrm{new}}$
      }        
    }     
  }
  \uIf{$f(\vector{x}) < f(\vector{x}^{\mathrm{prev}})$}{
    $\vector{x}^{\mathrm{new}} \leftarrow \vector{x} + (\vector{x} - \vector{x}^{\mathrm{prev}})$\;
    \lIf{$f(\vector{x}^{\mathrm{new}}) < f(\vector{x})$}{
      $\vector{x} \leftarrow \vector{x}^{\mathrm{new}}$
    }                
  }
  \lElse{
    $\sigma \leftarrow c \times \sigma$
  }      
}
 \caption{The Hooke-Jeeves method (HJ)}
\label{alg:hj}
\end{algorithm}\DecMargin{0.5em}

\IncMargin{0.5em}
\begin{algorithm}[t]
\small
\SetSideCommentRight
Initialize $\vector{x}$, $\sigma \leftarrow \sigma^{\mathrm{init}}$\;
\While{not happy}{
  $\vector{x}^{\mathrm{prev}} \leftarrow \vector{x}$\;    
  \For{$i \in \{1, \dots, D\}$}{
    $\vector{x}^{\mathrm{new}} \leftarrow \vector{x}$, $x^{\mathrm{new}}_i \leftarrow x_i - \sigma (x^{\mathrm{upper}}_i - x^{\mathrm{lower}}_i)$\;
      \lIf{$f(\vector{x}^{\mathrm{new}}) < f(\vector{x})$}{
        $\vector{x} \leftarrow \vector{x}^{\mathrm{new}}$
      }
      \Else{
        $\vector{x}^{\mathrm{new}} \leftarrow \vector{x}$, $x^{\mathrm{new}}_i \leftarrow x_i + 0.5 \times \sigma (x^{\mathrm{upper}}_i - x^{\mathrm{lower}}_i)$\;
        \lIf{$f(\vector{x}^{\mathrm{new}}) < f(\vector{x})$}{
          $\vector{x} \leftarrow \vector{x}^{\mathrm{new}}$
        }        
      }
      
  }

  \If{$f(\vector{x}) = f(\vector{x}^{\mathrm{prev}})$}{          
        $\sigma \leftarrow c \times \sigma$\;
        \lIf{$\sigma (x^{\mathrm{upper}}_1 - x^{\mathrm{lower}}_1) < 10^{-15}$}{
          $\sigma \leftarrow \sigma^{\mathrm{init}}$
        }
      }  
}
 \caption{MTS-LS1}
\label{alg:mtsls1}
\end{algorithm}\DecMargin{0.5em}

\section{CPU Timing}

In order to evaluate the CPU timing of the algorithms, we have run the three optimizers \textit{without} restarts on the entire {\ttfamily bbob-largescale} test suite \cite{varelas2020benchmarking} for $2 D$ function evaluations according to \cite{hansen2016exp}.
We conducted our experiments on a Ubuntu 18.04 with Intel(R) 52-Core Xeon Platinum 8270 (26-Core$\times 2$) 2.7GHz and (compile) options \texttt{-O2}.
Table \ref{tab:time} shows the computation time.
As seen from Table \ref{tab:time}, the C code is much faster than the Python code in terms of the CPU time.


\begin{table}[t]
\setlength{\tabcolsep}{1pt} 
\centering
\caption{\small Computation time of the three optimizers ($10^{-5}$ seconds).}
\label{tab:time}
{\footnotesize
\begin{tabular}{lccccccccccccccccccc}
  \toprule
Optimizers & Languages & 20-D & 40-D & 80-D & 160-D & 320-D & 640-D\\
  \midrule
  HJ & C & Na & $4.2$ & $5.9$ & $11$ & $21$ & $41$\\  
  MTS-LS1 & C & $4.1$ & $2.0$ & $5.8$ & $11$ & $21$ & $42$\\
  BSrr & Python & $13$ & $20$ & $33$ & $62$ & $120$ & $270$\\  
  \bottomrule
\end{tabular}
}
\end{table}


\section{Results}

\begin{figure*}[htp]
\centering
\begin{tabular}{@{}c@{}c@{}c@{}c@{}}
\includegraphics[width=0.24\textwidth]{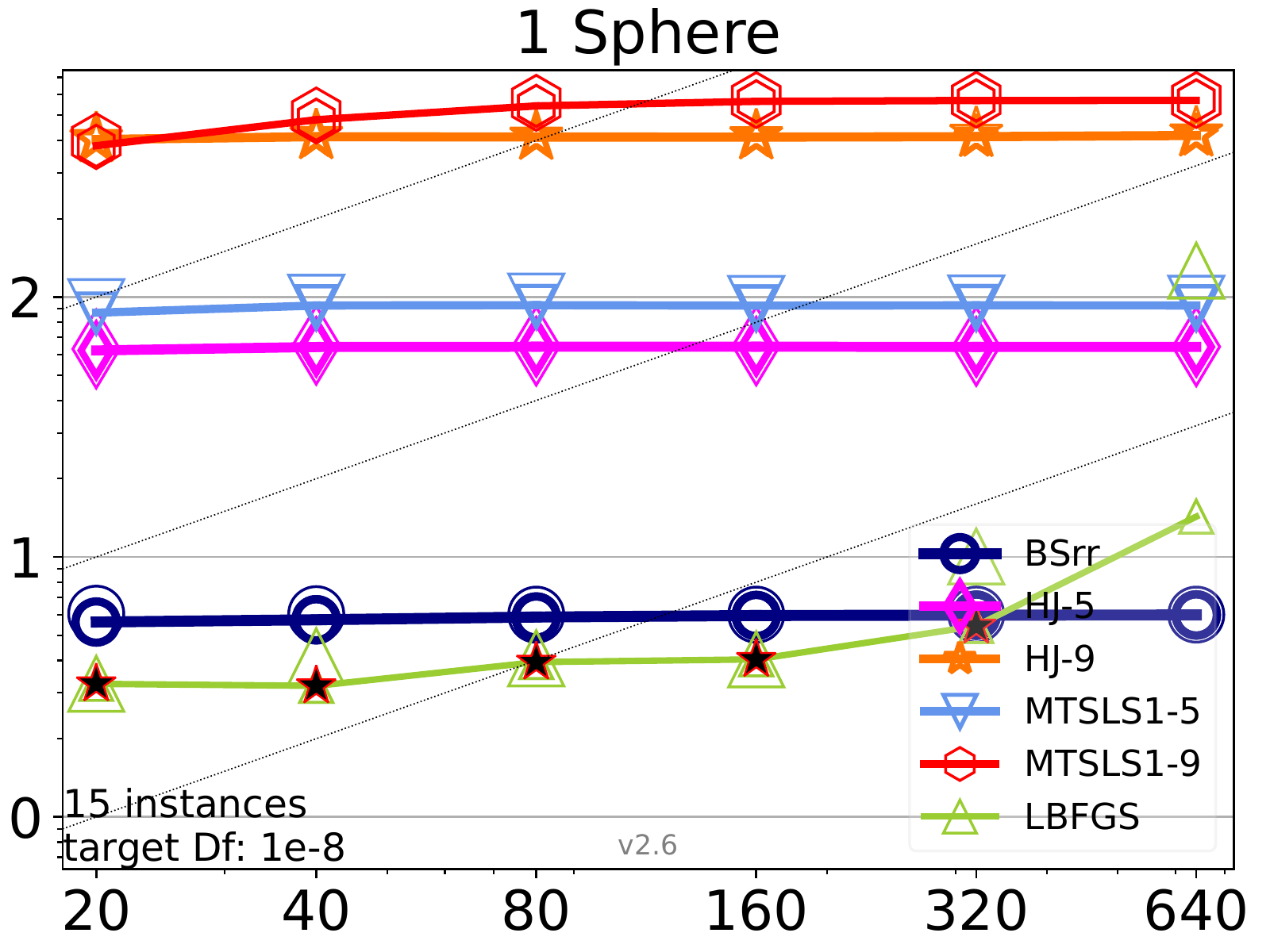}&
\includegraphics[width=0.24\textwidth]{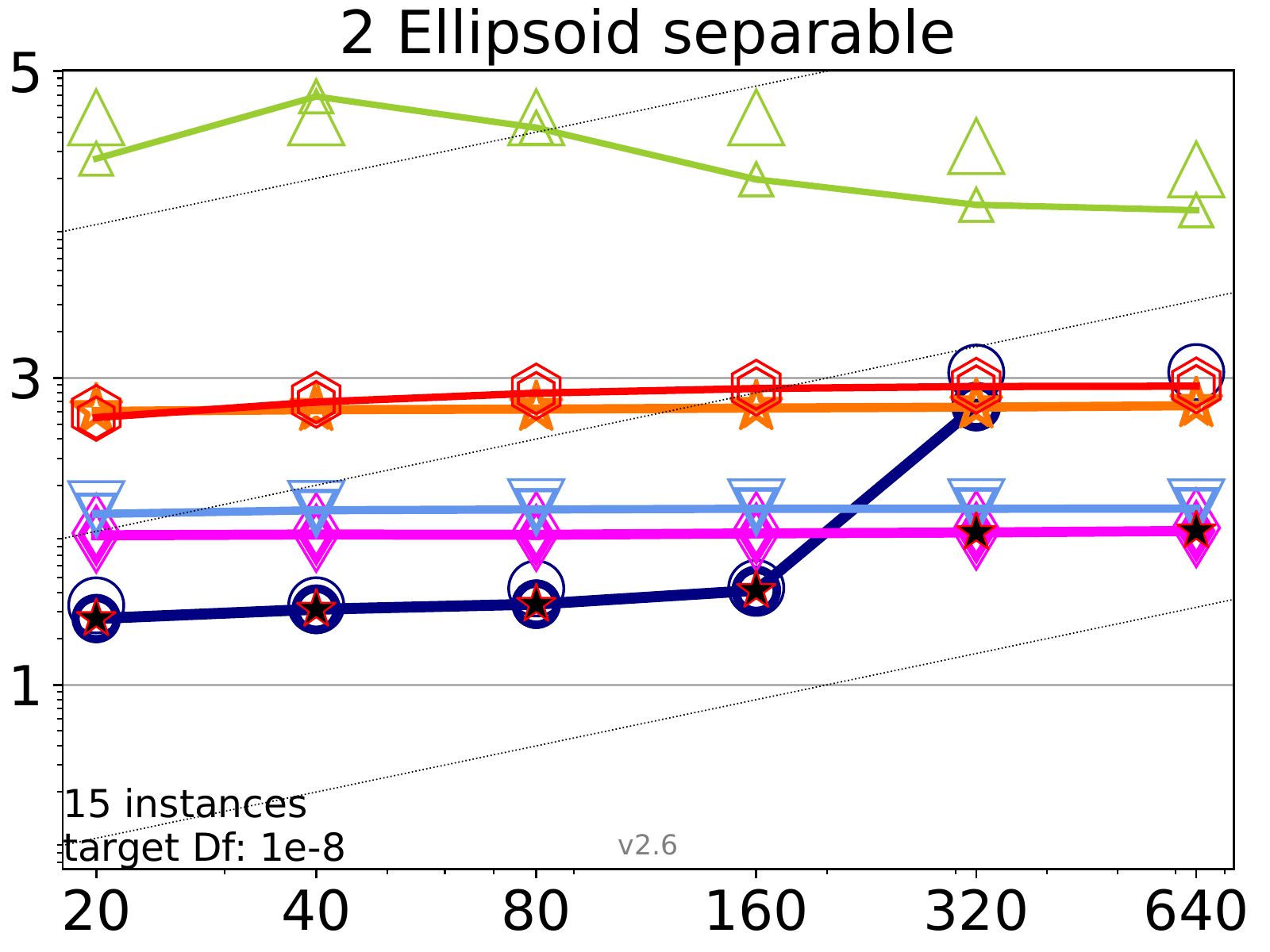}&
\includegraphics[width=0.24\textwidth]{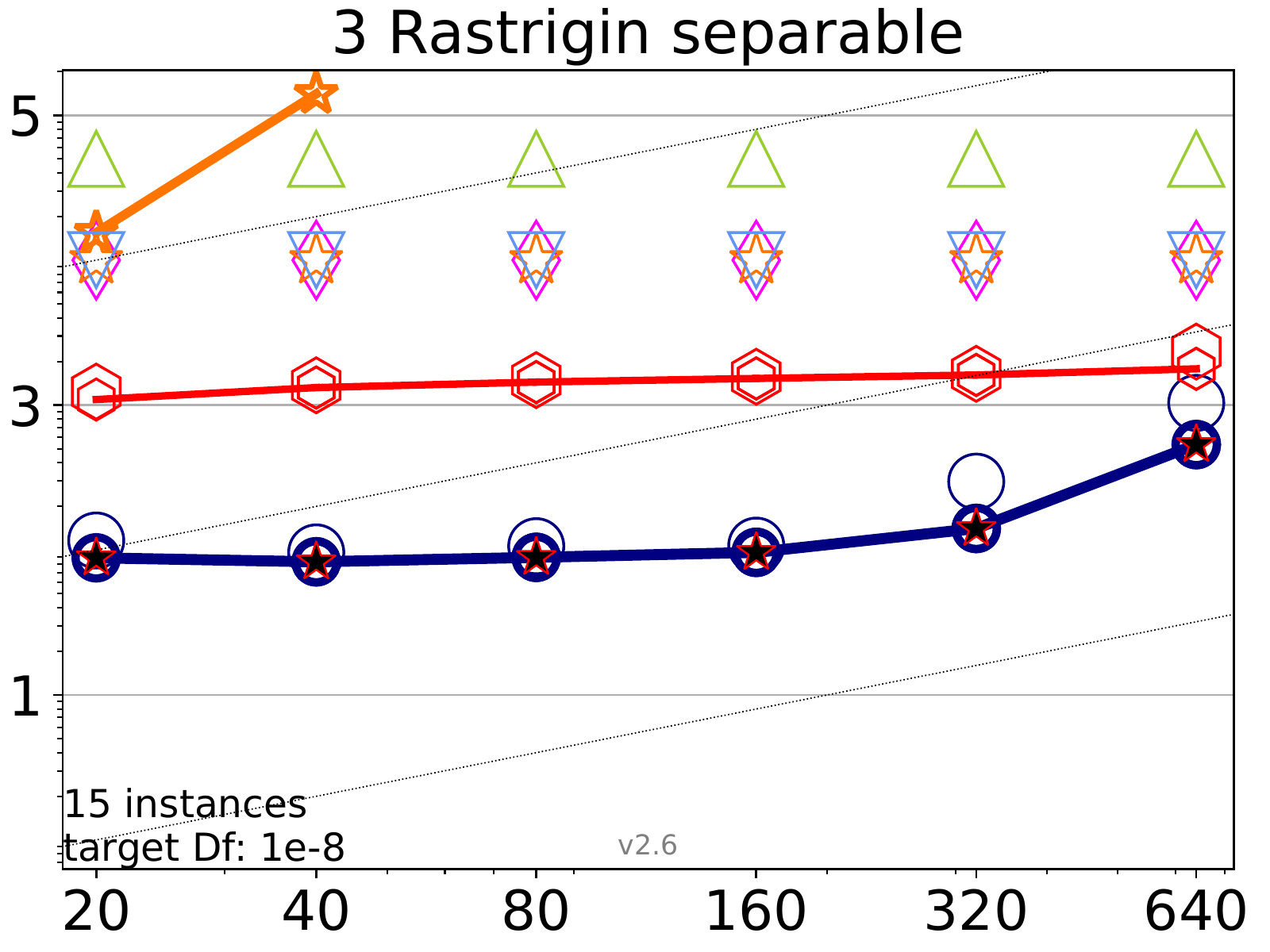}&
\includegraphics[width=0.24\textwidth]{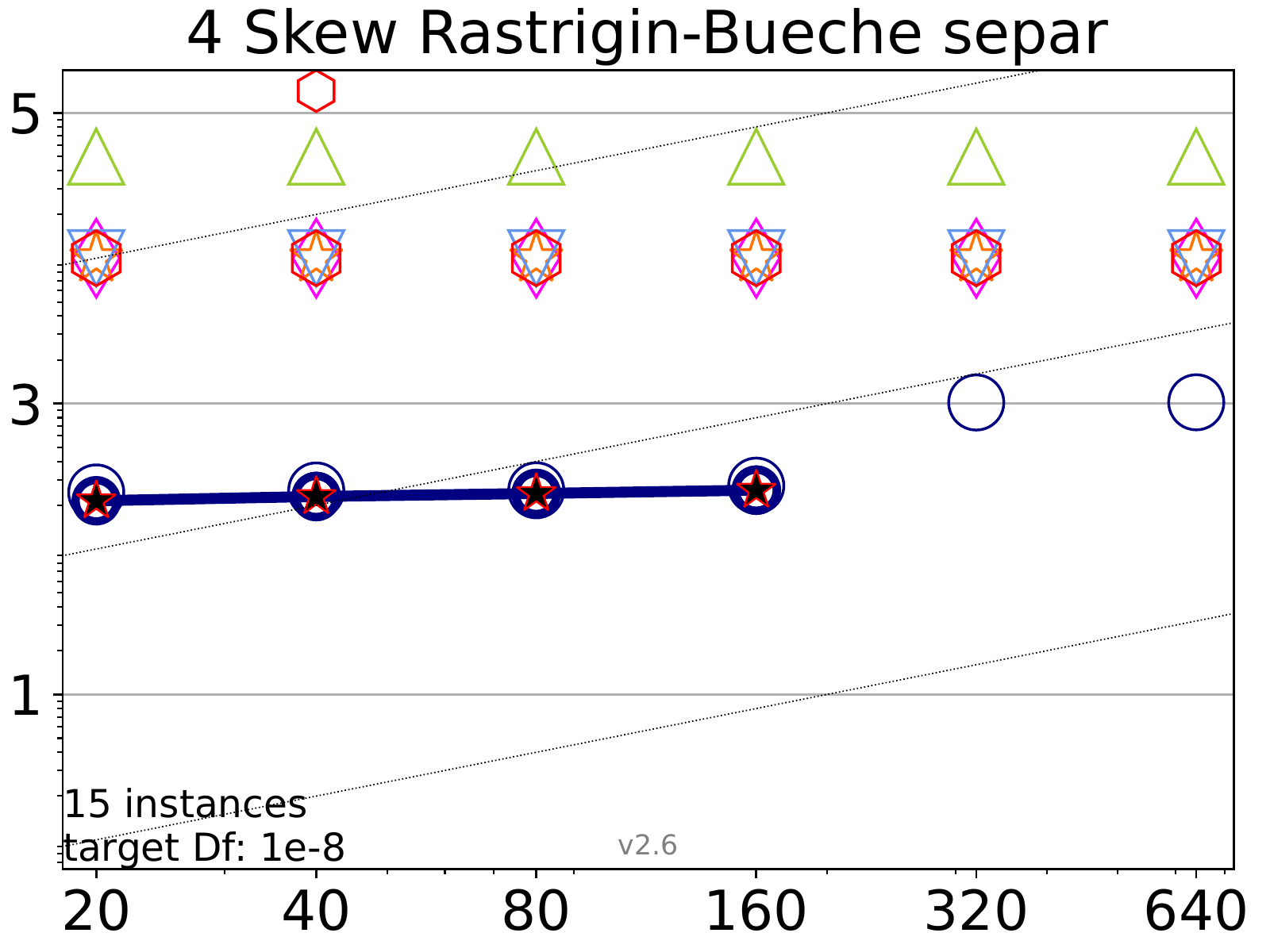}\\[-0.25em]
\includegraphics[width=0.24\textwidth]{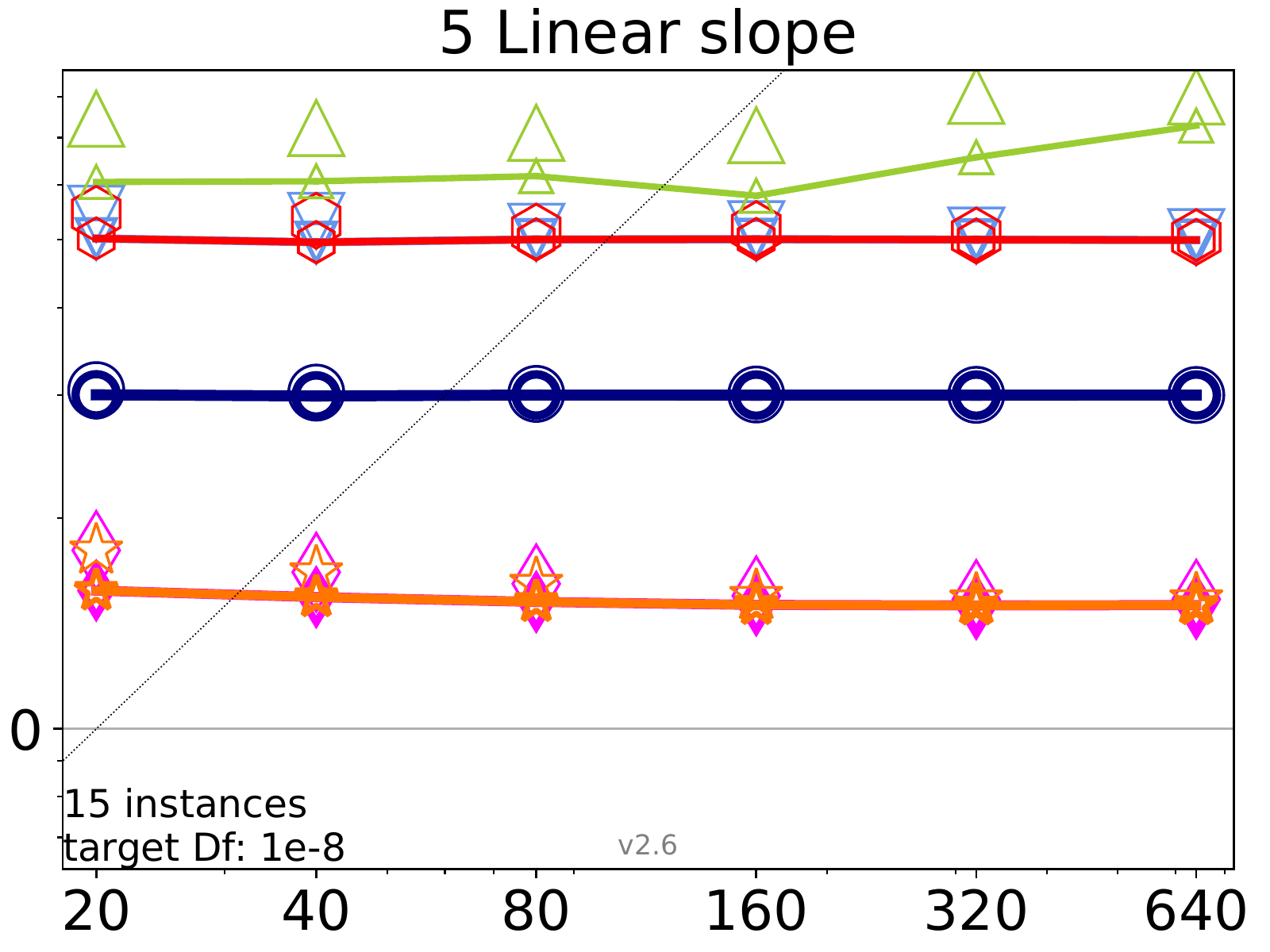}&
\includegraphics[width=0.24\textwidth]{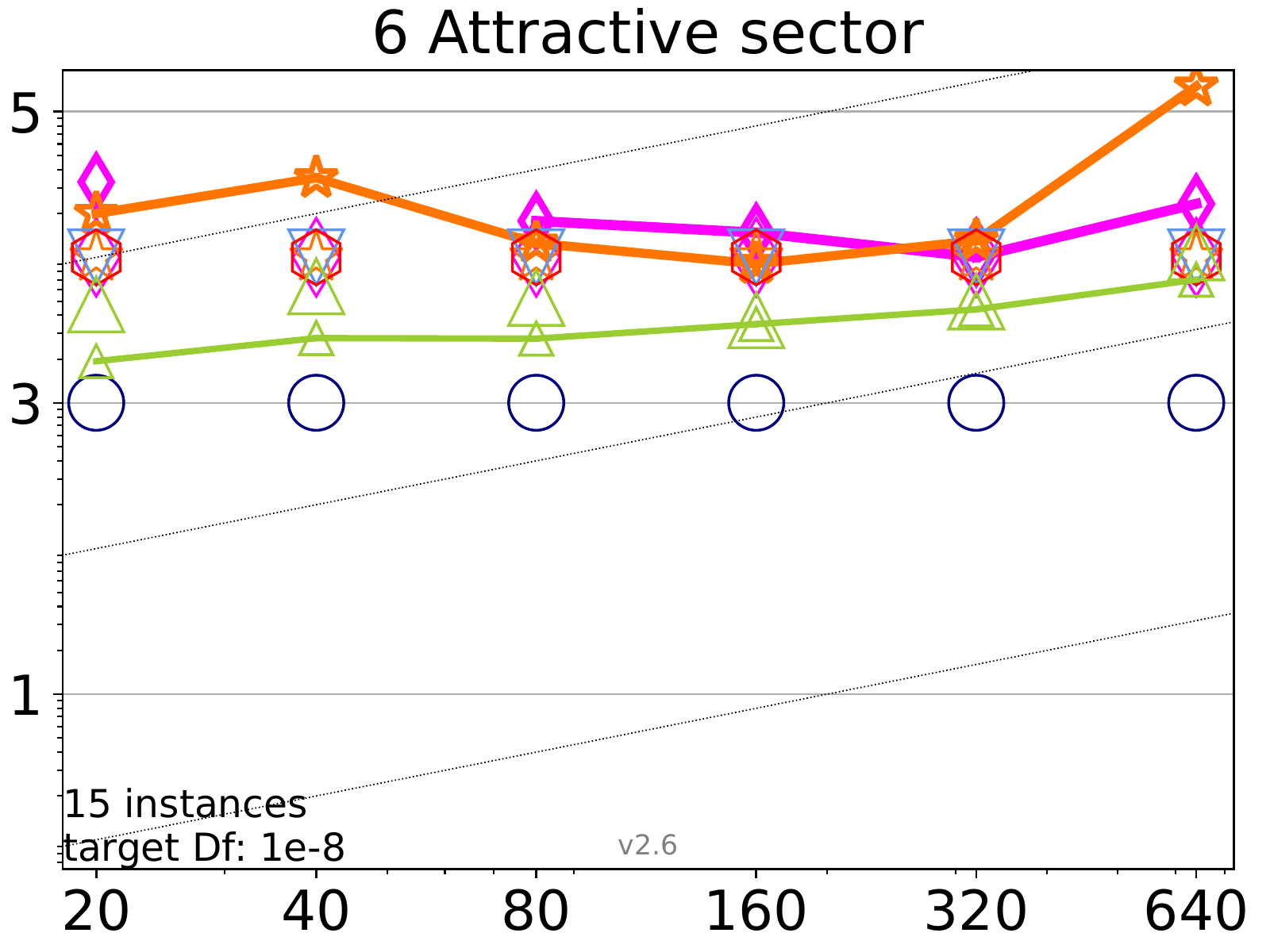}&
\includegraphics[width=0.24\textwidth]{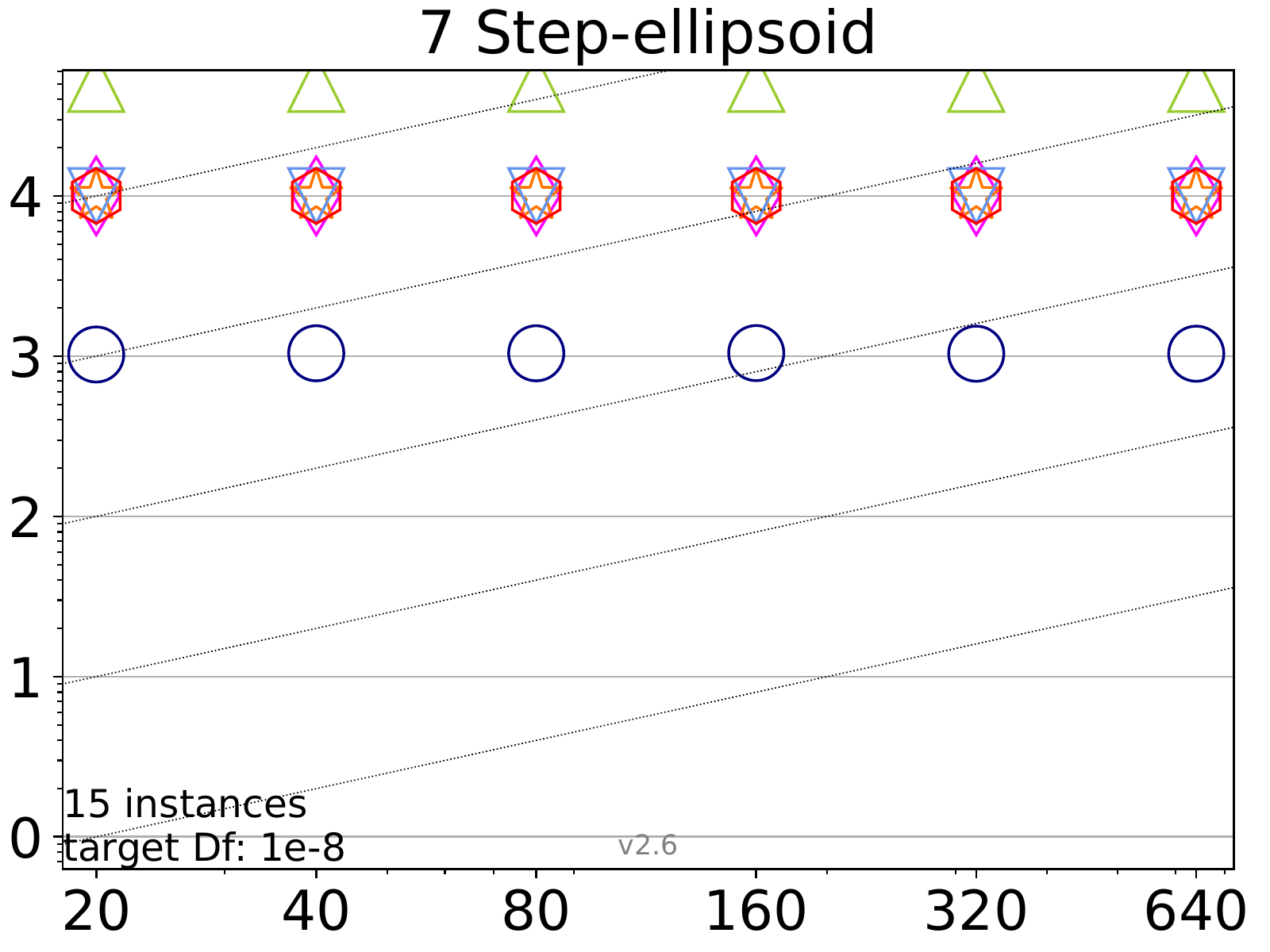}&
\includegraphics[width=0.24\textwidth]{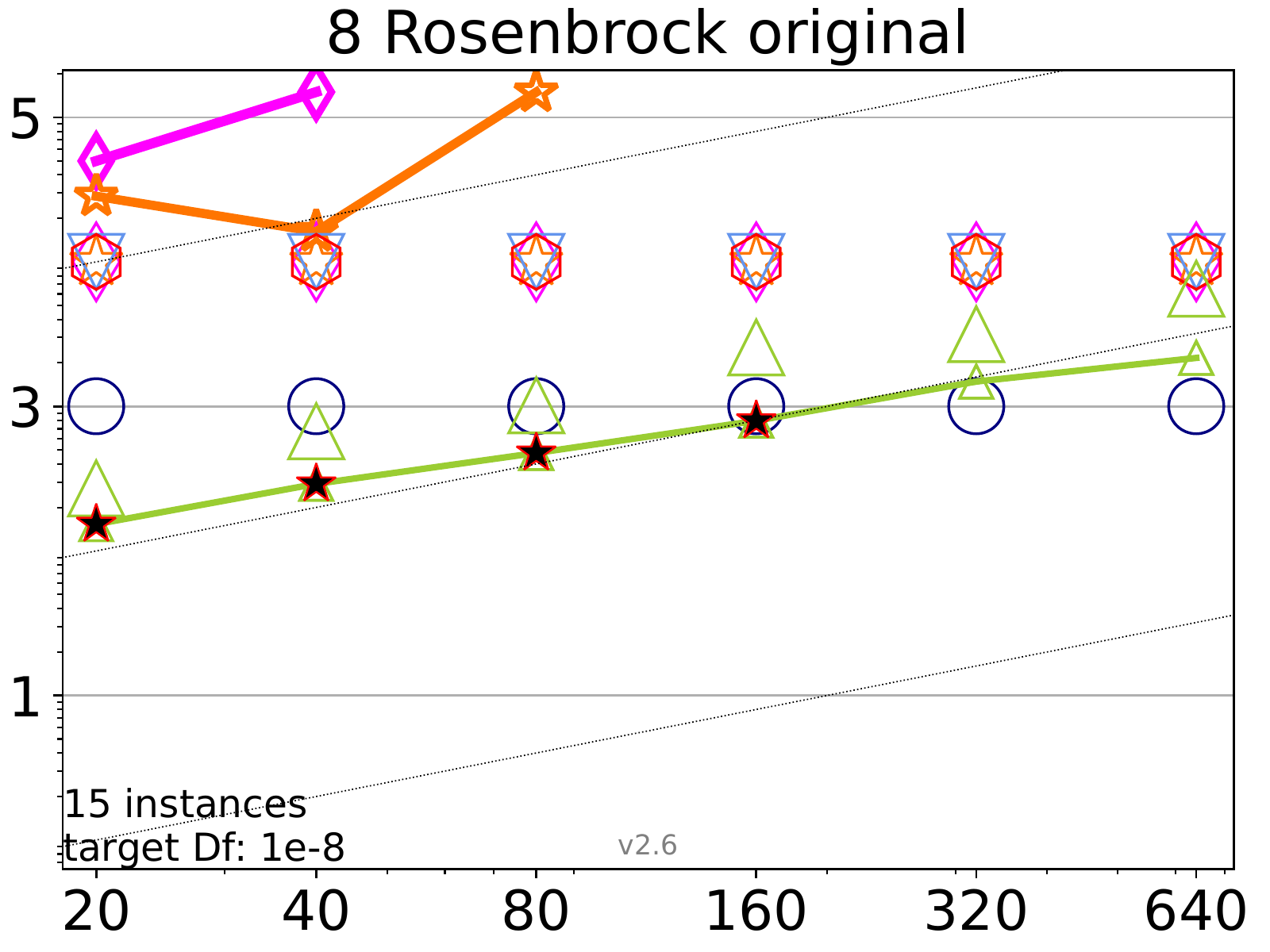}\\[-0.25em]
\includegraphics[width=0.24\textwidth]{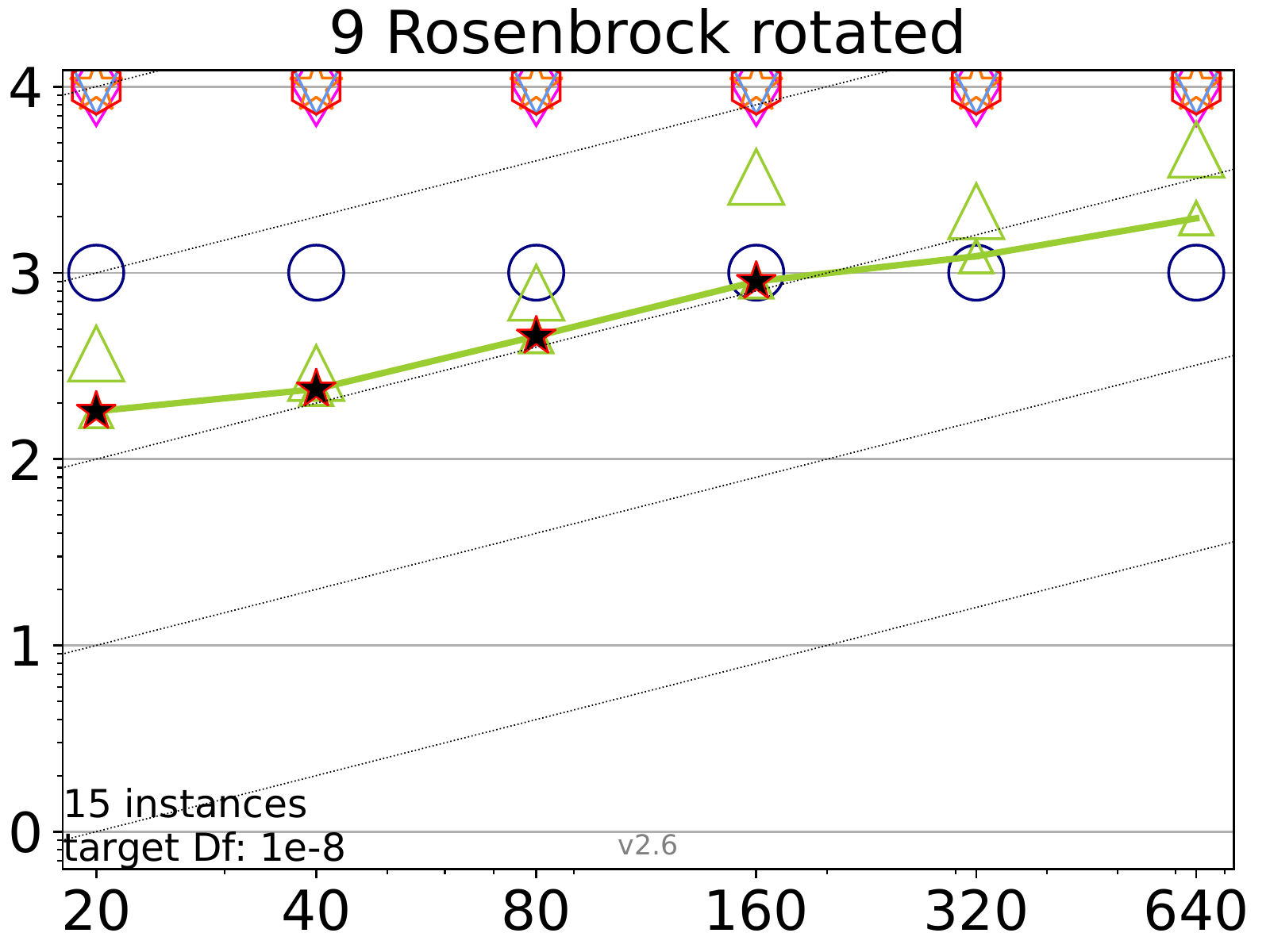}&
\includegraphics[width=0.24\textwidth]{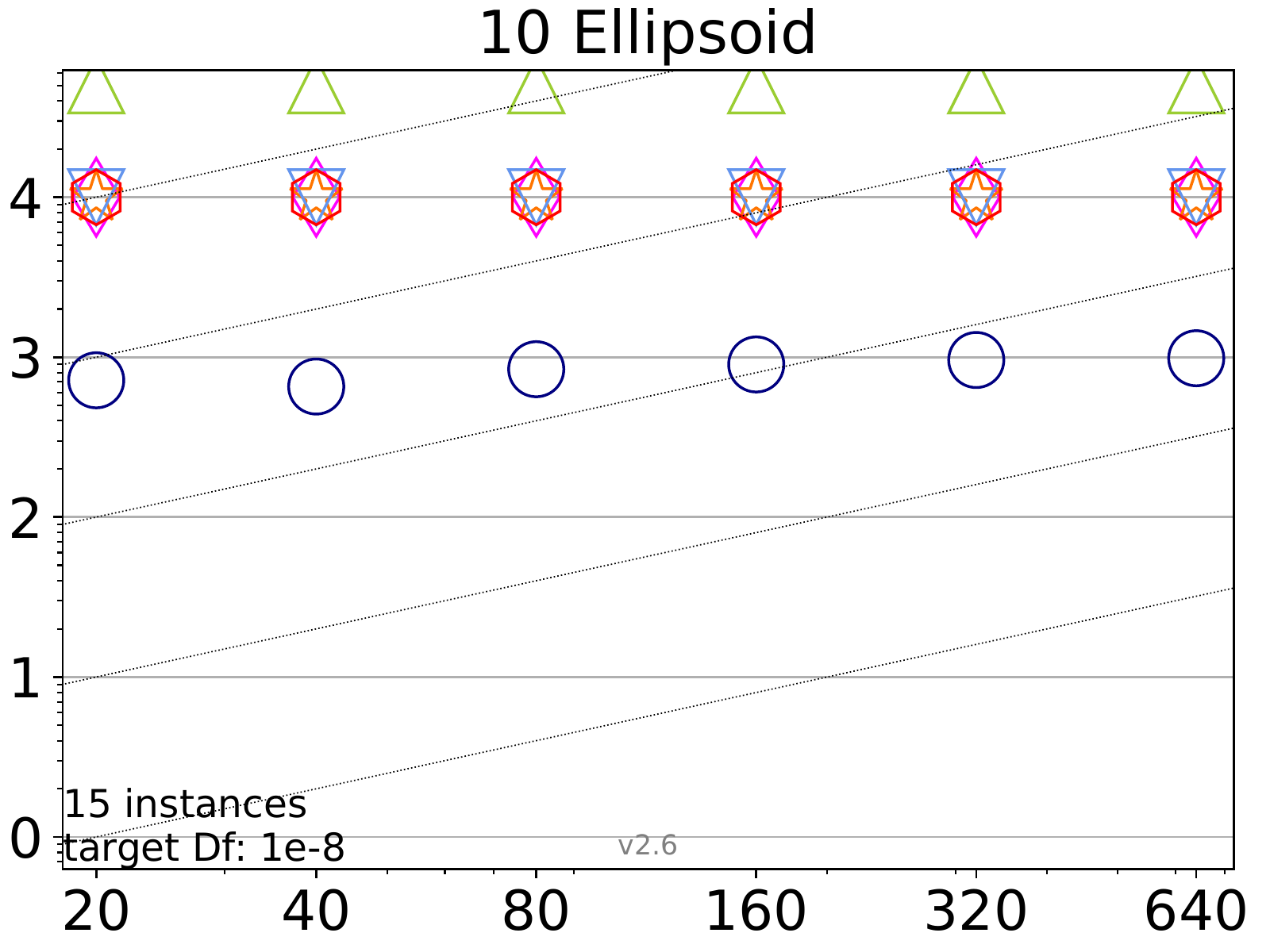}&
\includegraphics[width=0.24\textwidth]{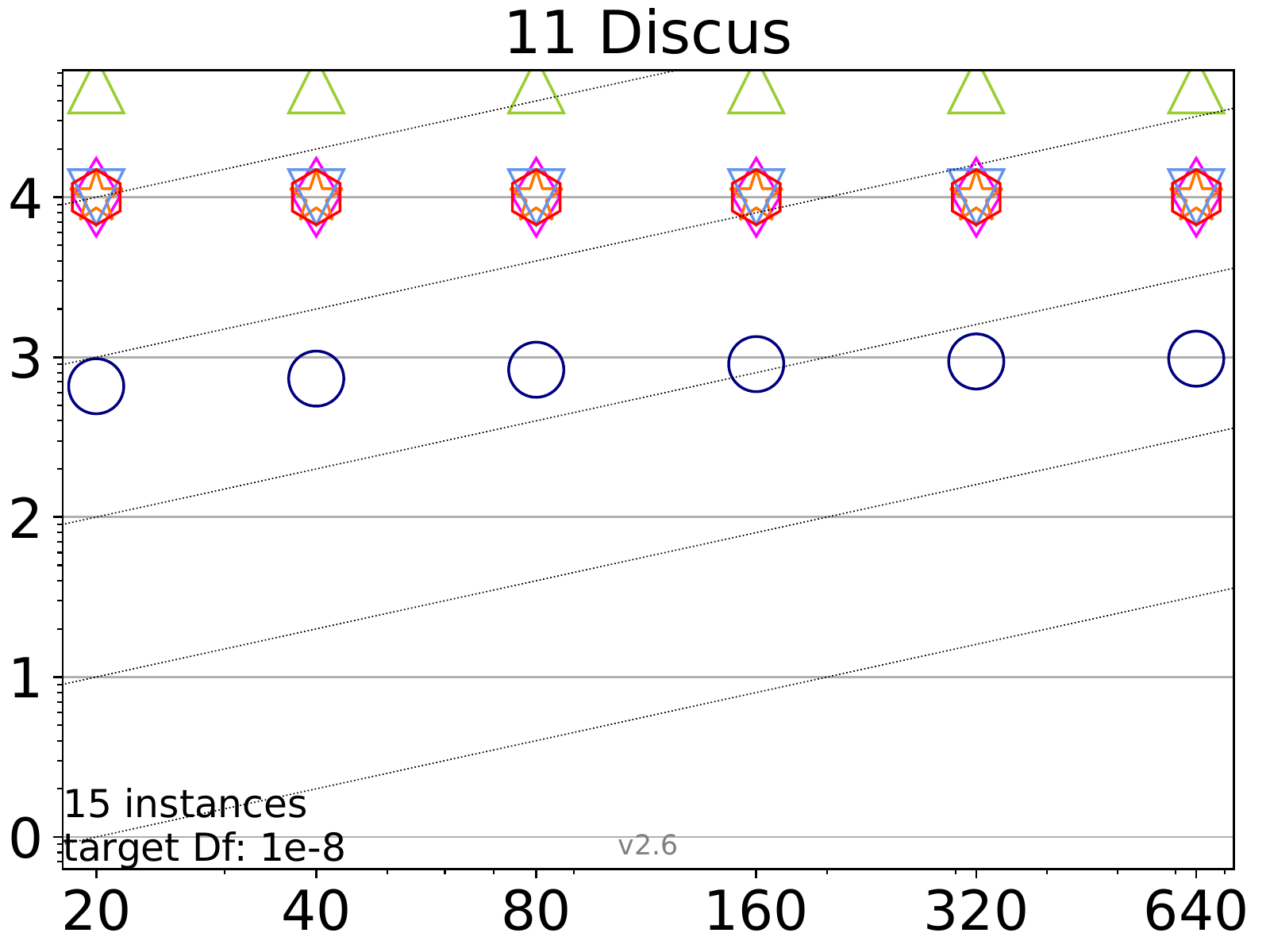}&
\includegraphics[width=0.24\textwidth]{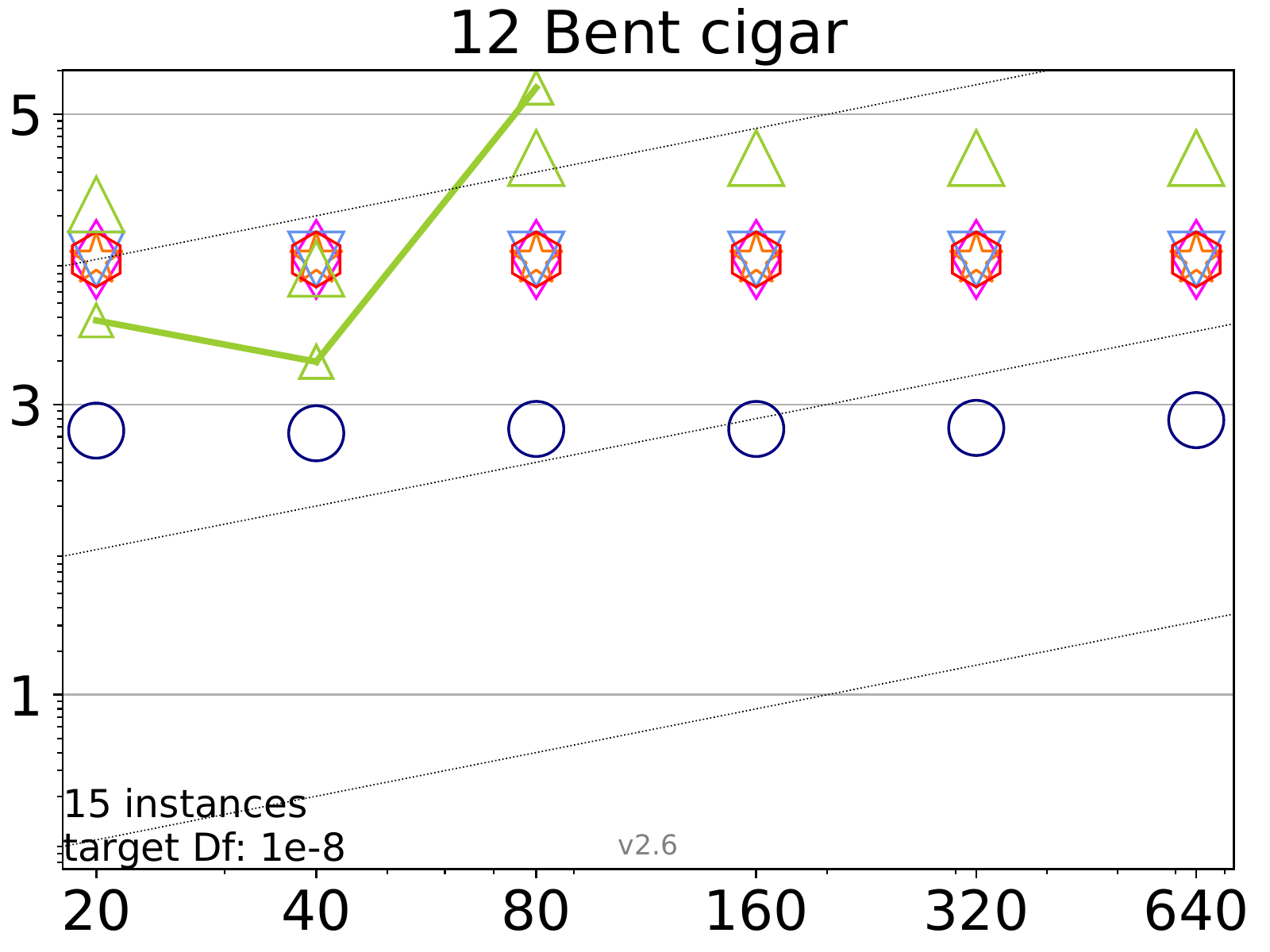}\\[-0.25em]
\includegraphics[width=0.24\textwidth]{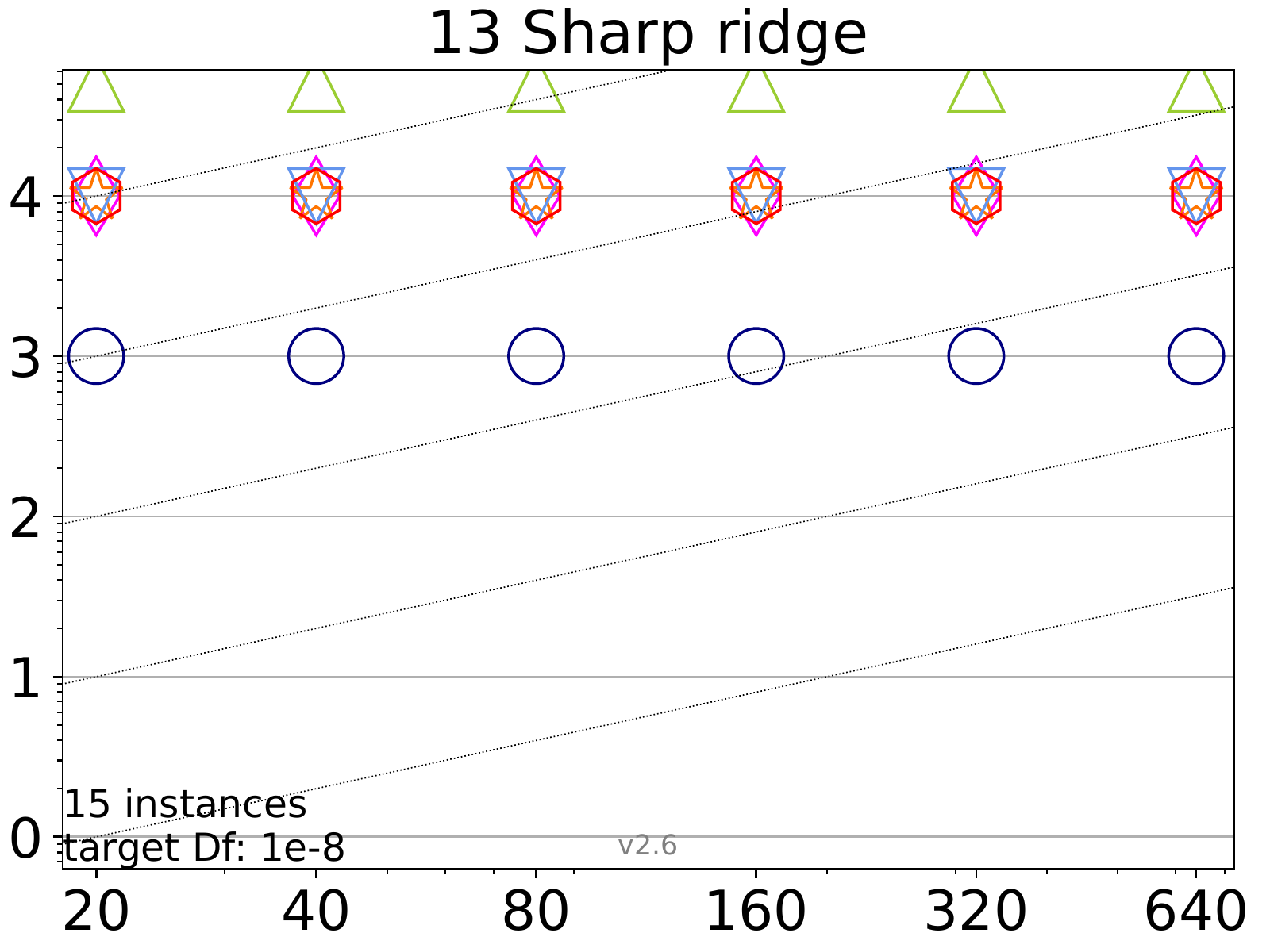}&
\includegraphics[width=0.24\textwidth]{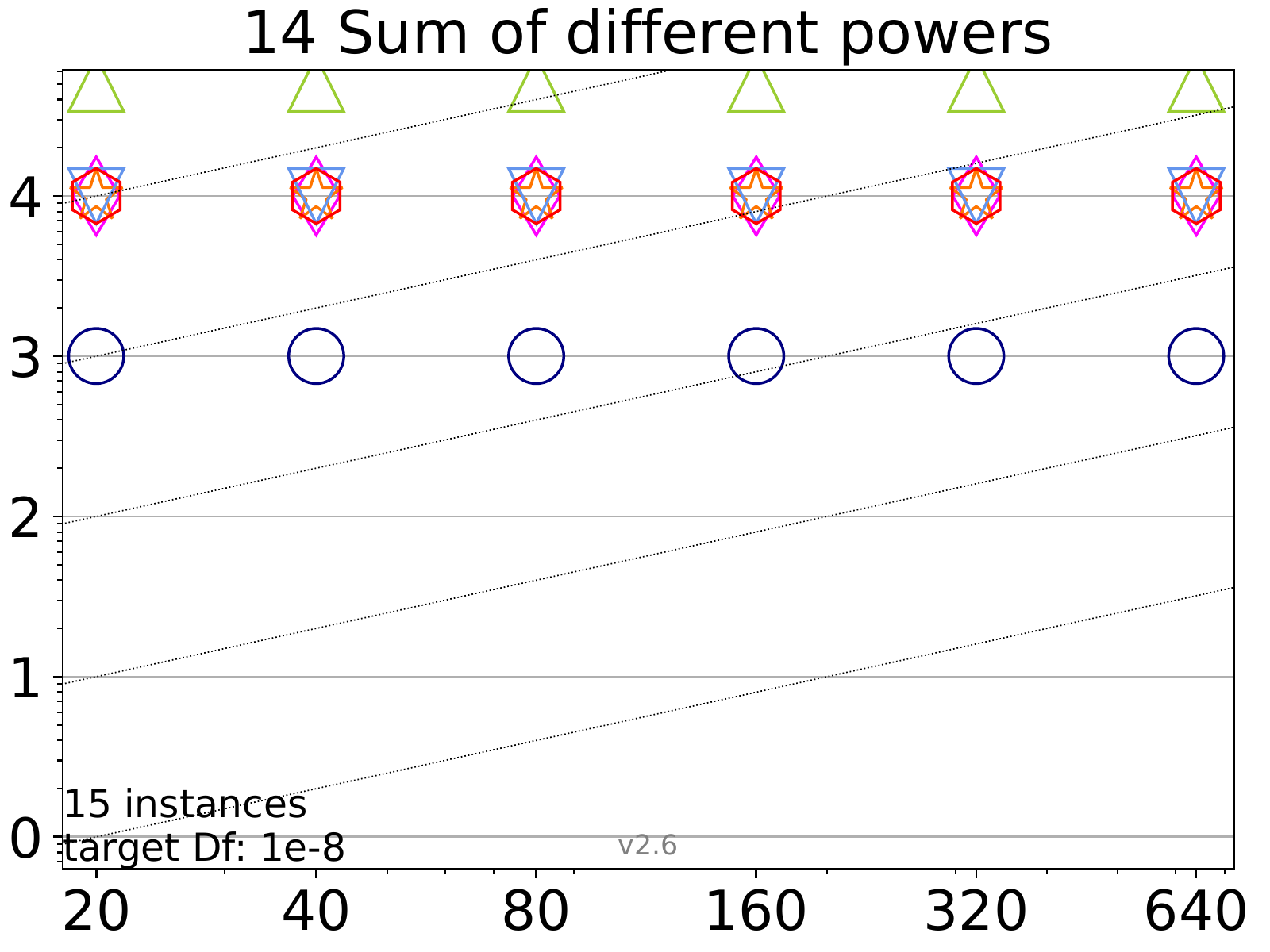}&
\includegraphics[width=0.24\textwidth]{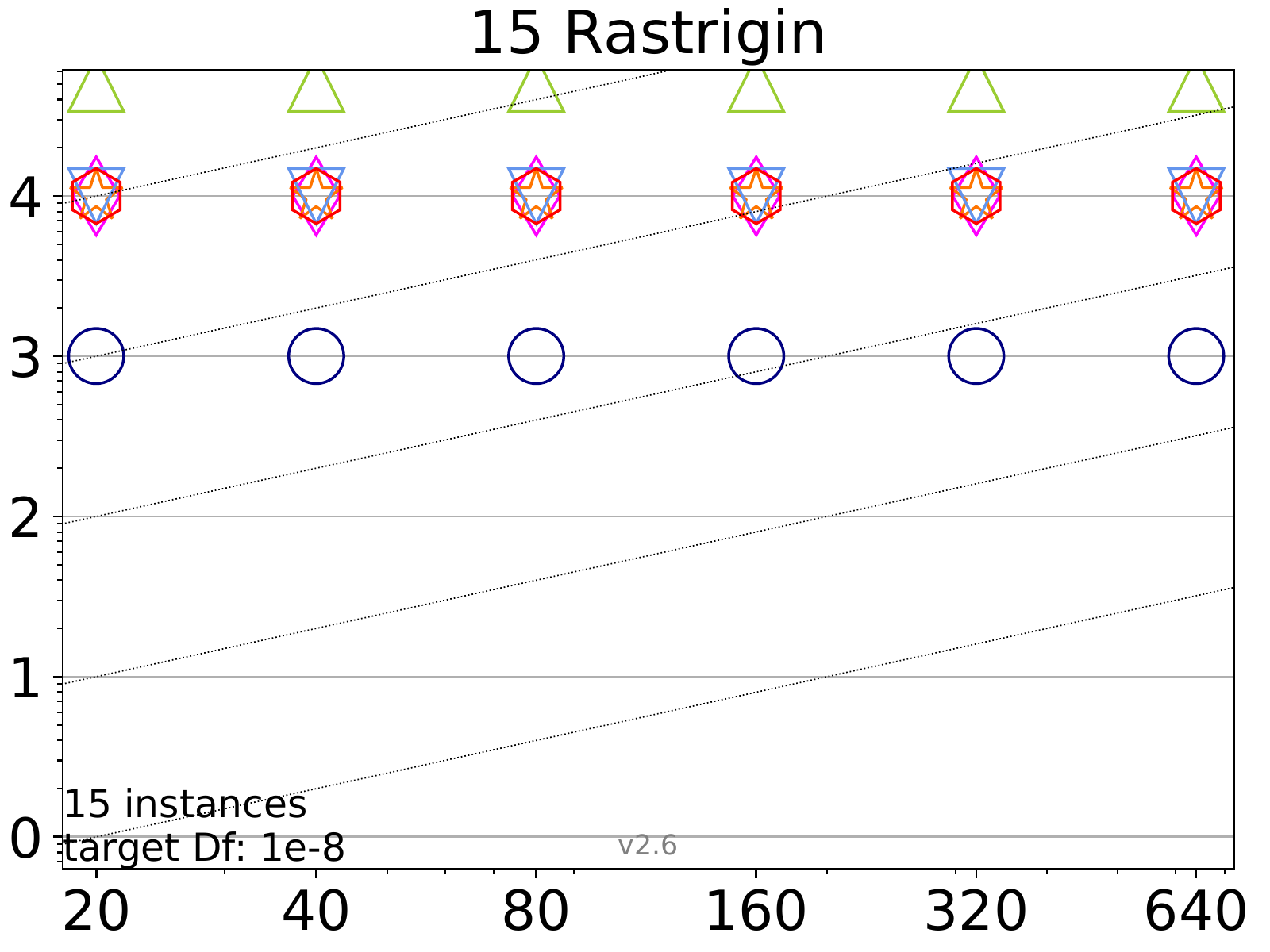}&
\includegraphics[width=0.24\textwidth]{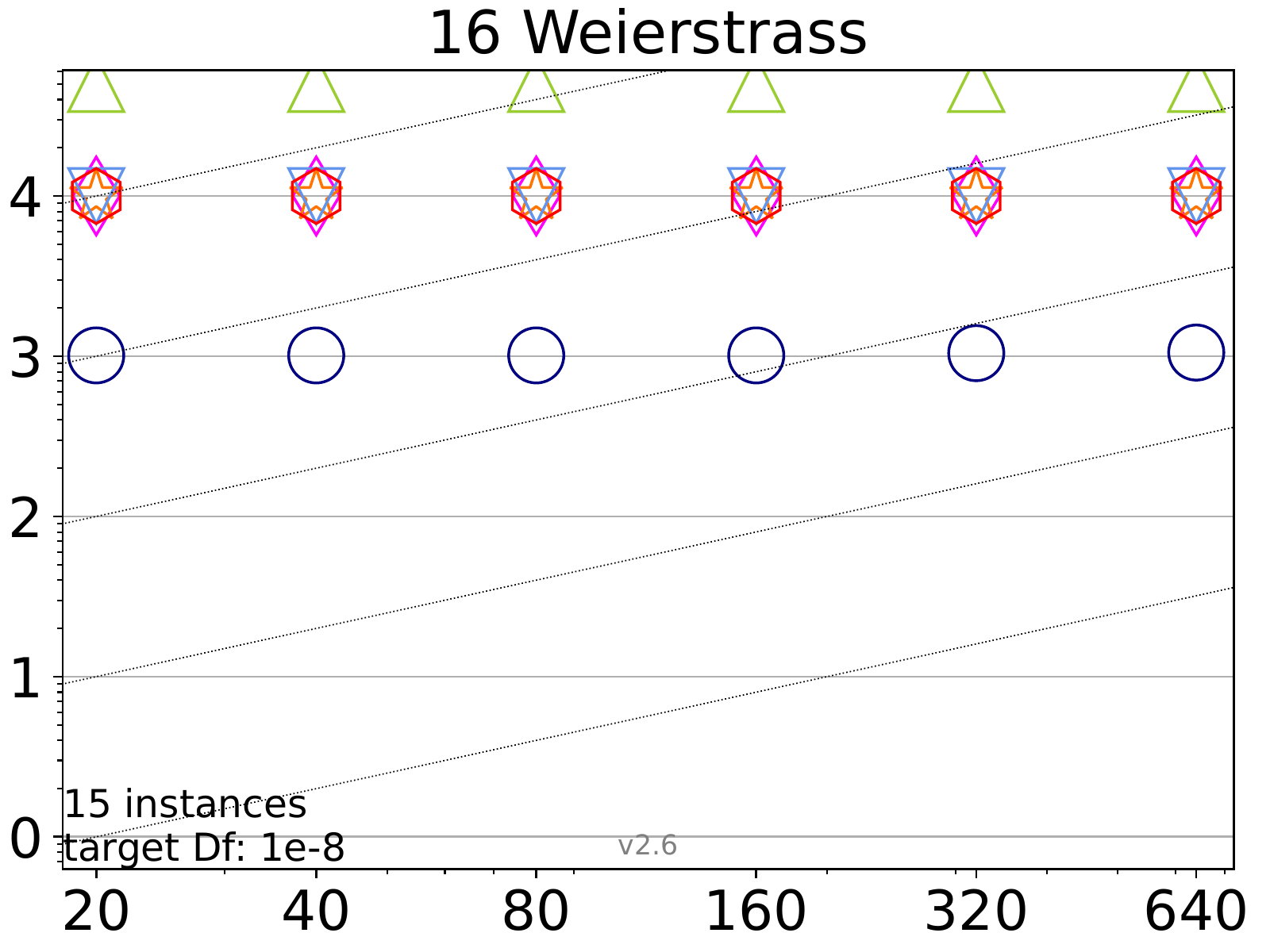}\\[-0.25em]
\includegraphics[width=0.24\textwidth]{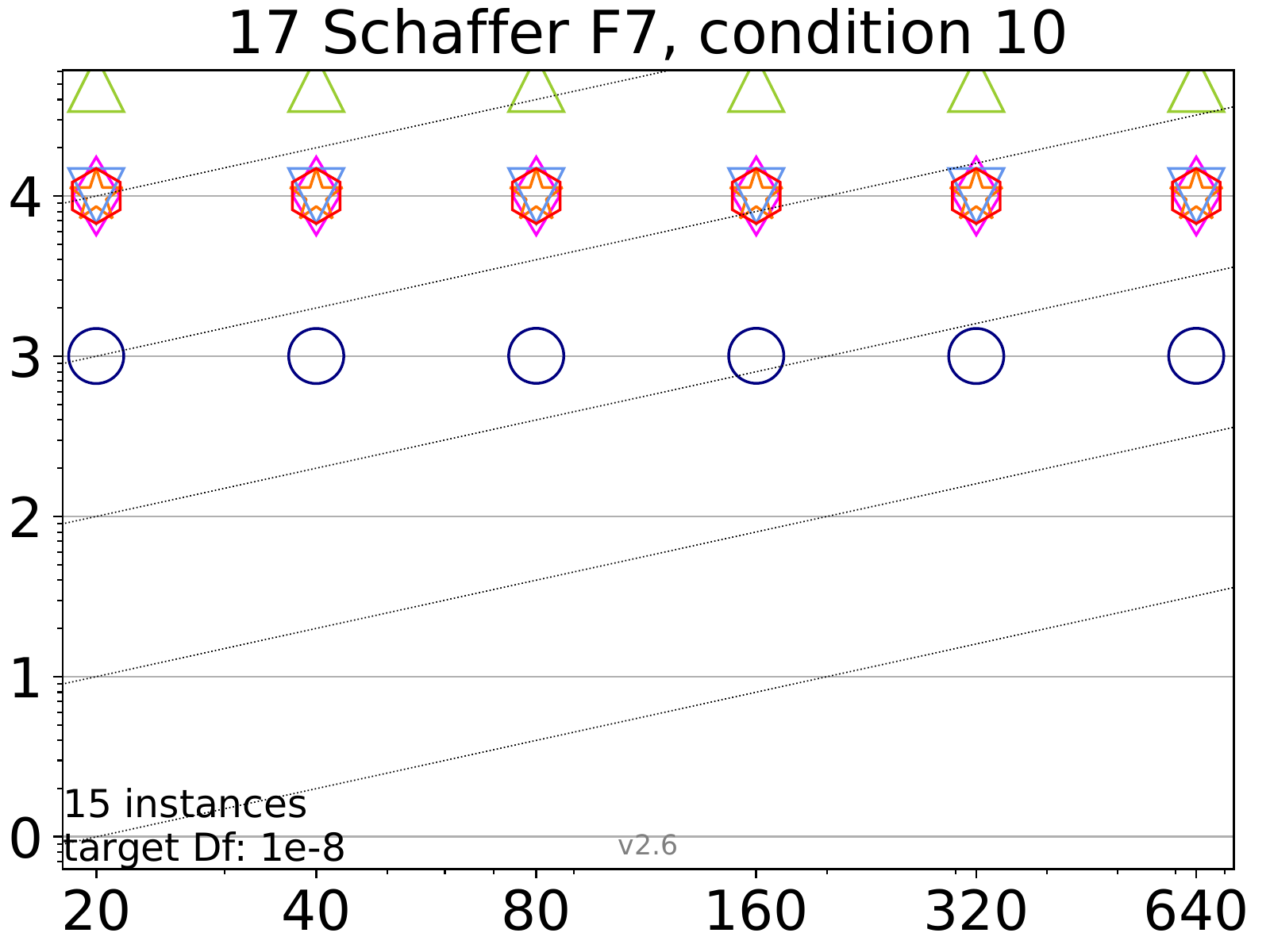}&
\includegraphics[width=0.24\textwidth]{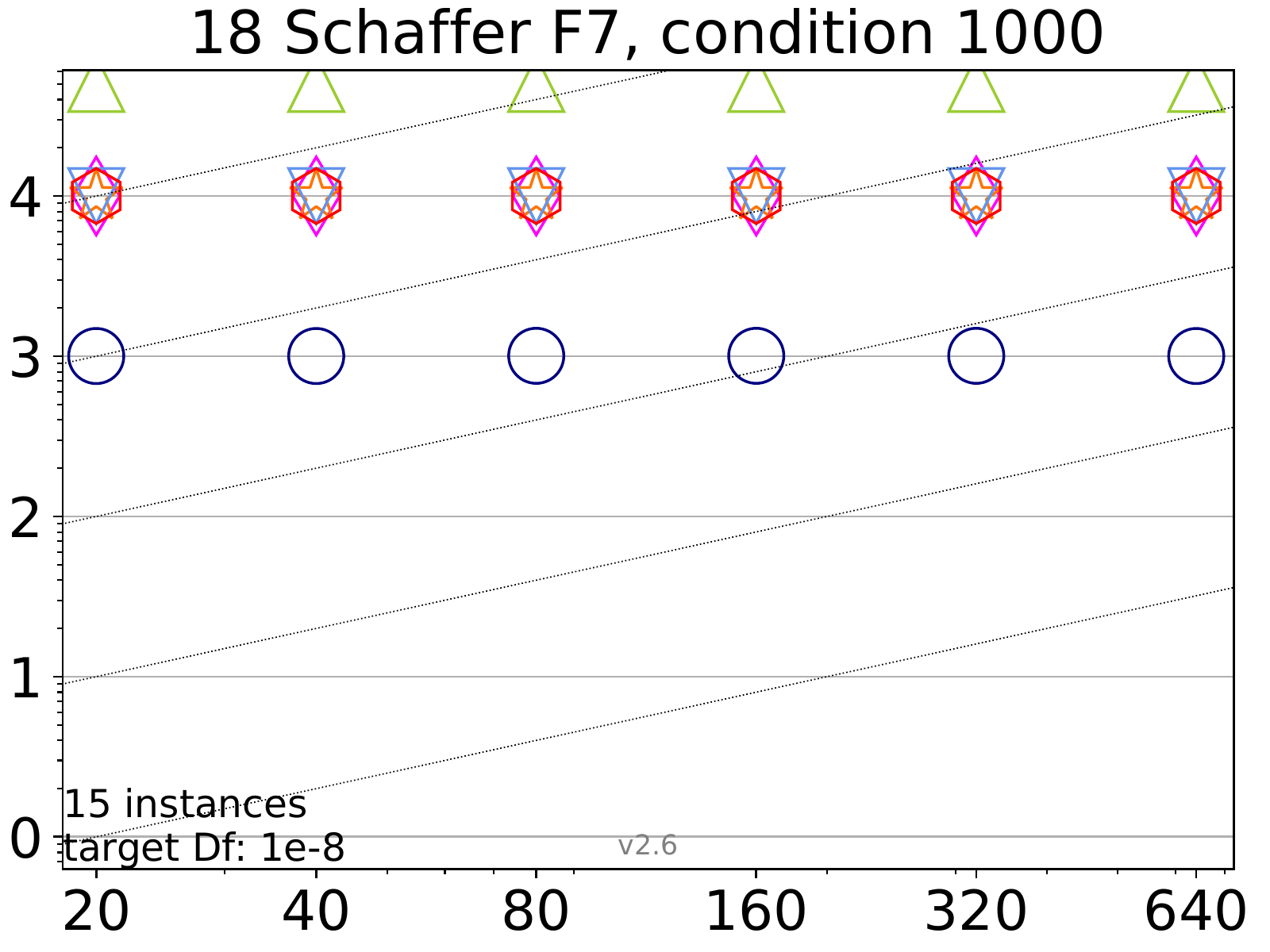}&
\includegraphics[width=0.24\textwidth]{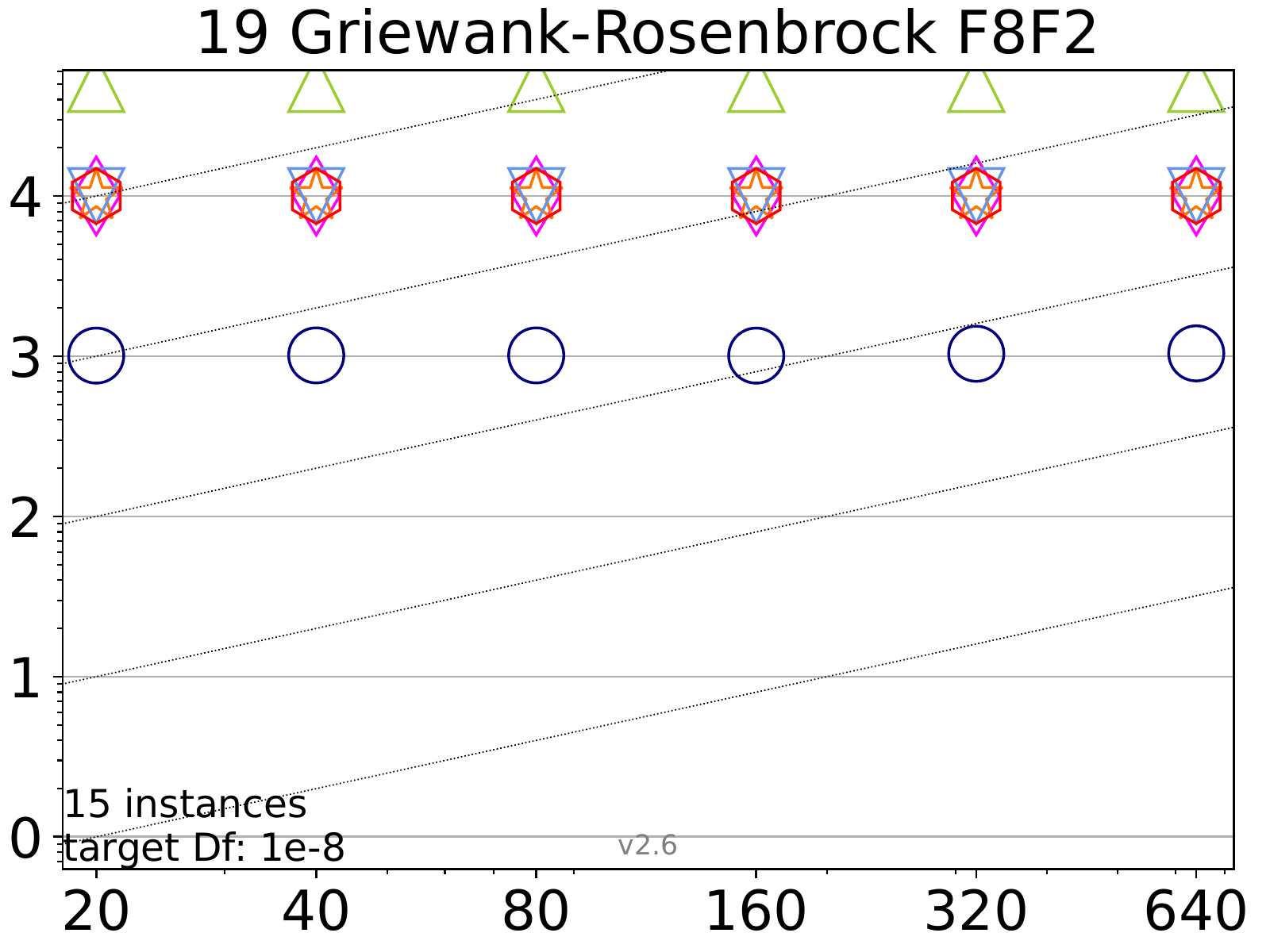}&
\includegraphics[width=0.24\textwidth]{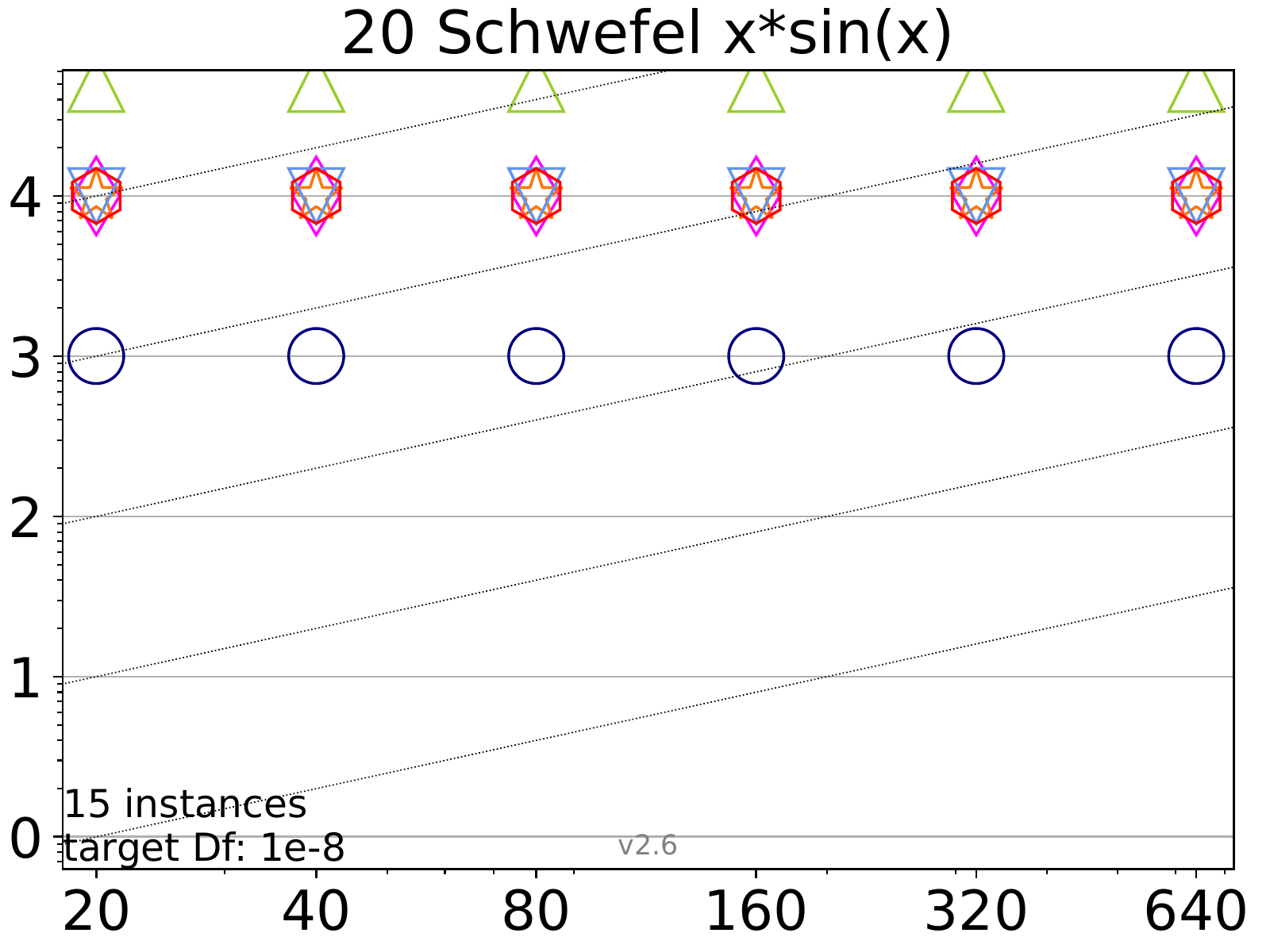}\\[-0.25em]
\includegraphics[width=0.24\textwidth]{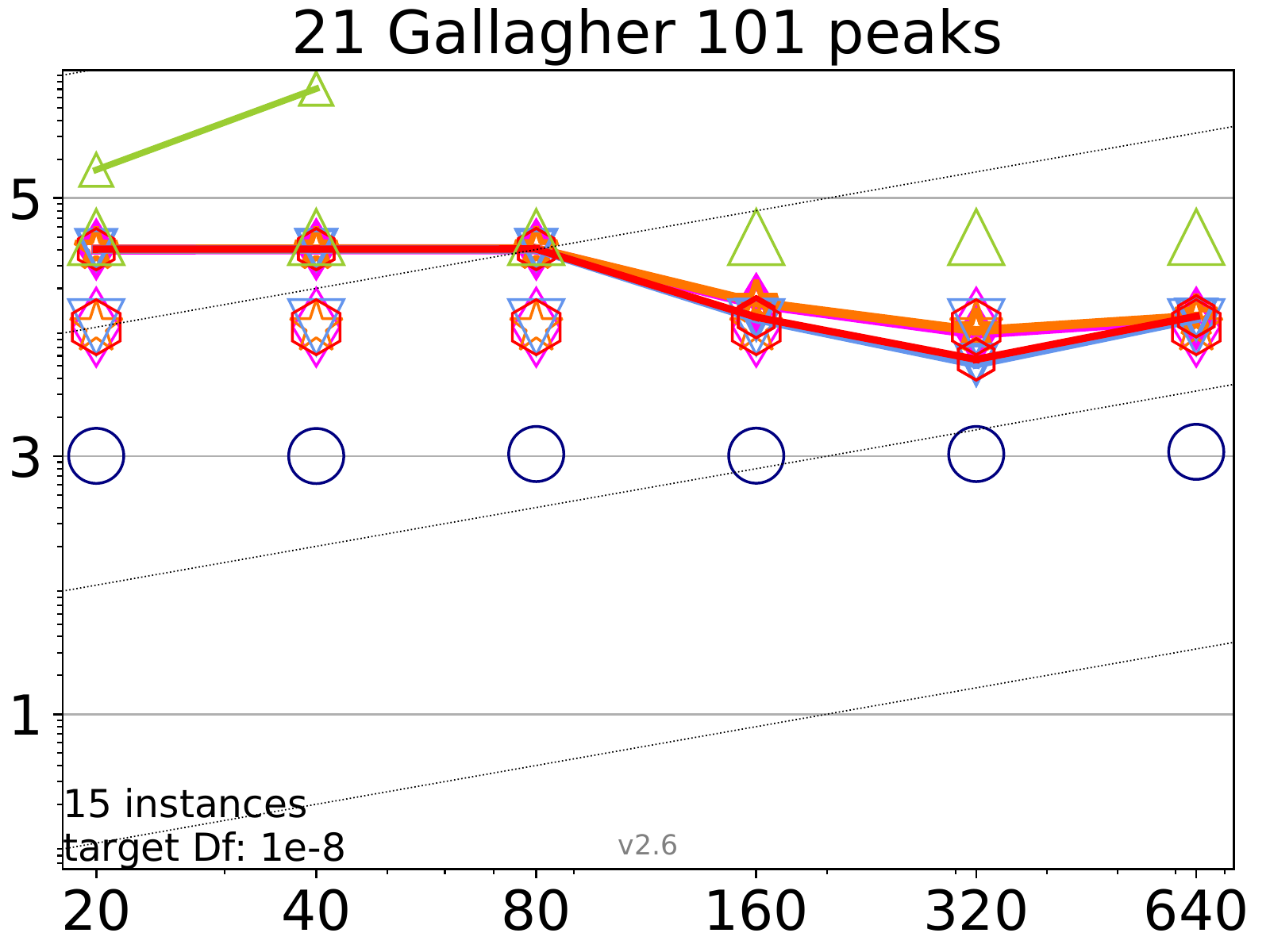}&
\includegraphics[width=0.24\textwidth]{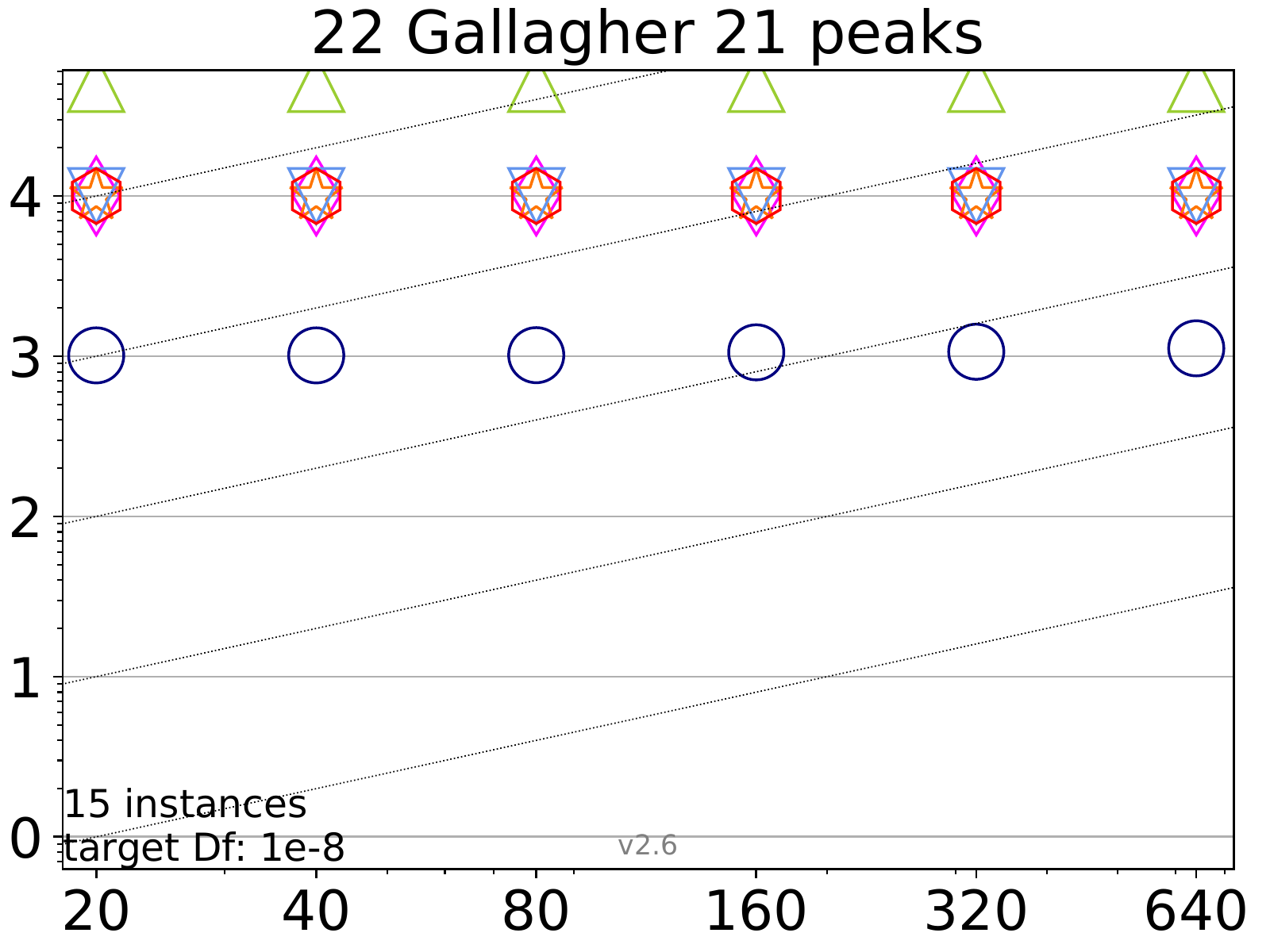}&
\includegraphics[width=0.24\textwidth]{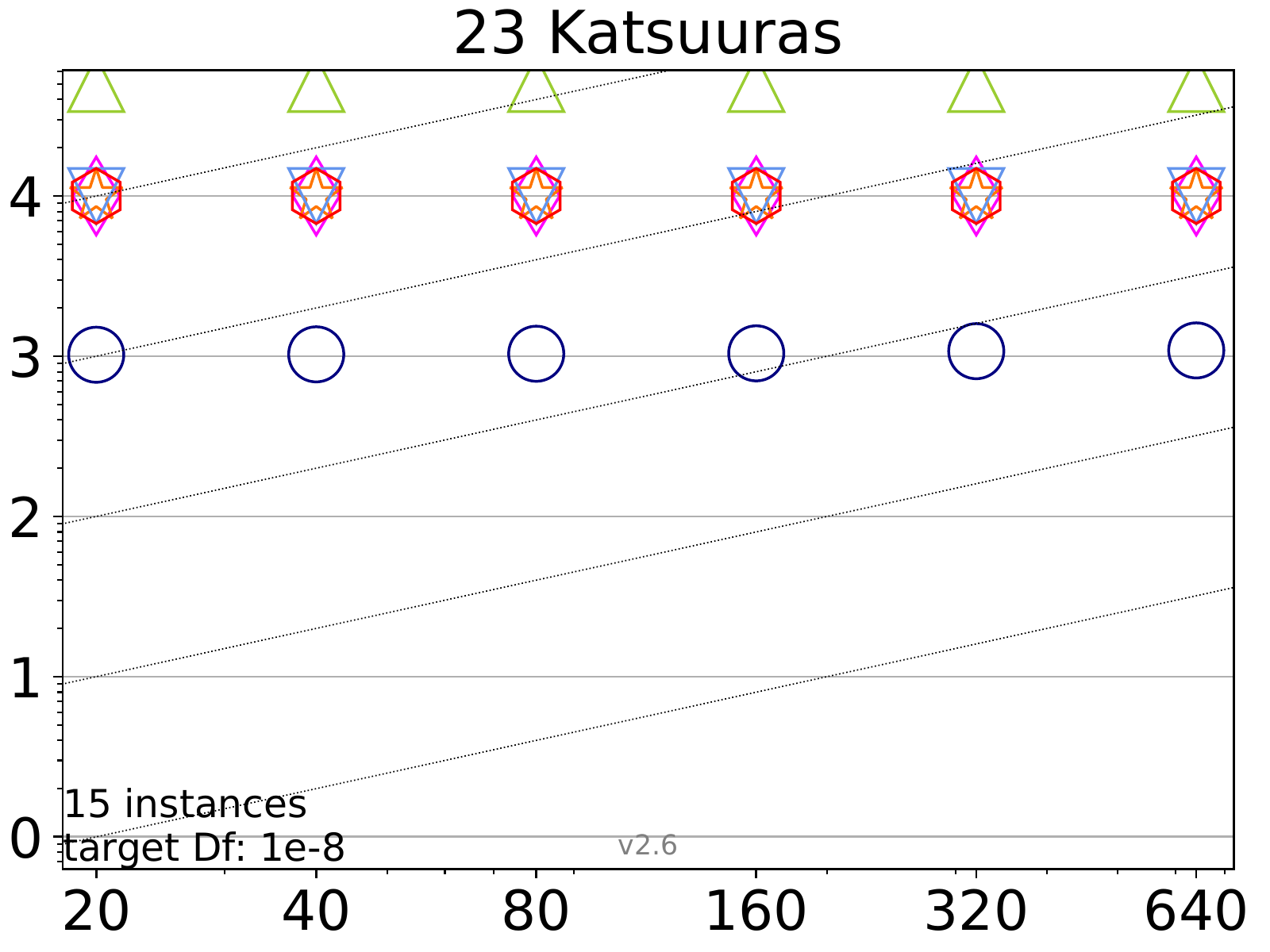}&
\includegraphics[width=0.24\textwidth]{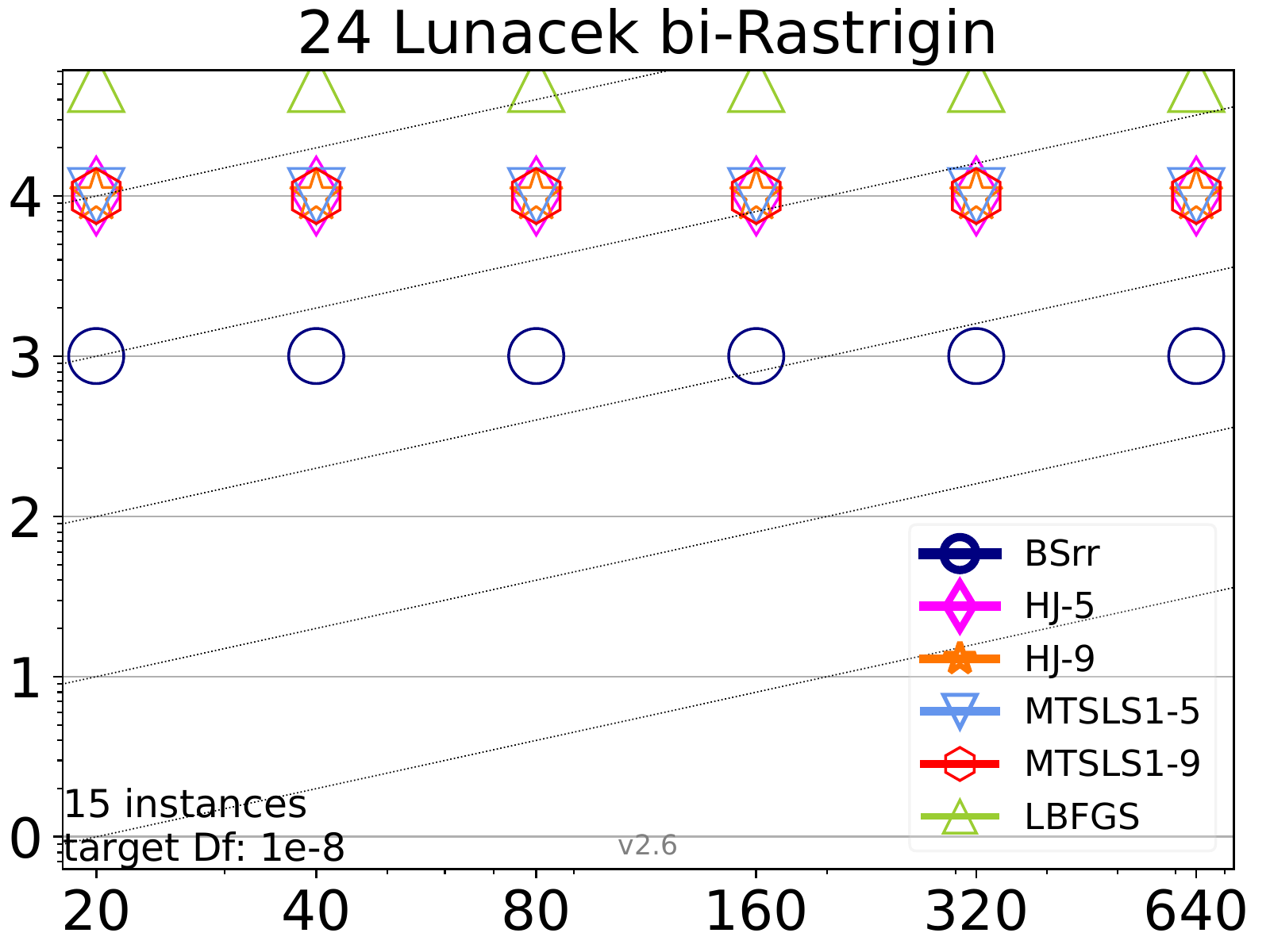}
\end{tabular}
\vspace*{-0.2cm}
\caption[Expected running time (\ERT) divided by dimension
versus dimension in log-log presentation]{
\label{fig:scaling}
\bbobppfigslegend{$f_1$ and $f_{24}$}. 
}
\end{figure*}

\begin{figure*}[htp]
\begin{tabular}{@{}l@{}l@{}l@{}}
    separable fcts & moderate fcts & ill-conditioned fcts\\
 \includegraphics[height=0.19\textheight]{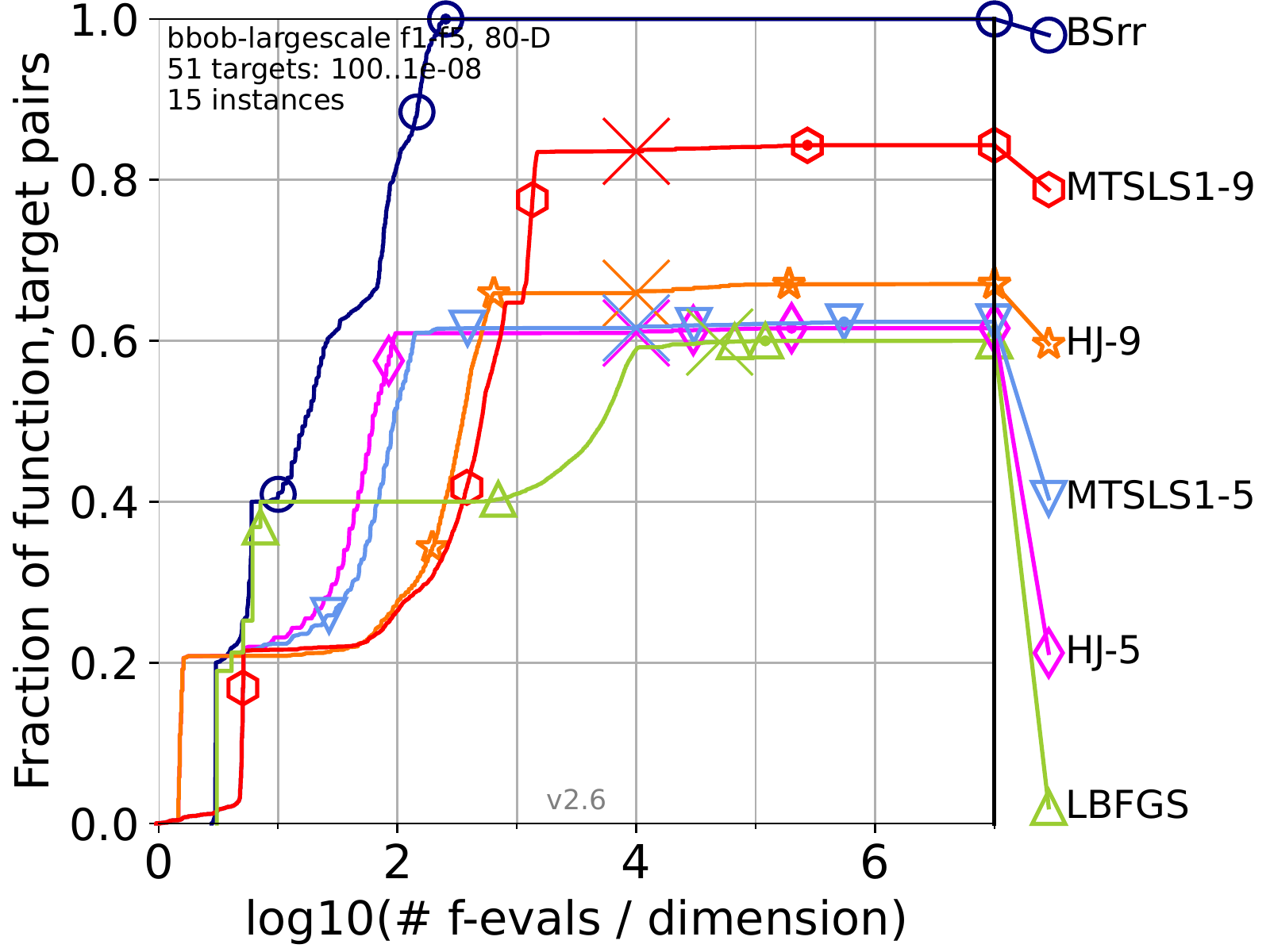} &
 \includegraphics[height=0.19\textheight]{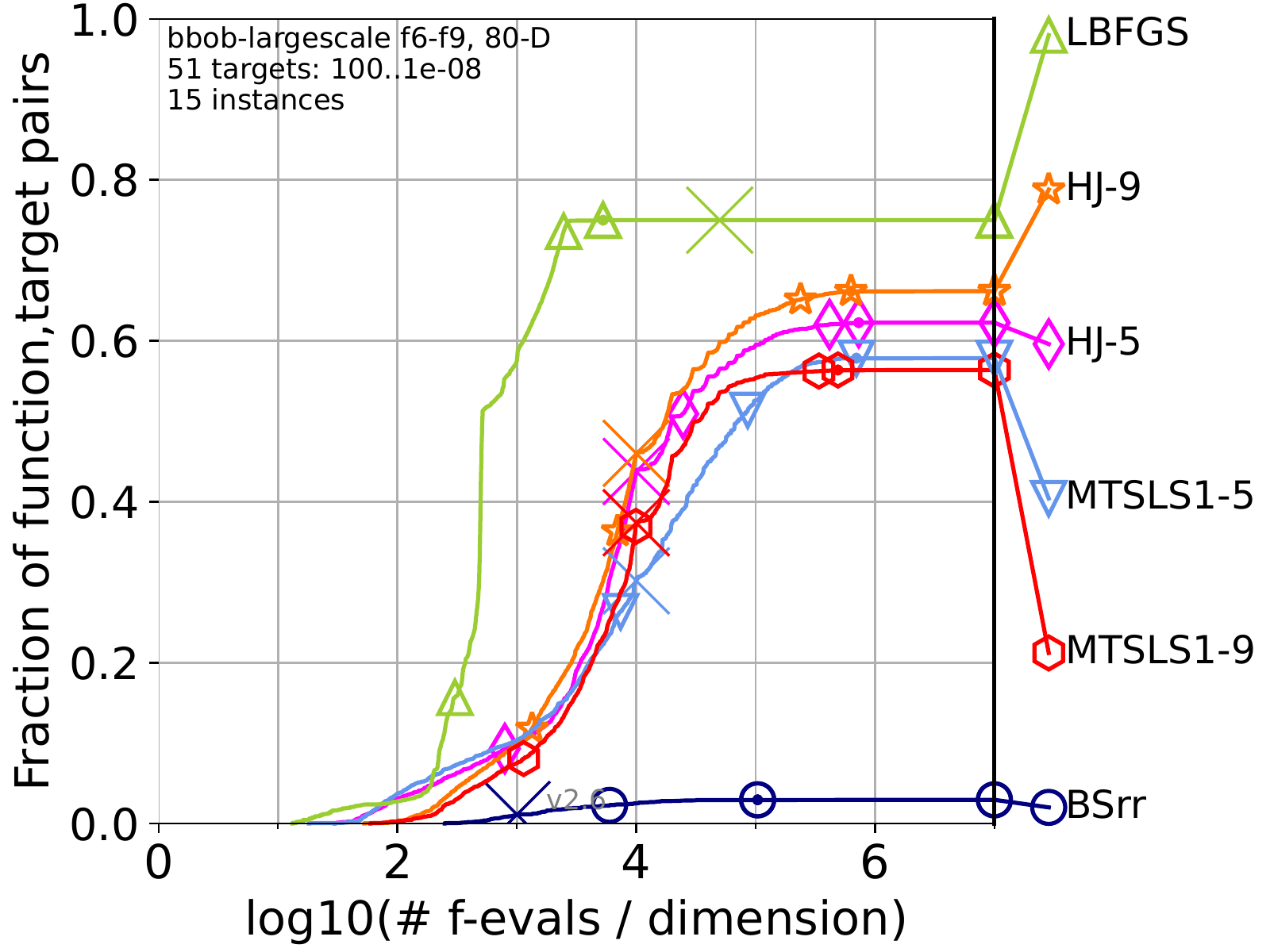} & 
 \includegraphics[height=0.19\textheight]{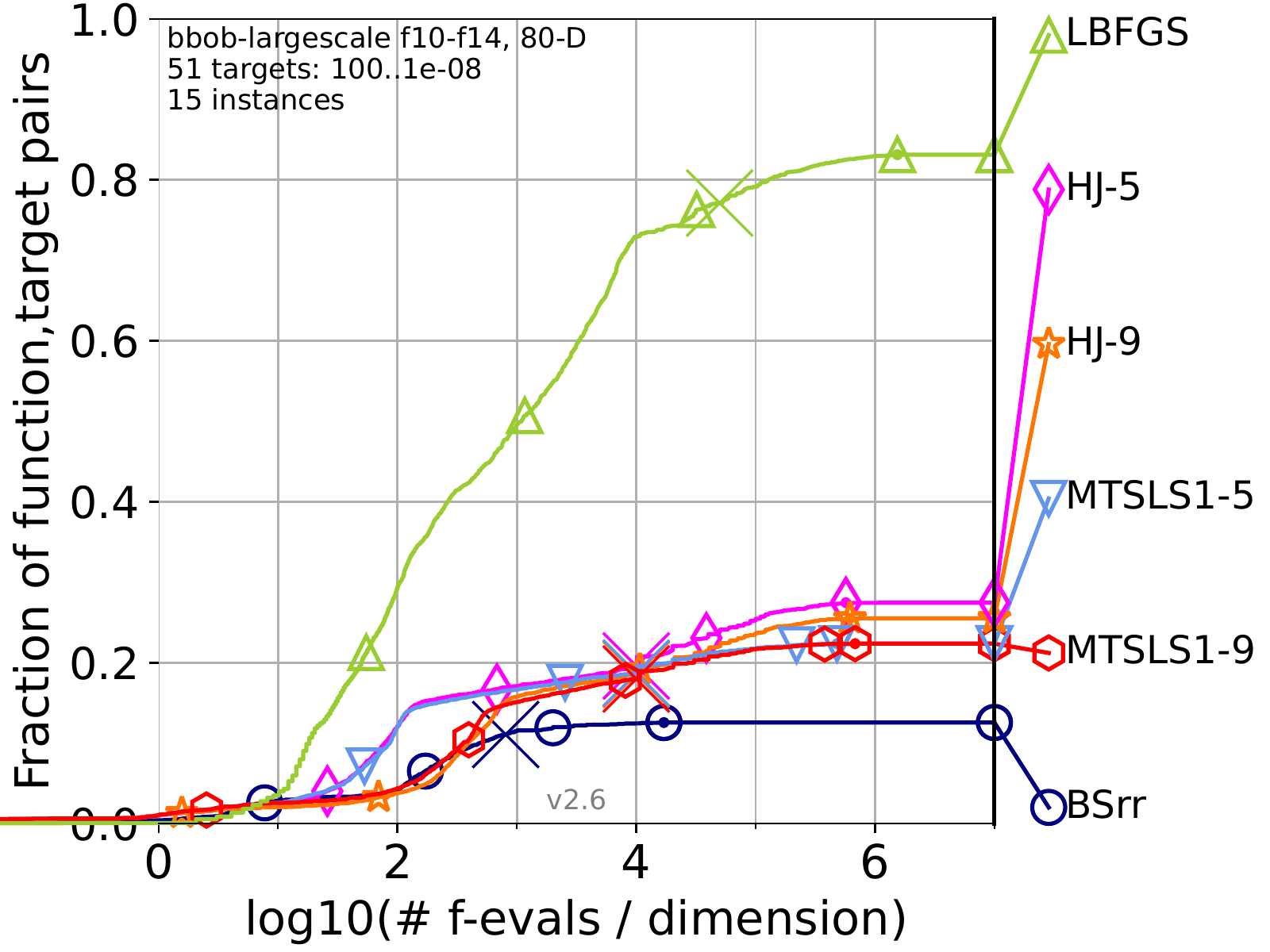}\\
 multi-modal fcts & weakly structured multi-modal fcts & all functions\\
 \includegraphics[height=0.19\textheight]{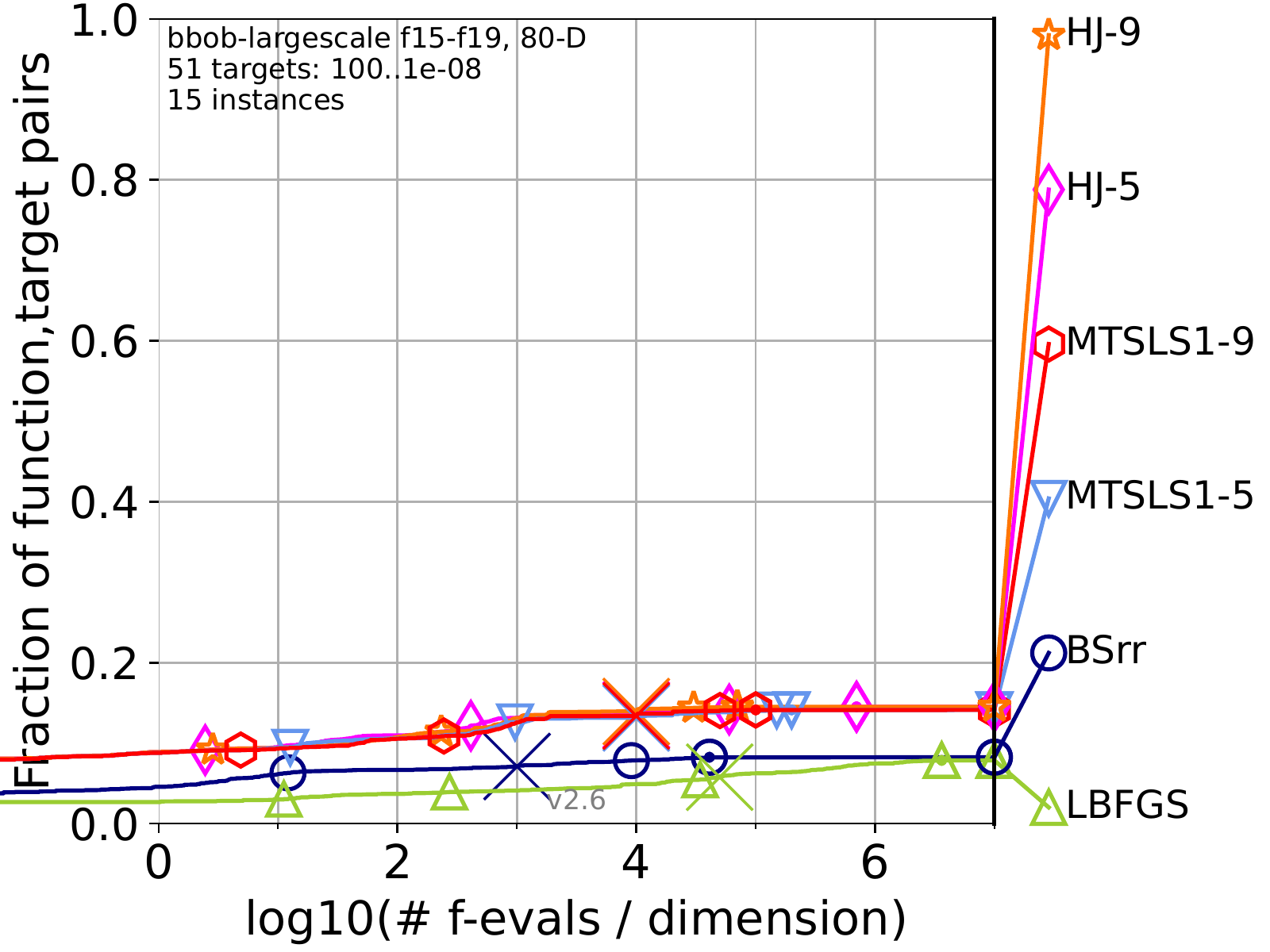} &
 \includegraphics[height=0.19\textheight]{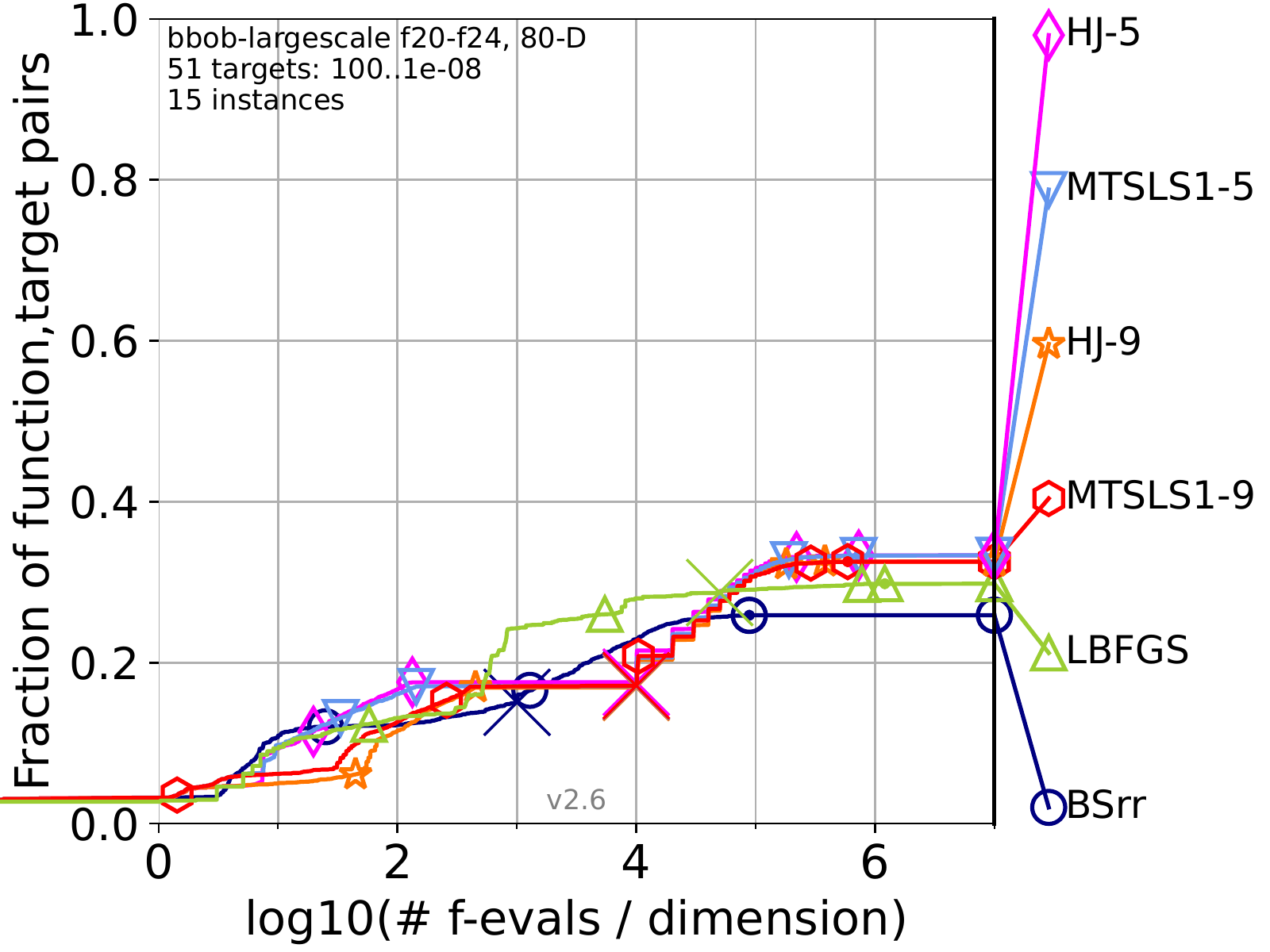} & 
 \includegraphics[height=0.19\textheight]{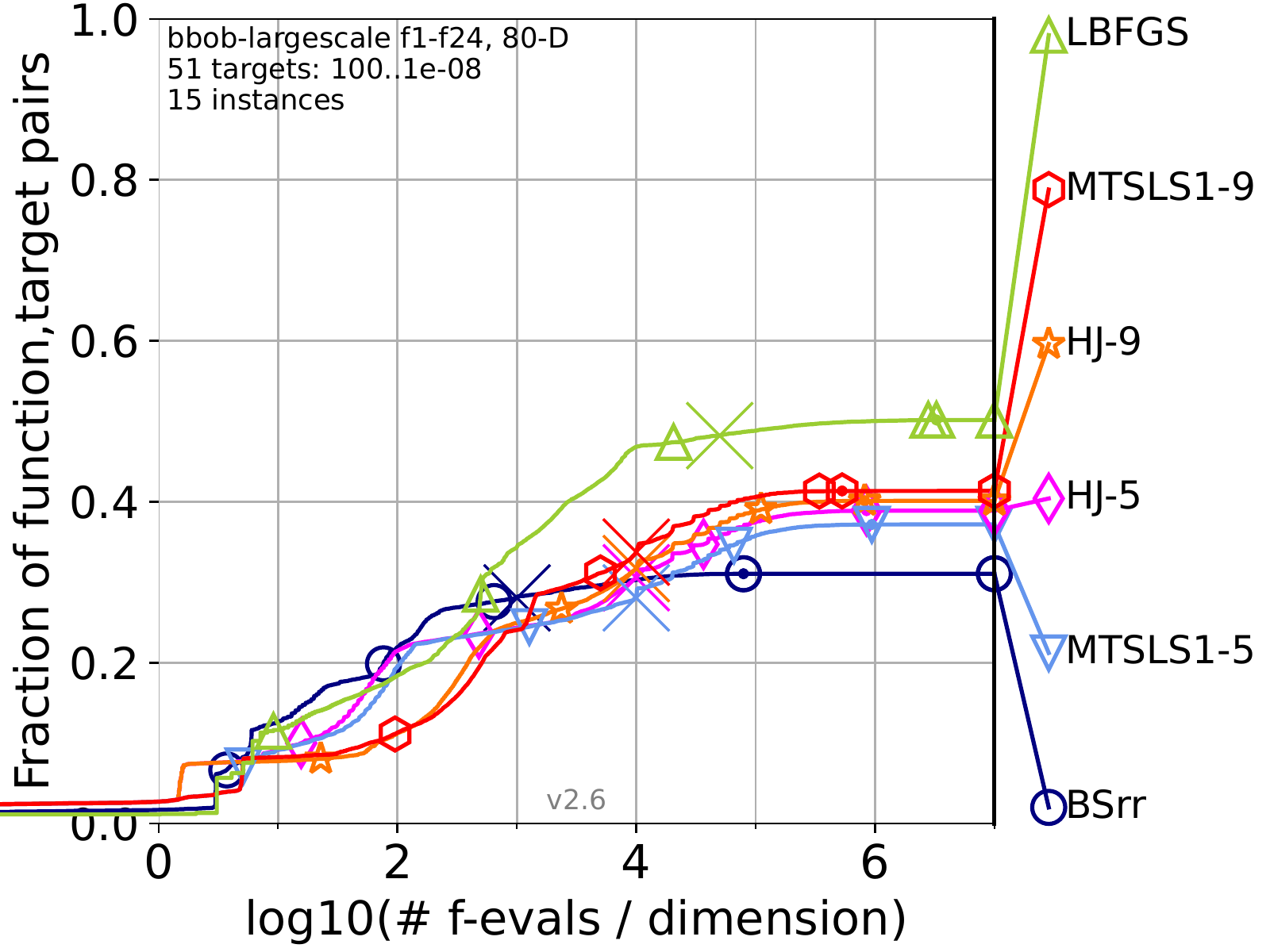} 
 \end{tabular}
\caption{
\label{fig:ECDFs80D}
\bbobECDFslegend{80}
}
\begin{tabular}{@{}l@{}l@{}l@{}}
    separable fcts & moderate fcts & ill-conditioned fcts\\
 \includegraphics[height=0.19\textheight]{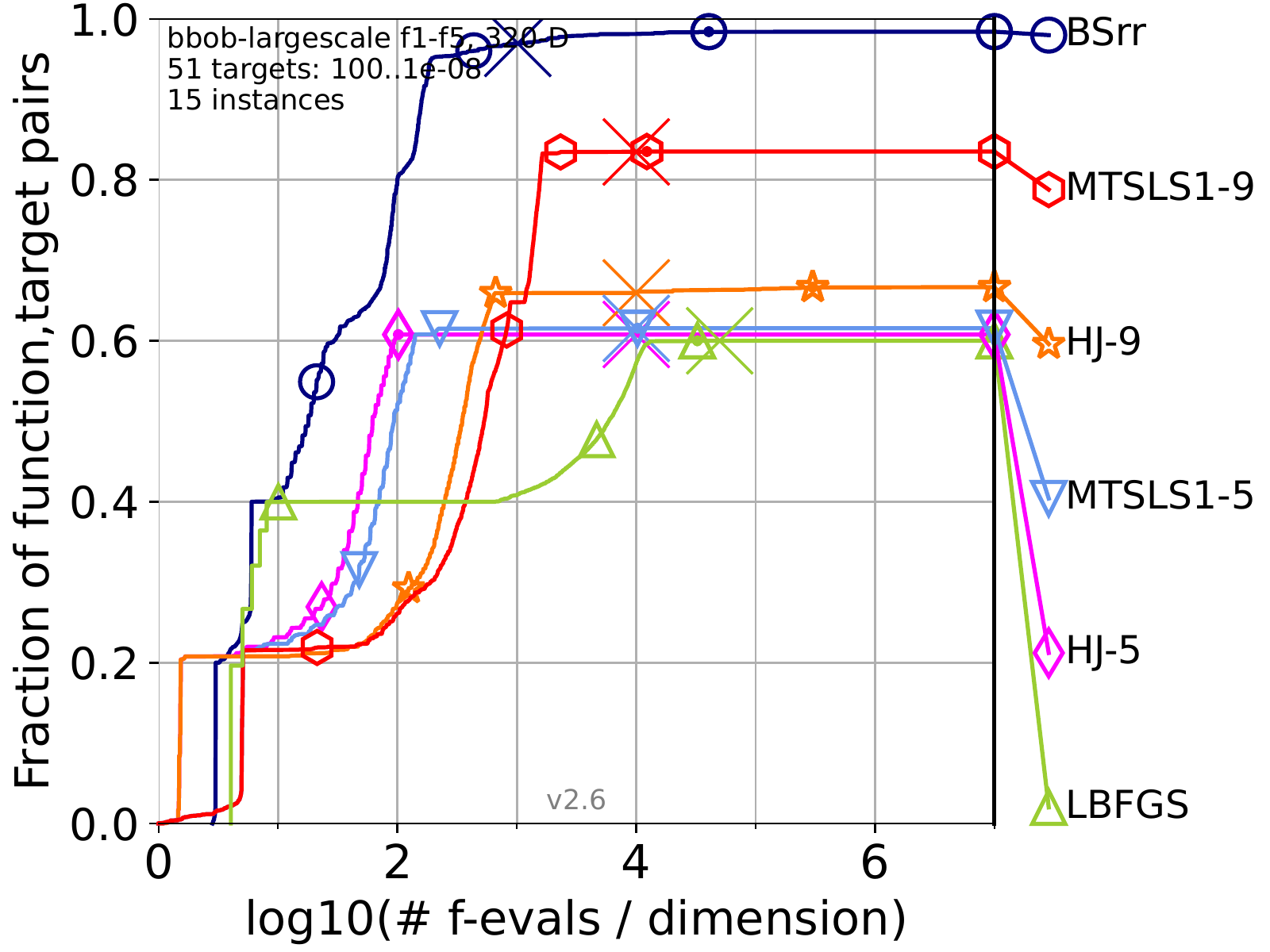} &
 \includegraphics[height=0.19\textheight]{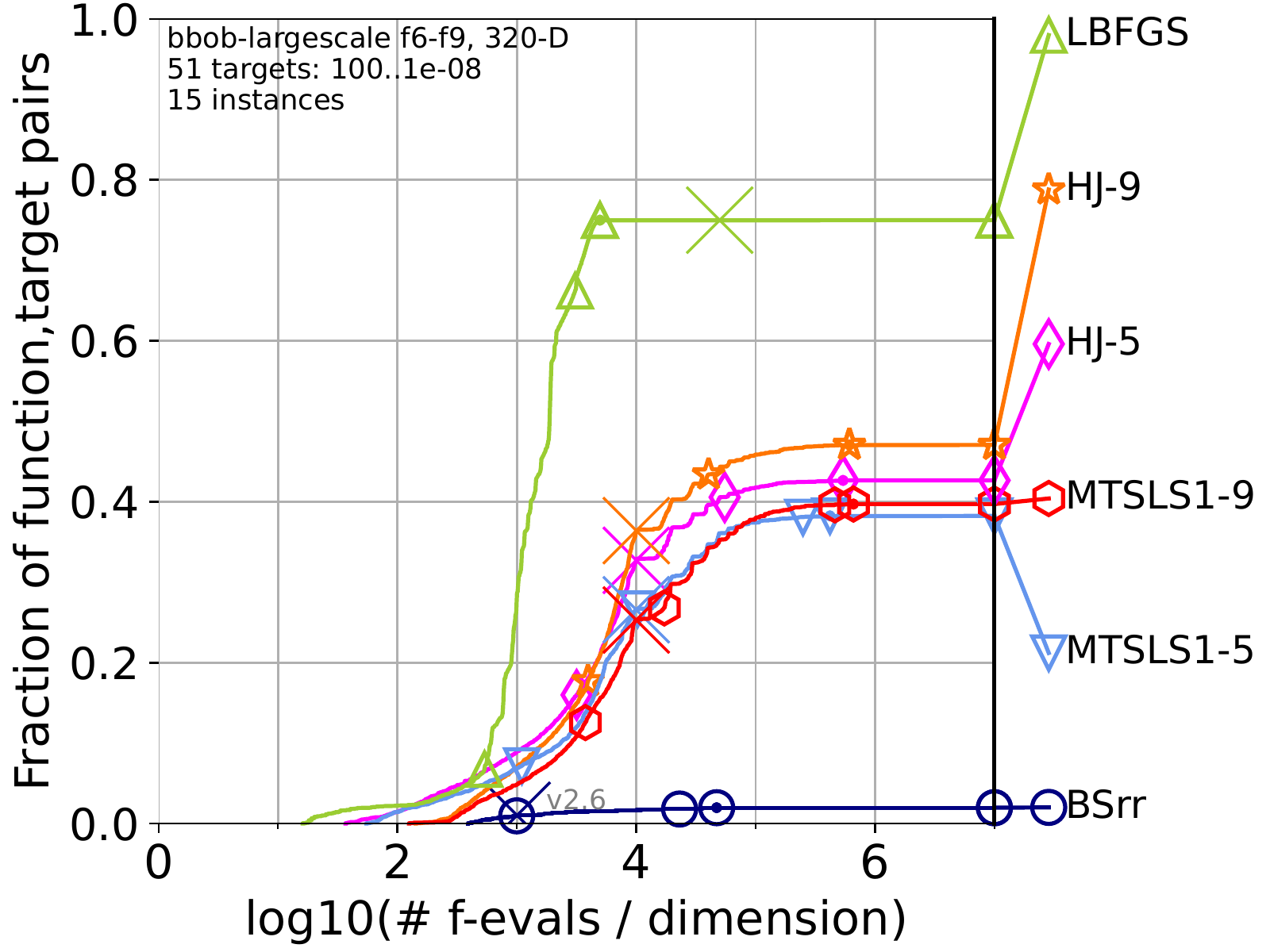} &
 \includegraphics[height=0.19\textheight]{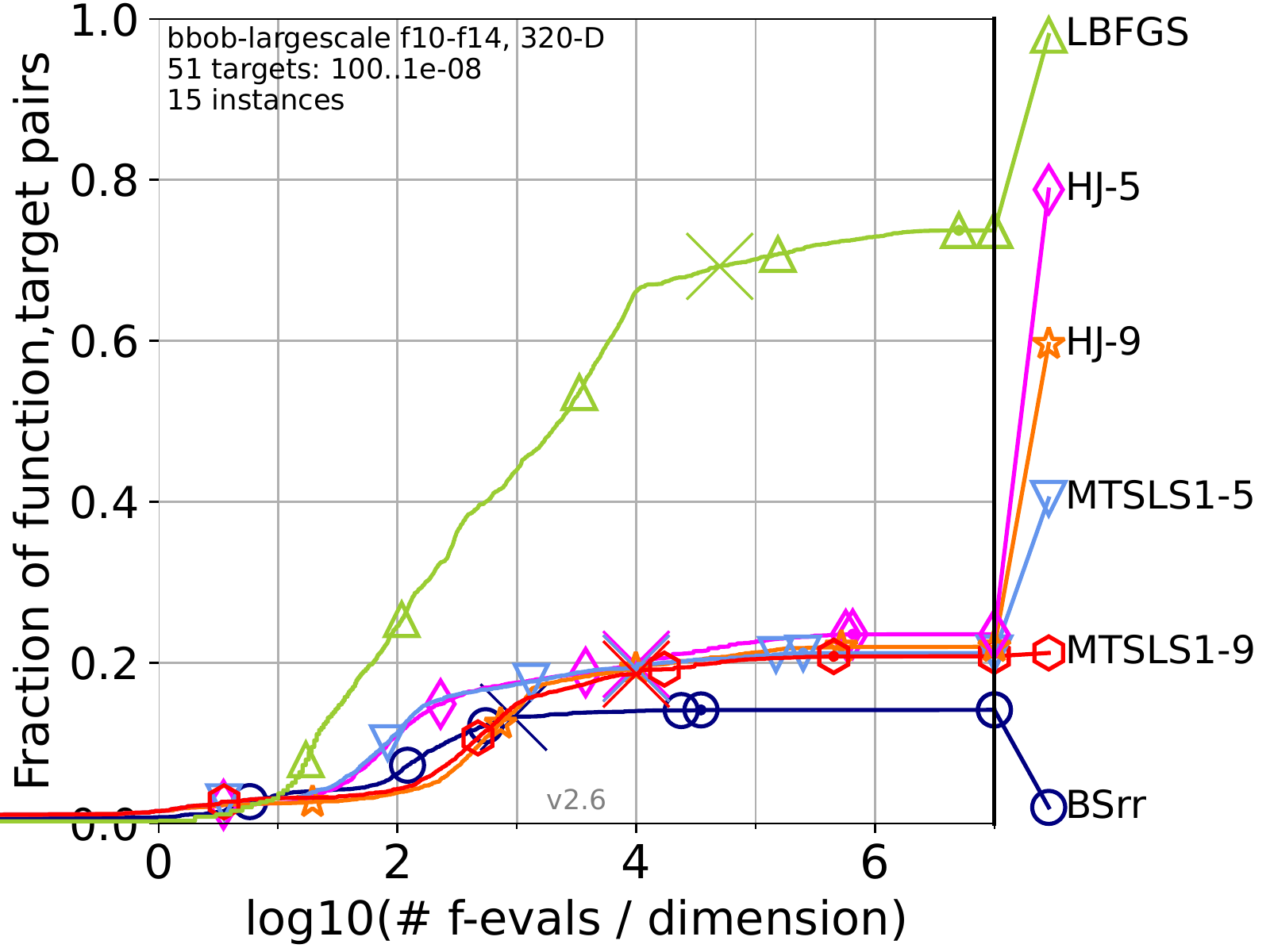}\\
 multi-modal fcts & weakly structured multi-modal fcts & all functions\\
 \includegraphics[height=0.19\textheight]{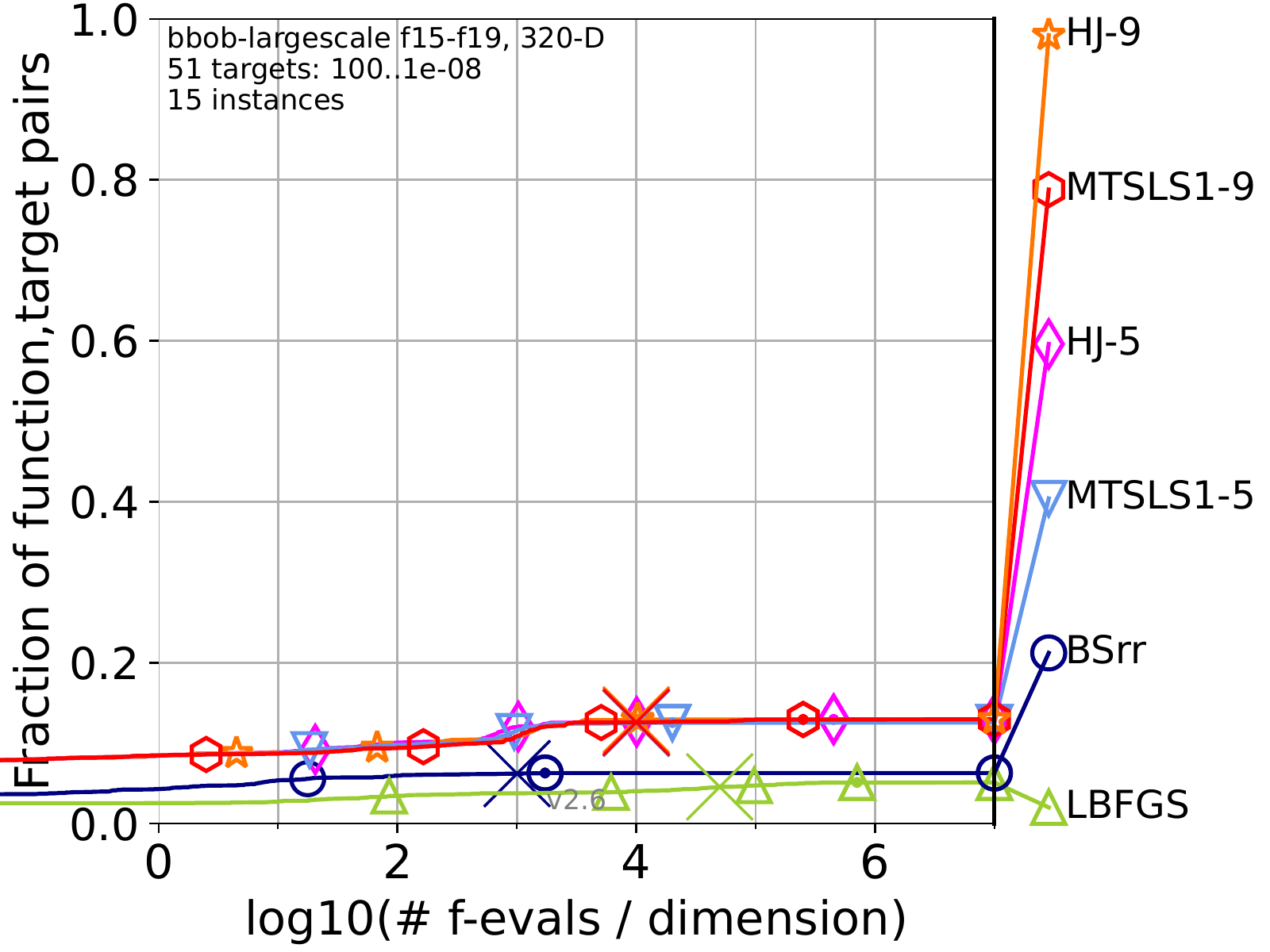} &
 \includegraphics[height=0.19\textheight]{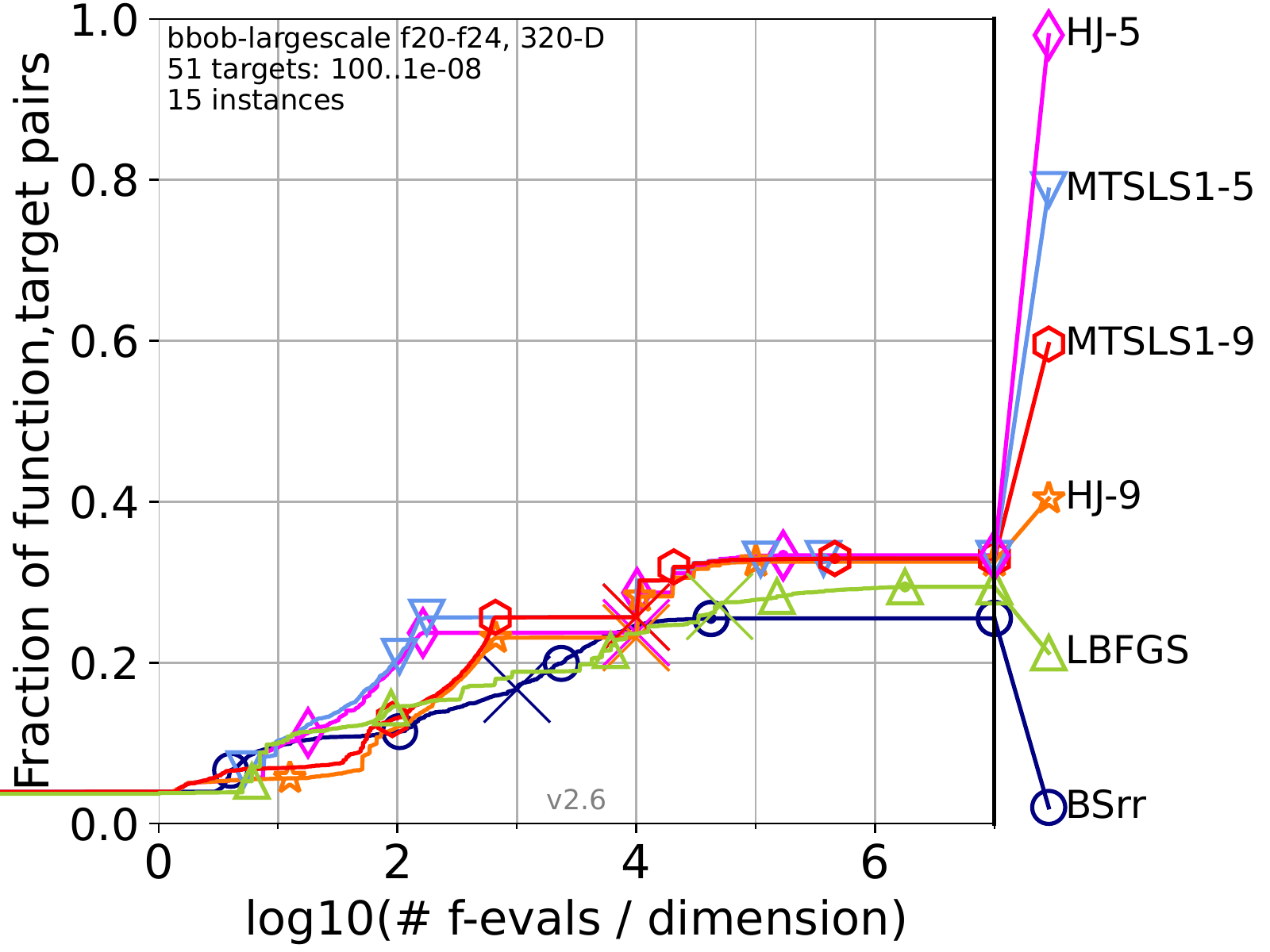} & 
 \includegraphics[height=0.19\textheight]{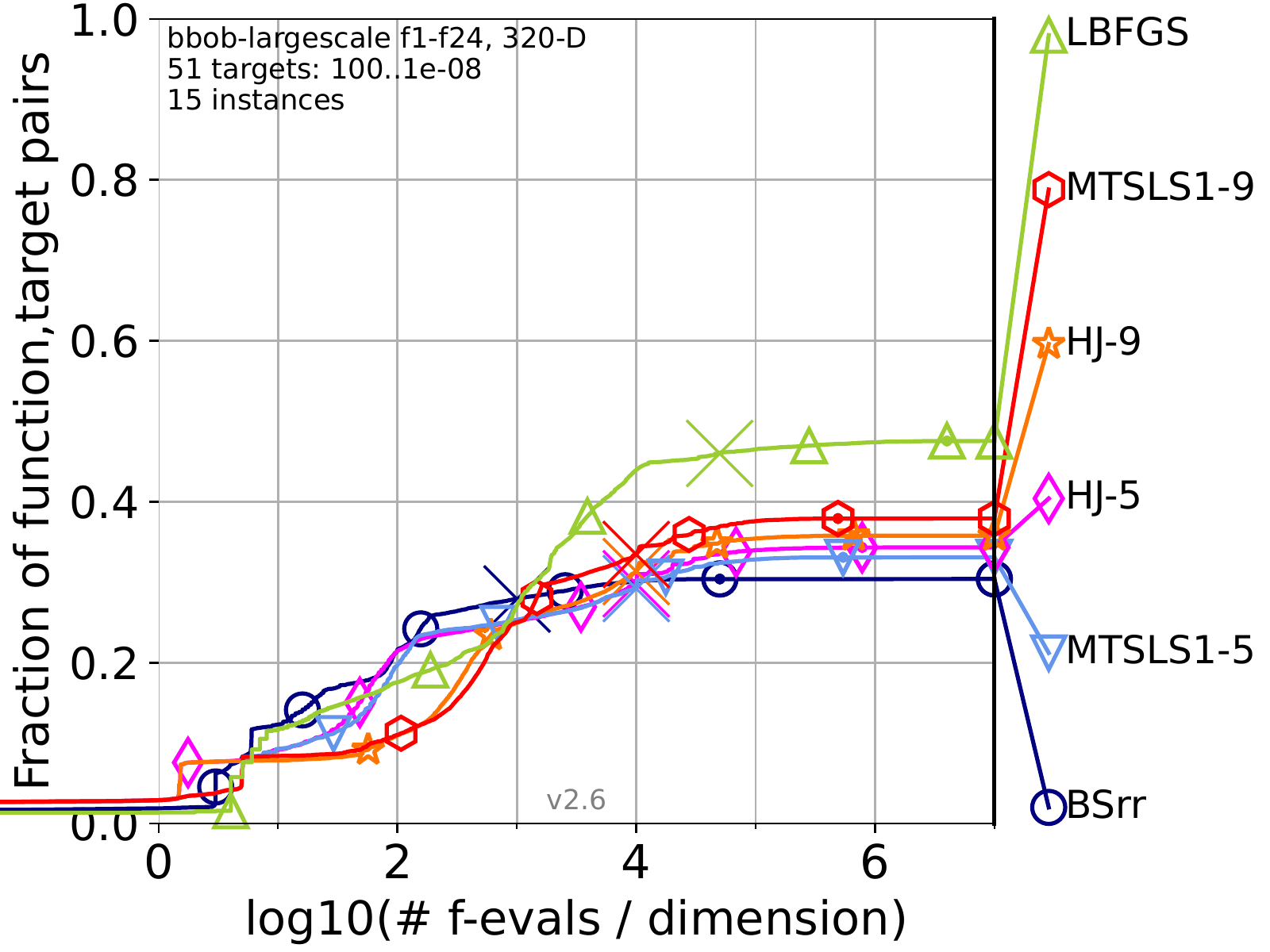} 
 \end{tabular}
\caption{
\label{fig:ECDFs320D}
\bbobECDFslegend{320}
}
\end{figure*}

\begin{figure*}
  \newcommand{\widthvar}{0.2}    
  \centering
\begin{tabular}{@{}l@{}l@{}l@{}l@{}l@{}}
\includegraphics[width=\widthvar\textwidth]{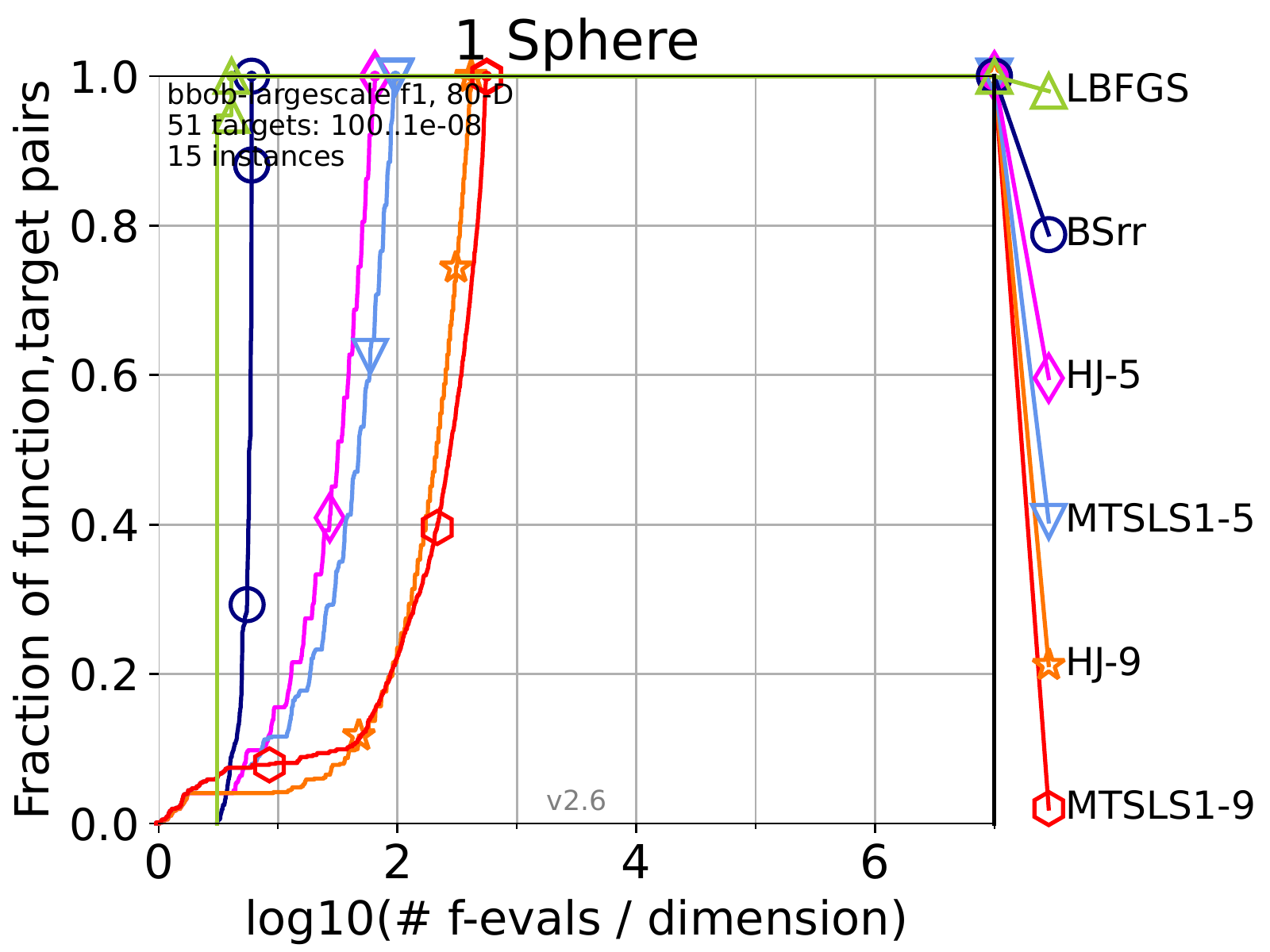}&
\includegraphics[width=\widthvar\textwidth]{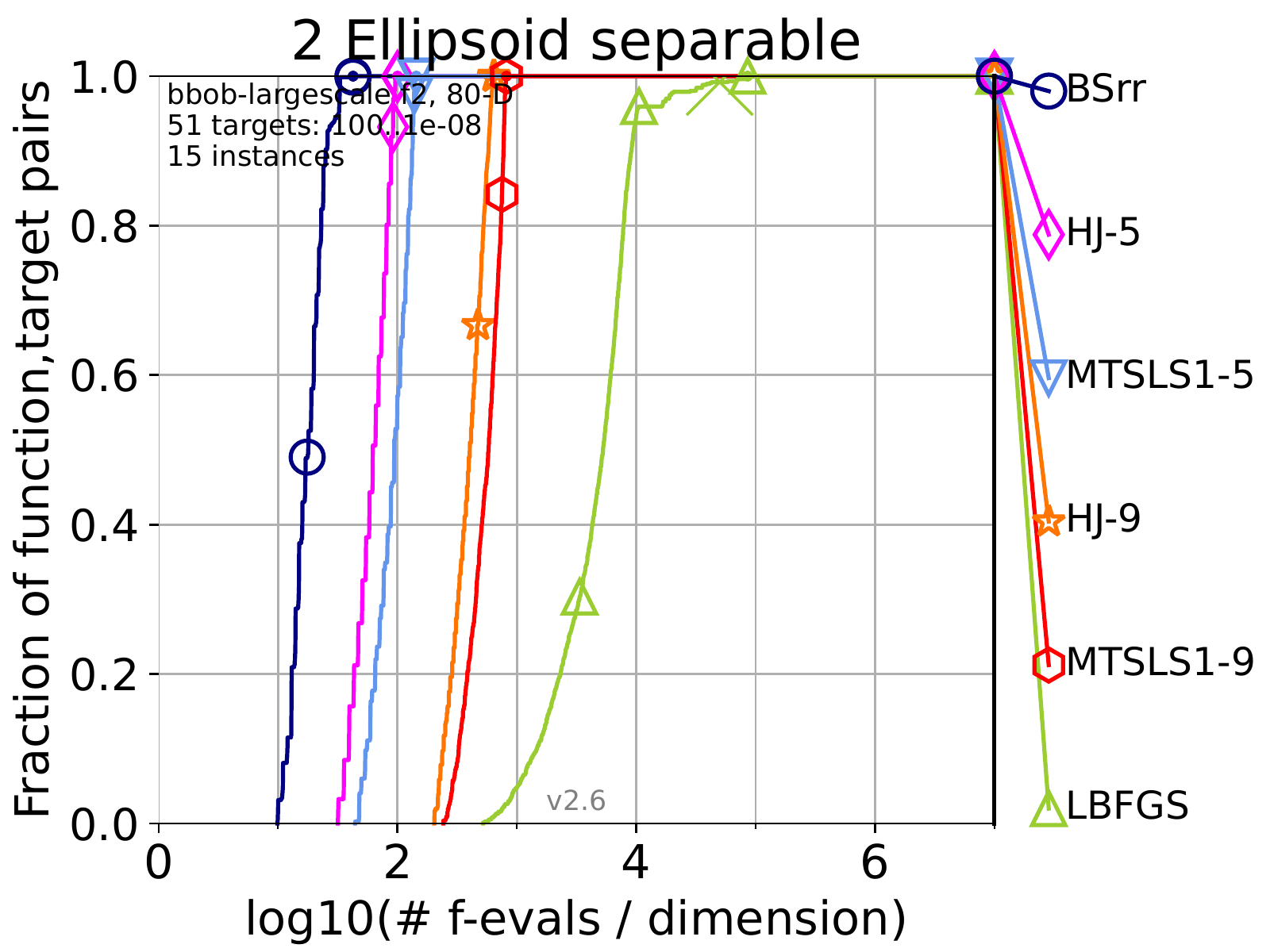}&
\includegraphics[width=\widthvar\textwidth]{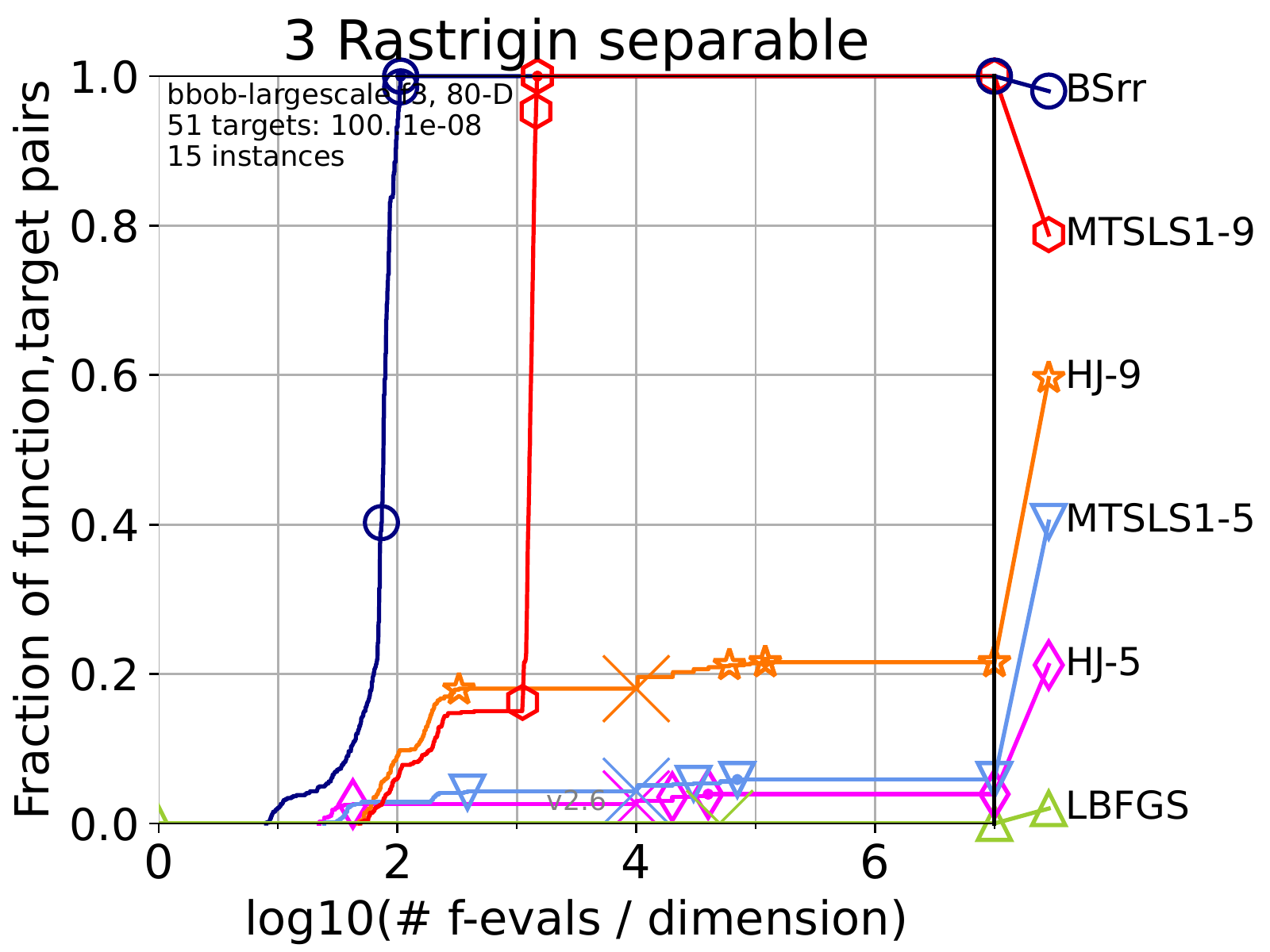}&
\includegraphics[width=\widthvar\textwidth]{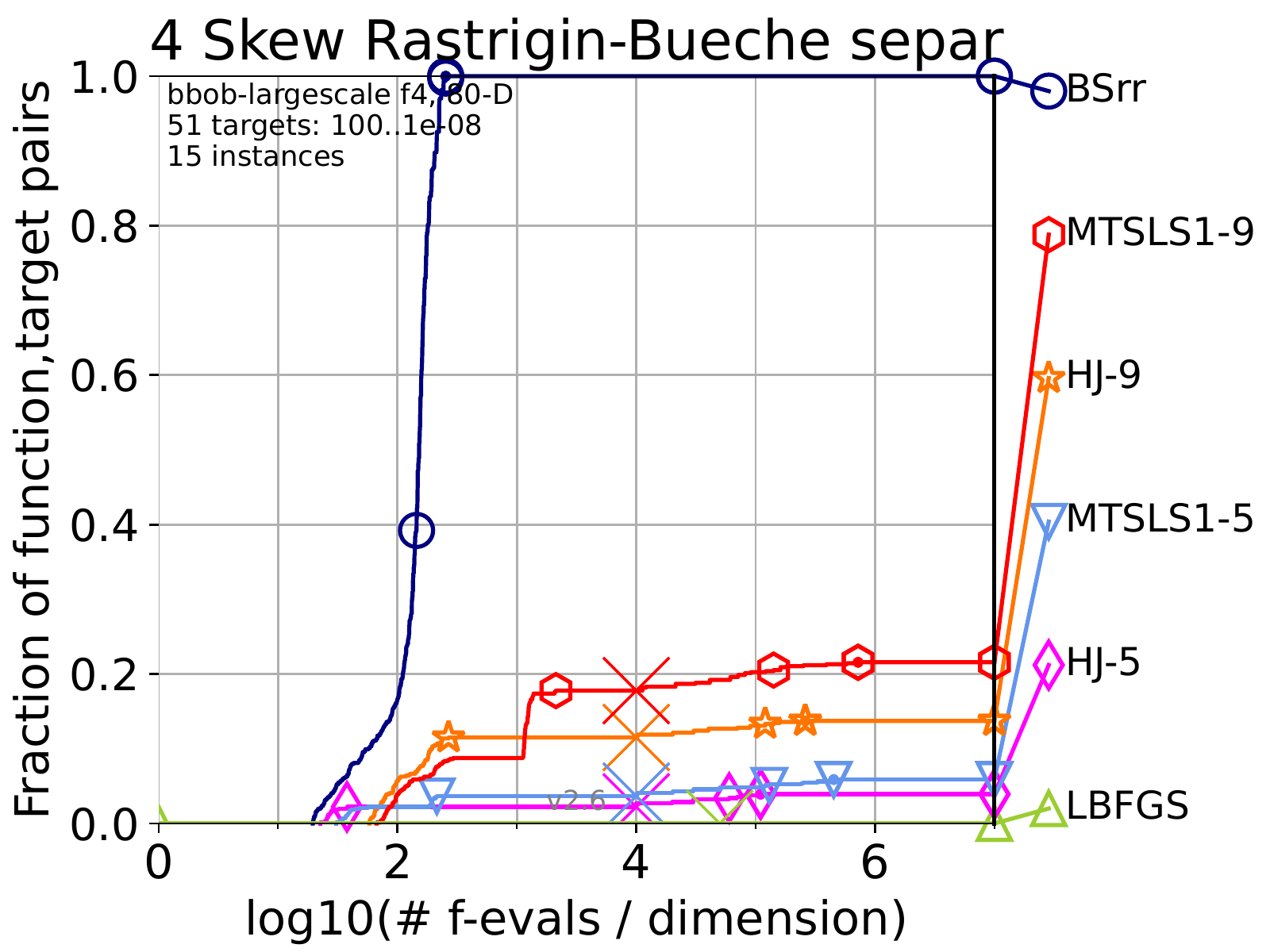}\\
\includegraphics[width=\widthvar\textwidth]{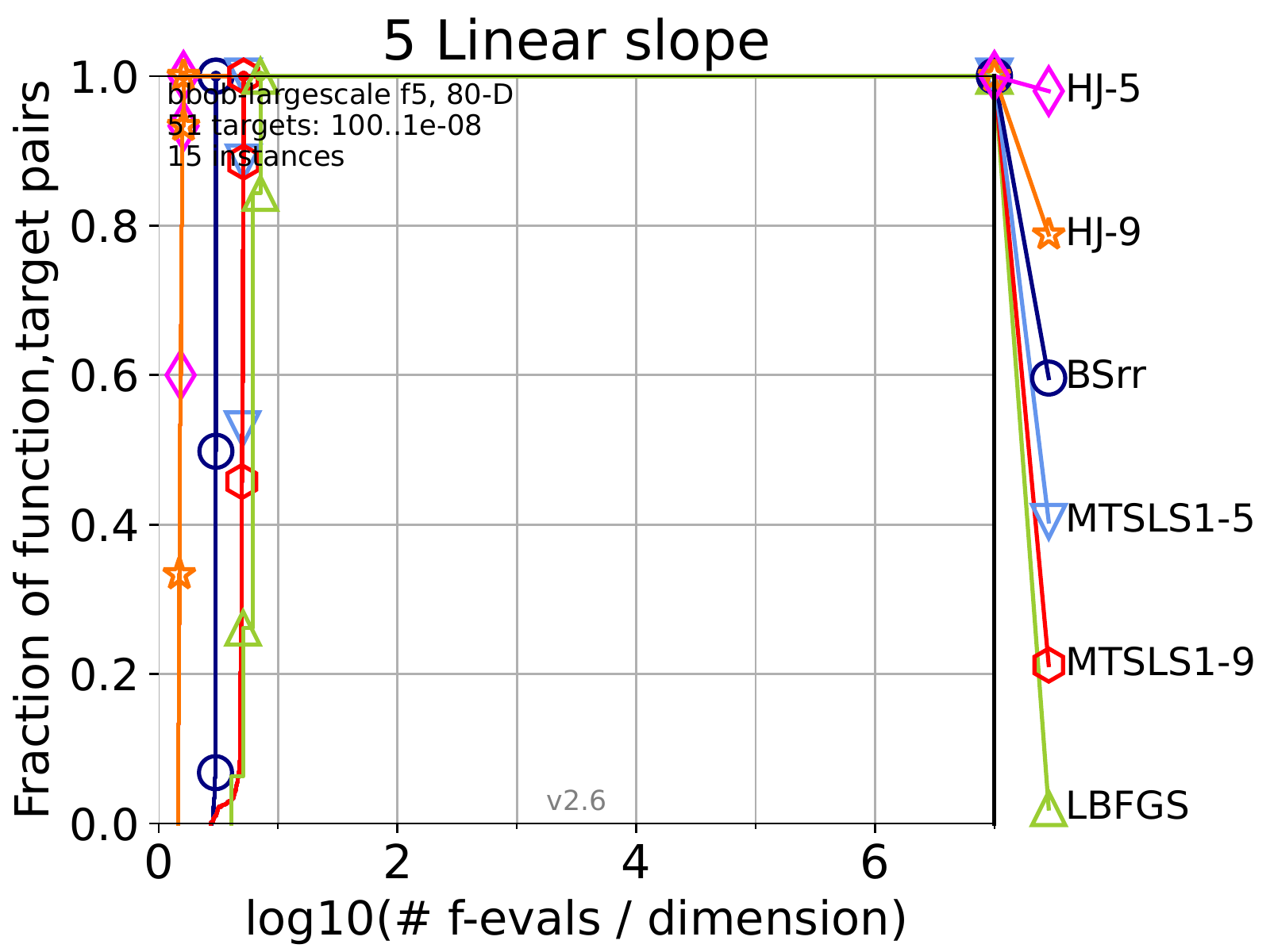}&
\includegraphics[width=\widthvar\textwidth]{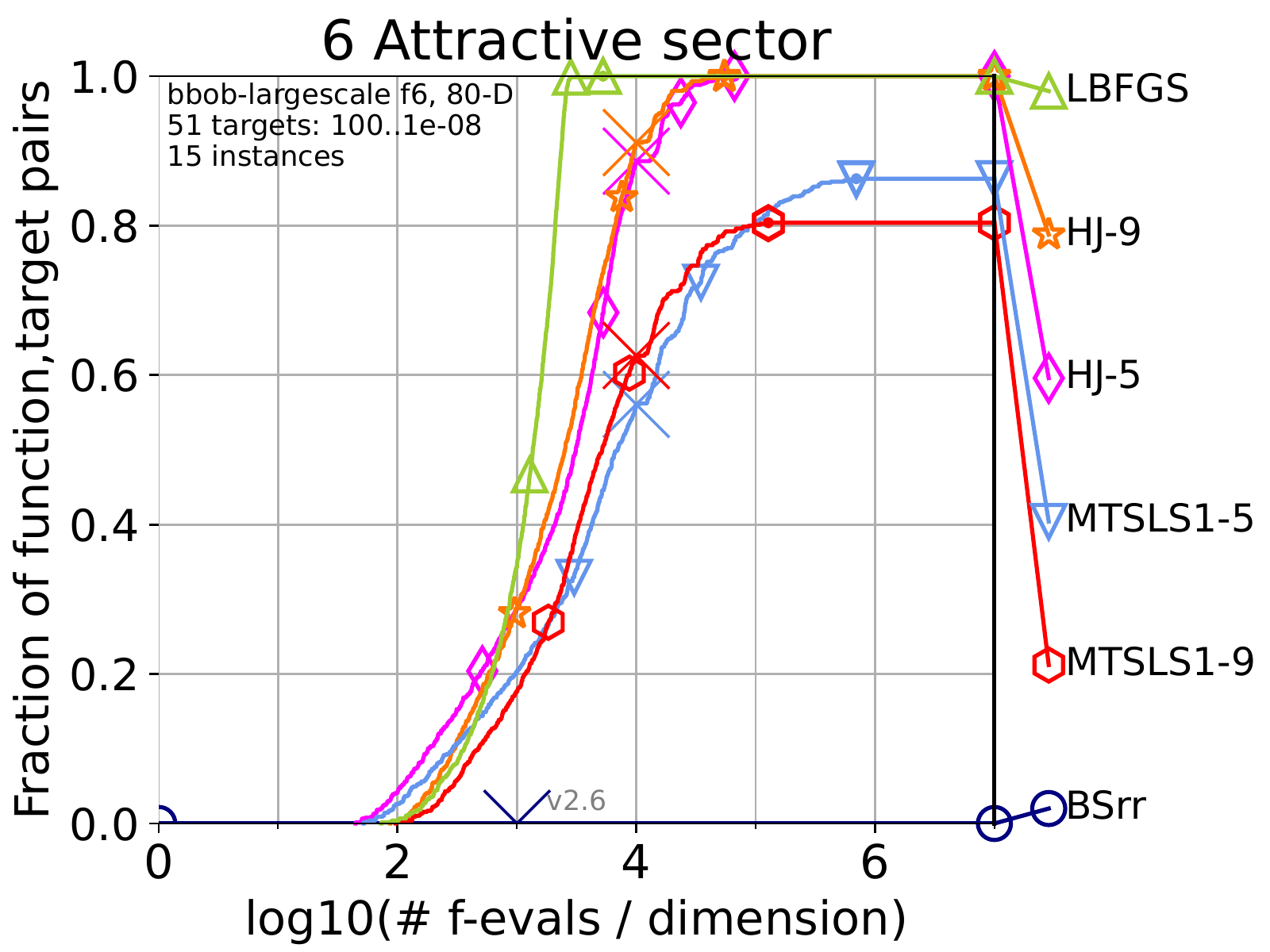}&
\includegraphics[width=\widthvar\textwidth]{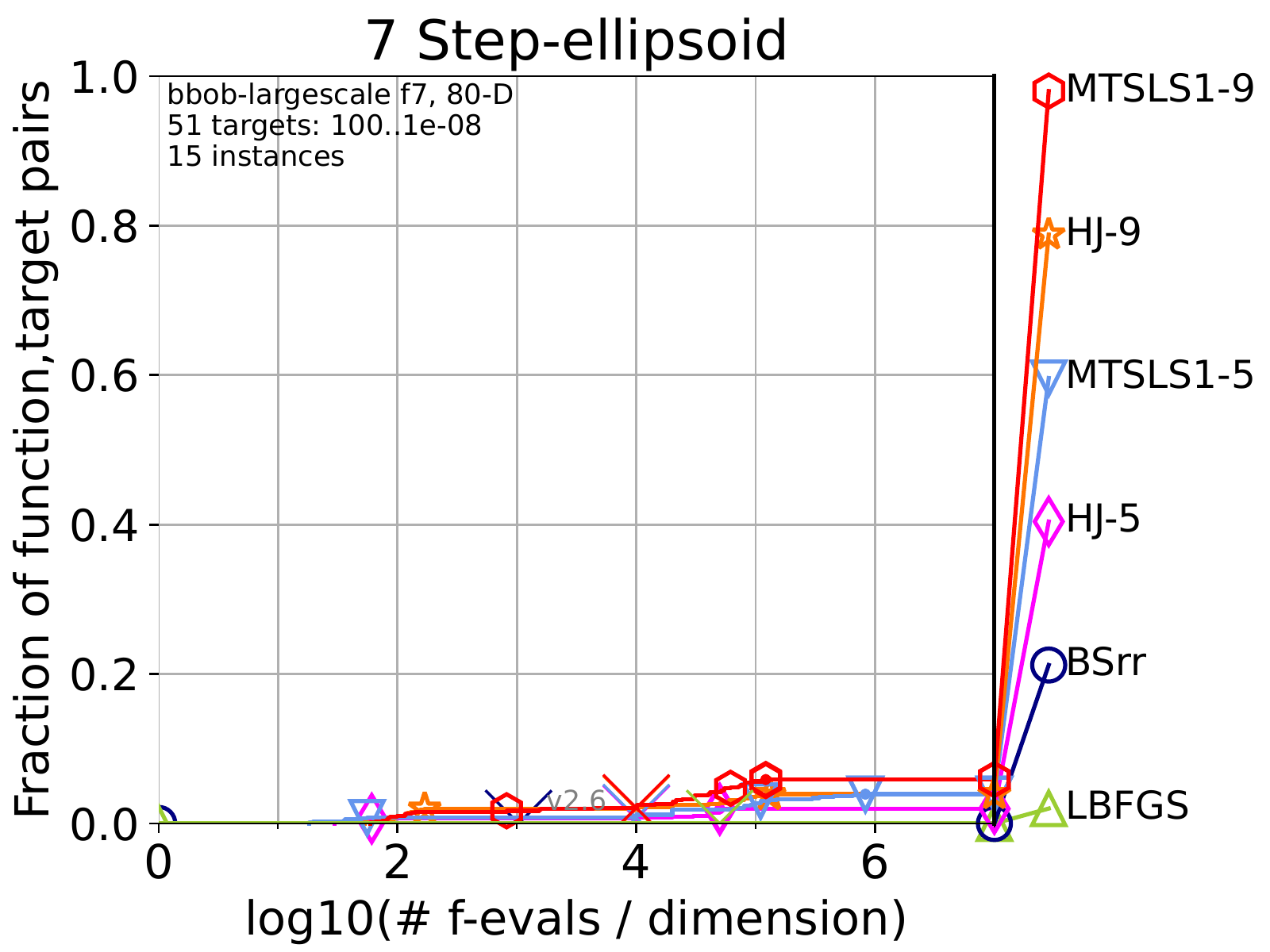}&
\includegraphics[width=\widthvar\textwidth]{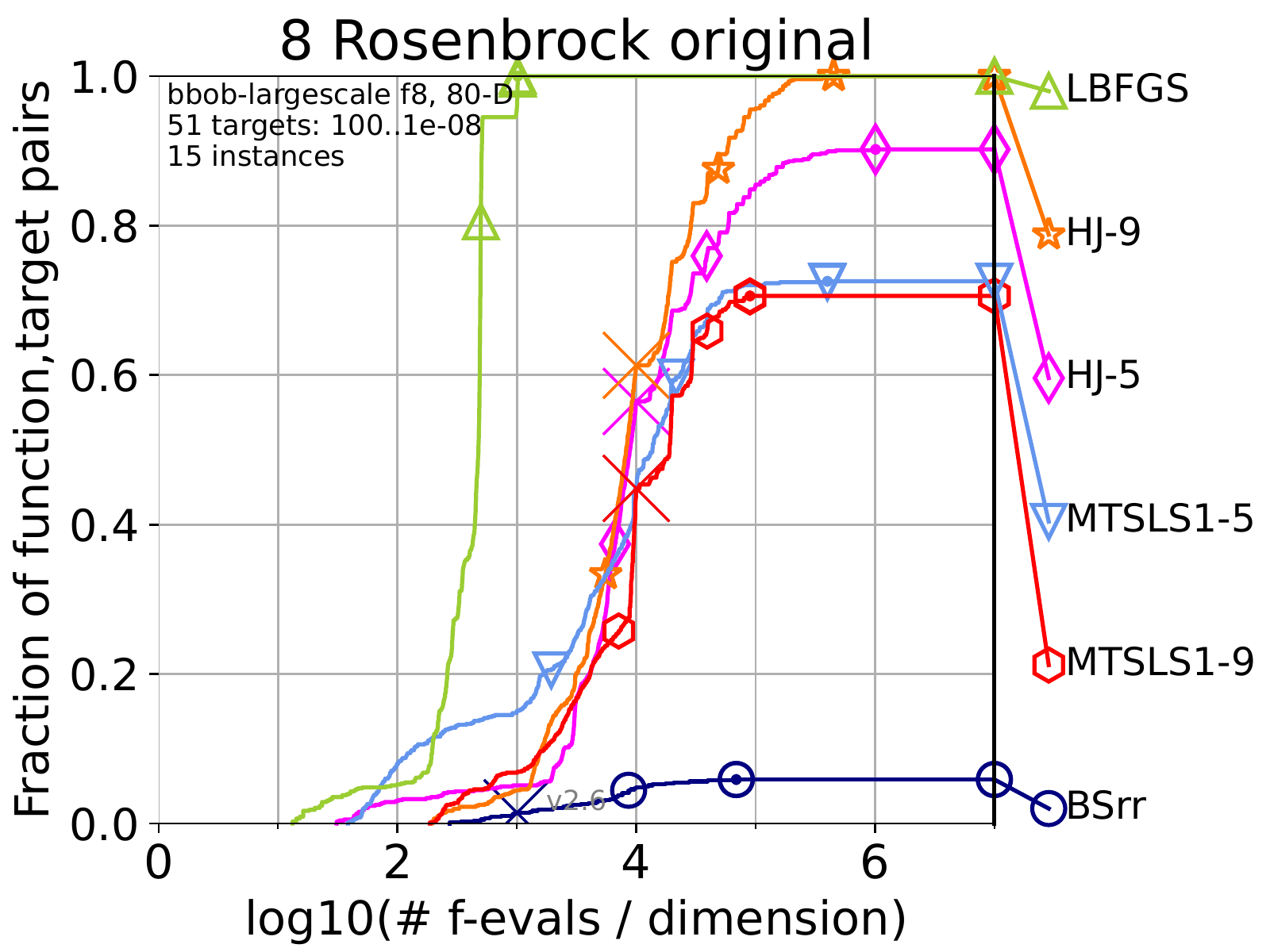}\\
\includegraphics[width=\widthvar\textwidth]{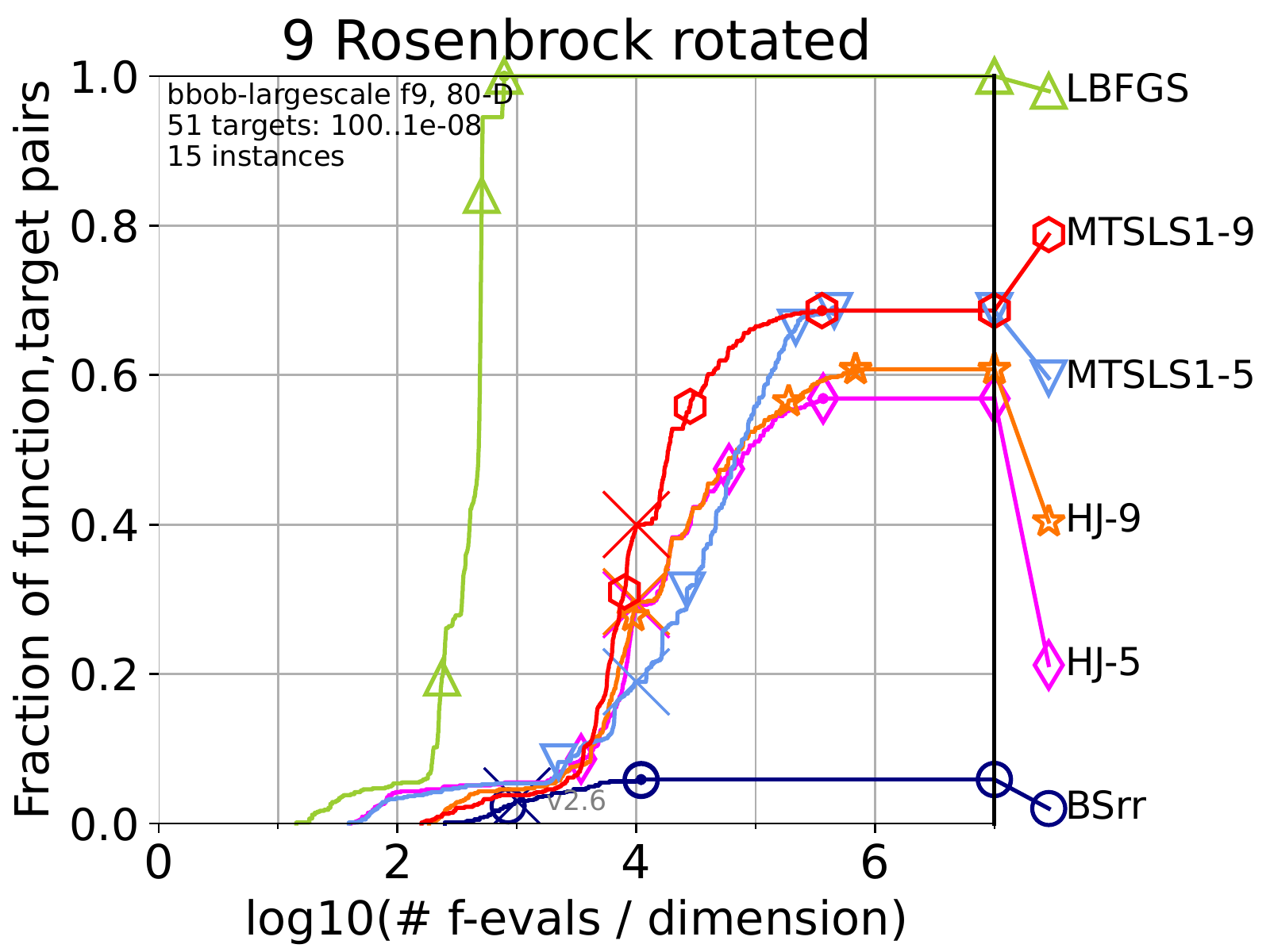}&
\includegraphics[width=\widthvar\textwidth]{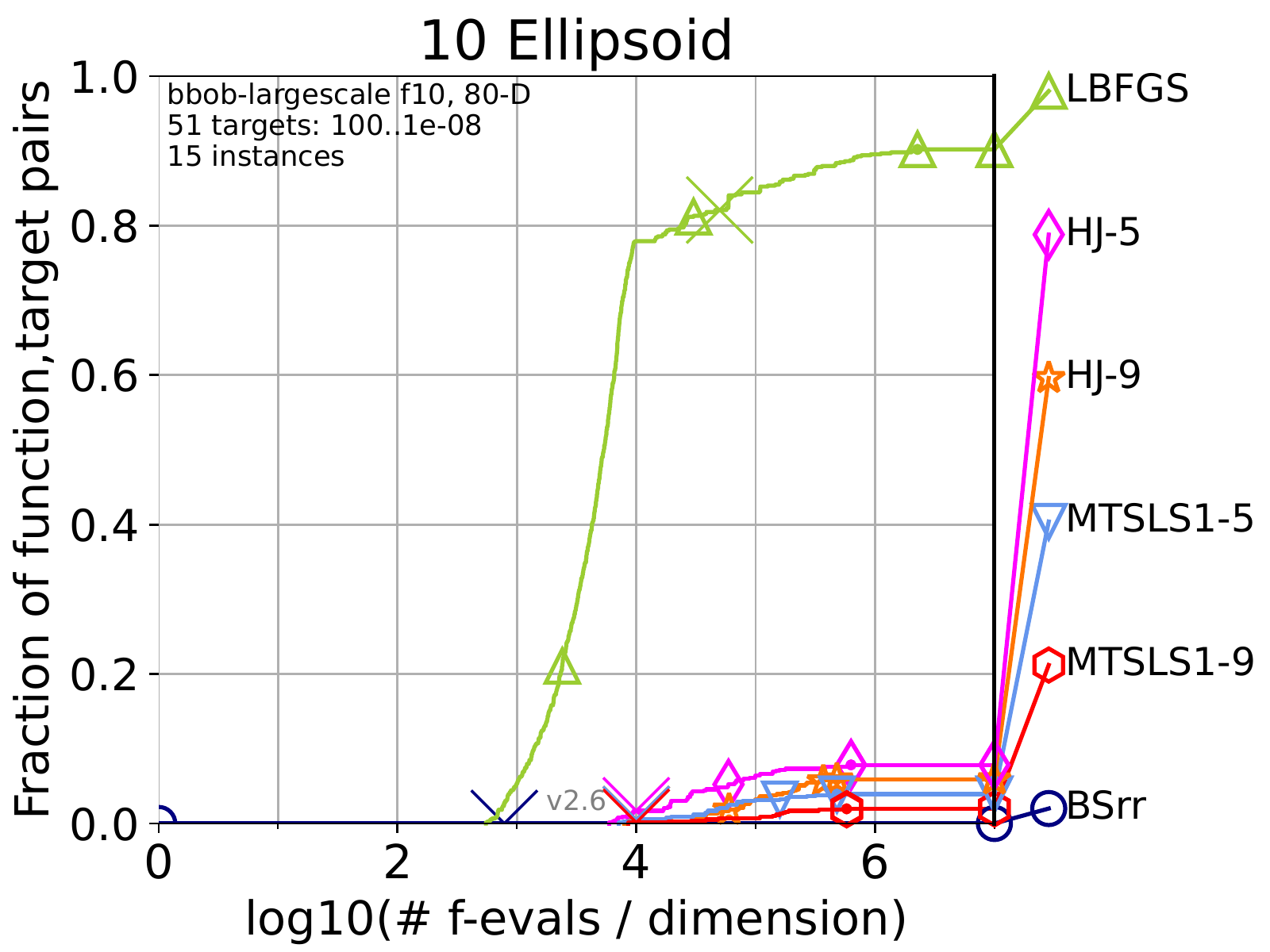}&
\includegraphics[width=\widthvar\textwidth]{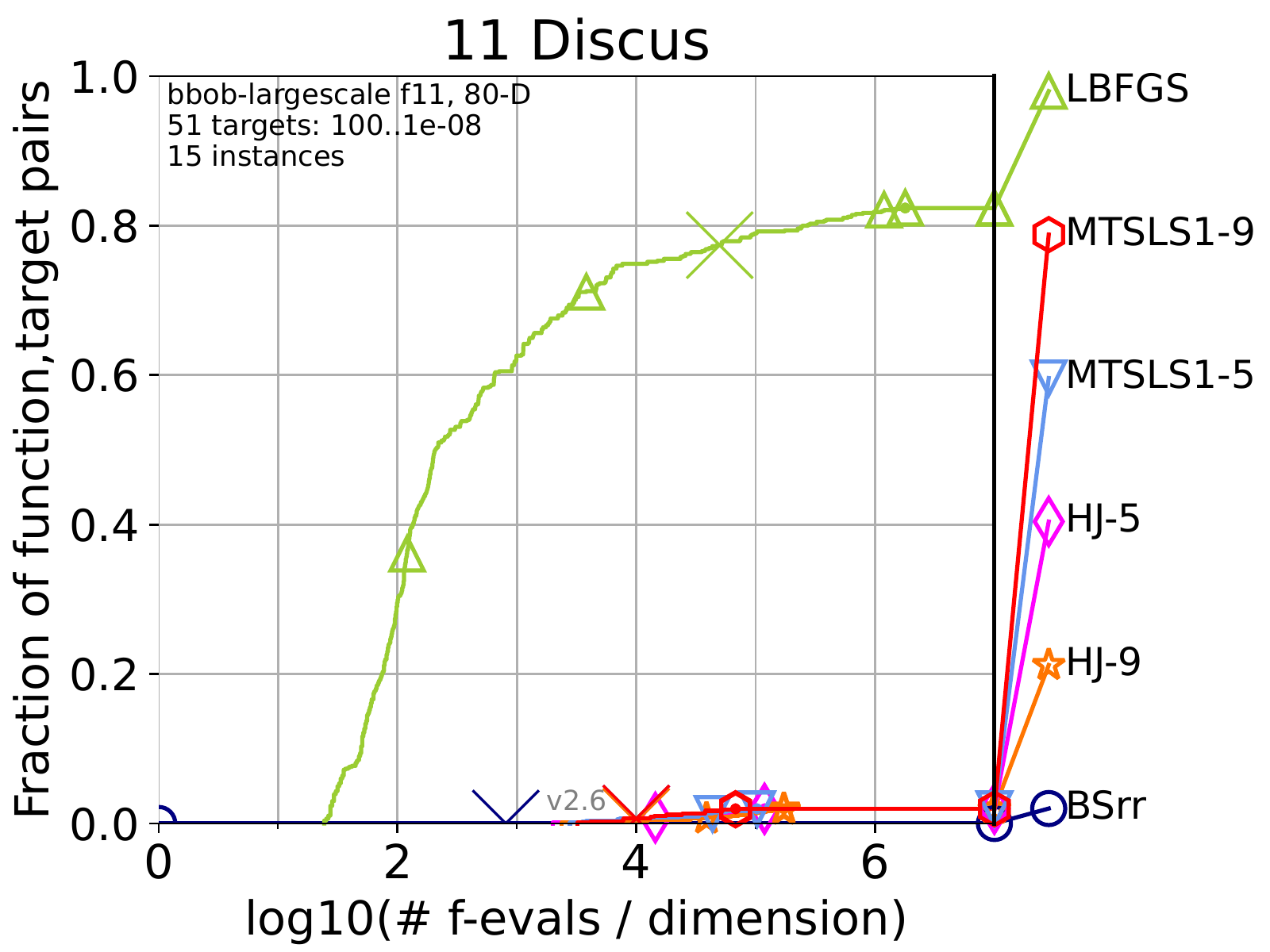}&
\includegraphics[width=\widthvar\textwidth]{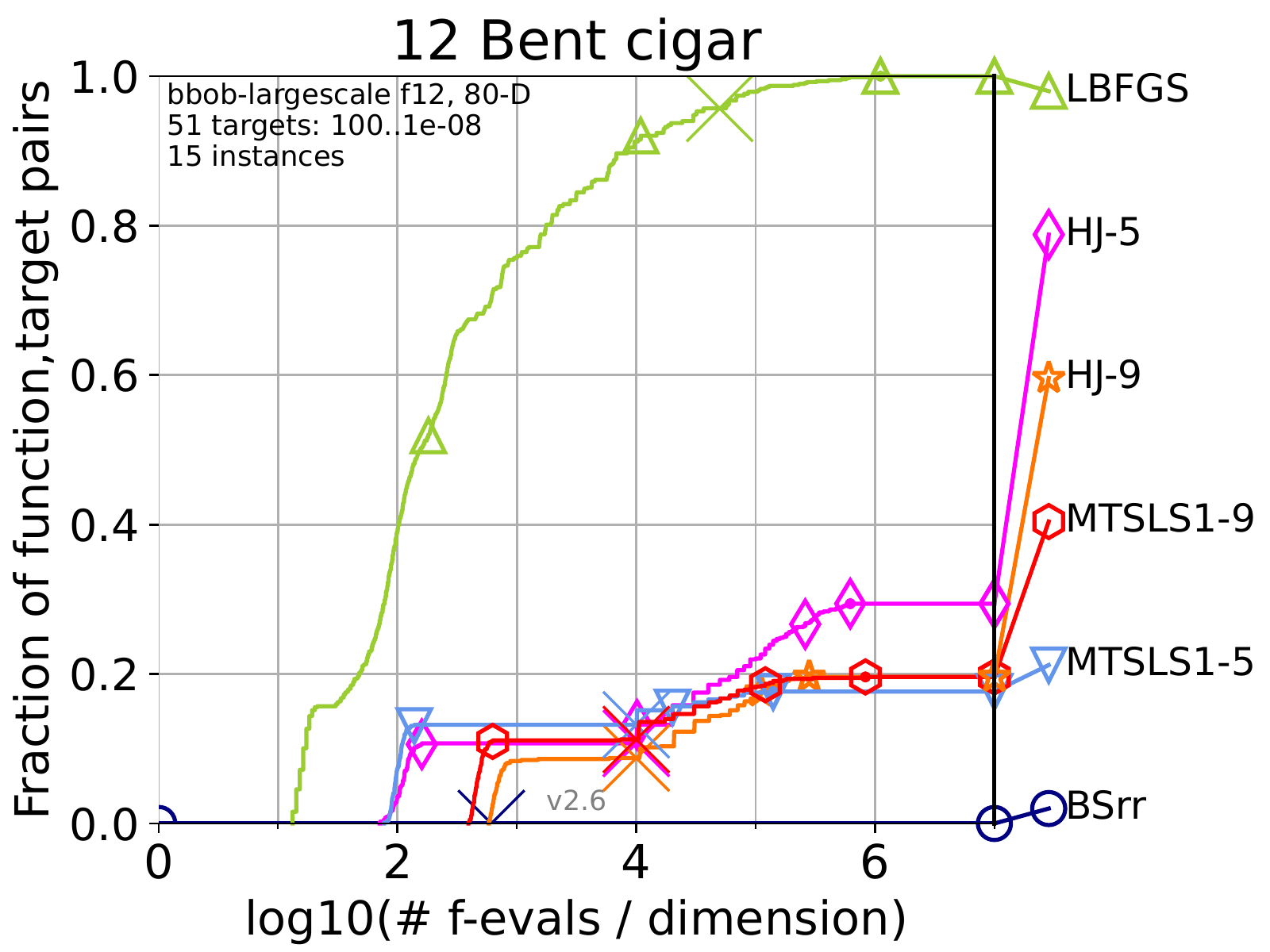}\\
\includegraphics[width=\widthvar\textwidth]{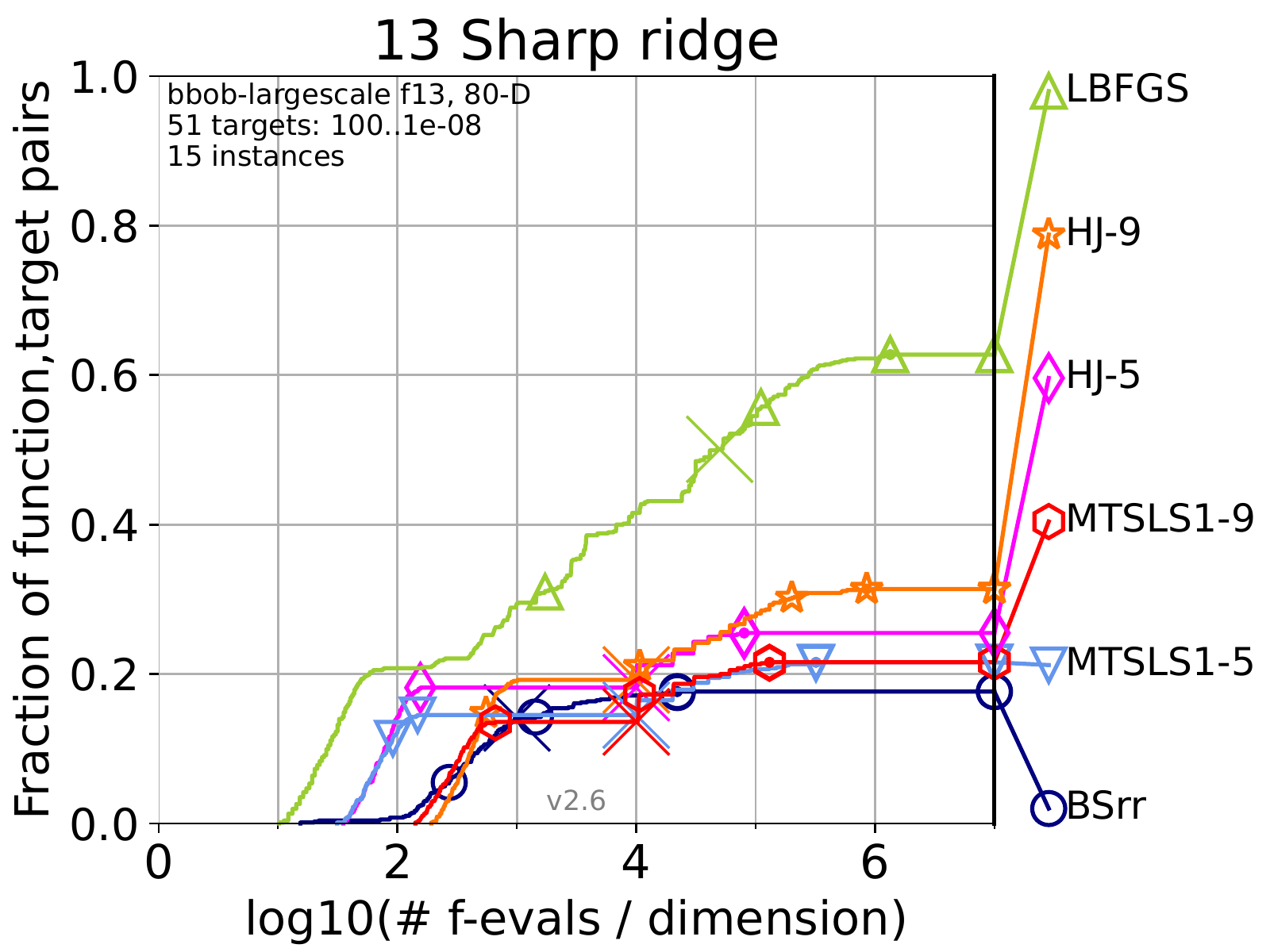}&
\includegraphics[width=\widthvar\textwidth]{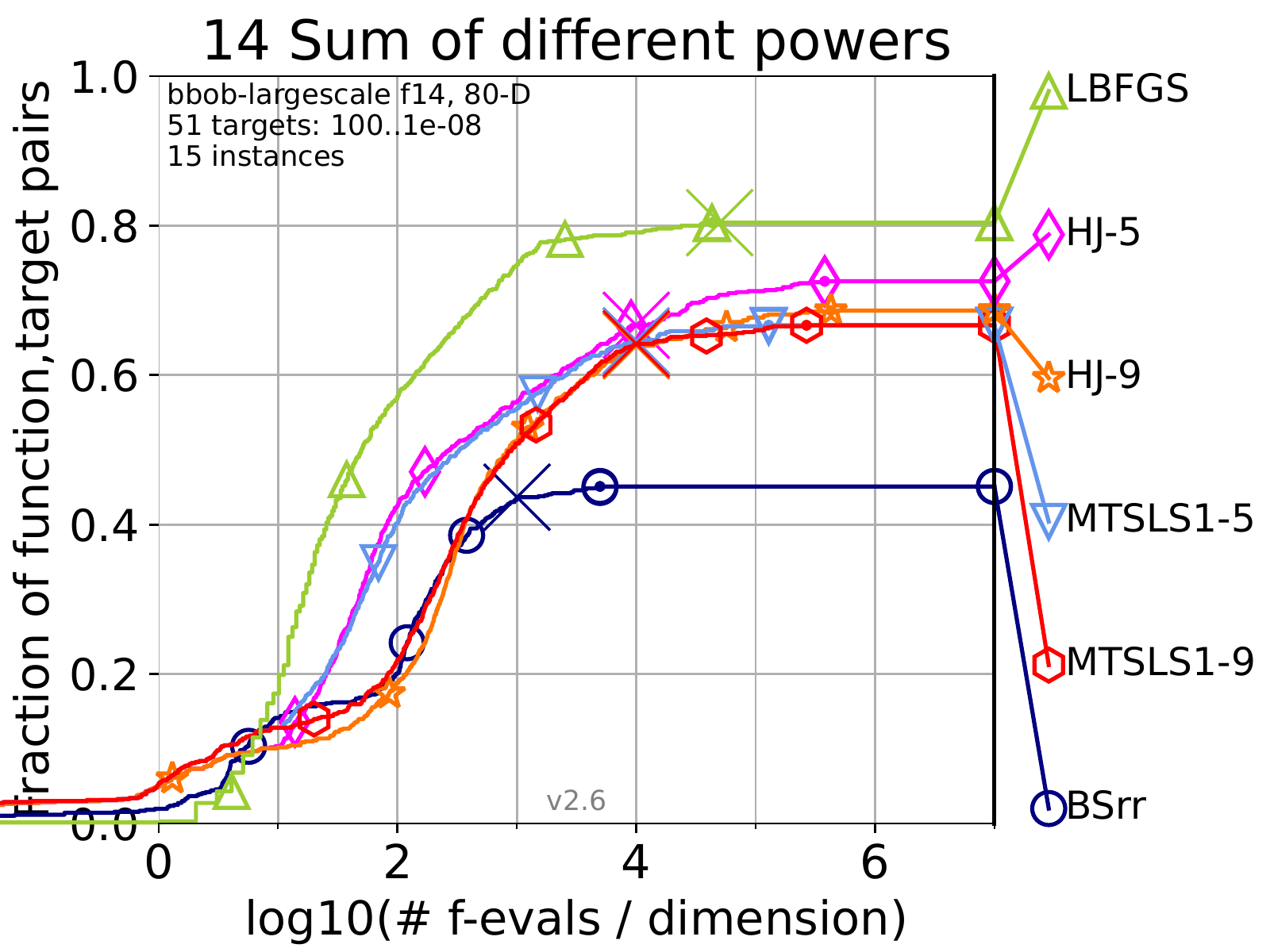}&
\includegraphics[width=\widthvar\textwidth]{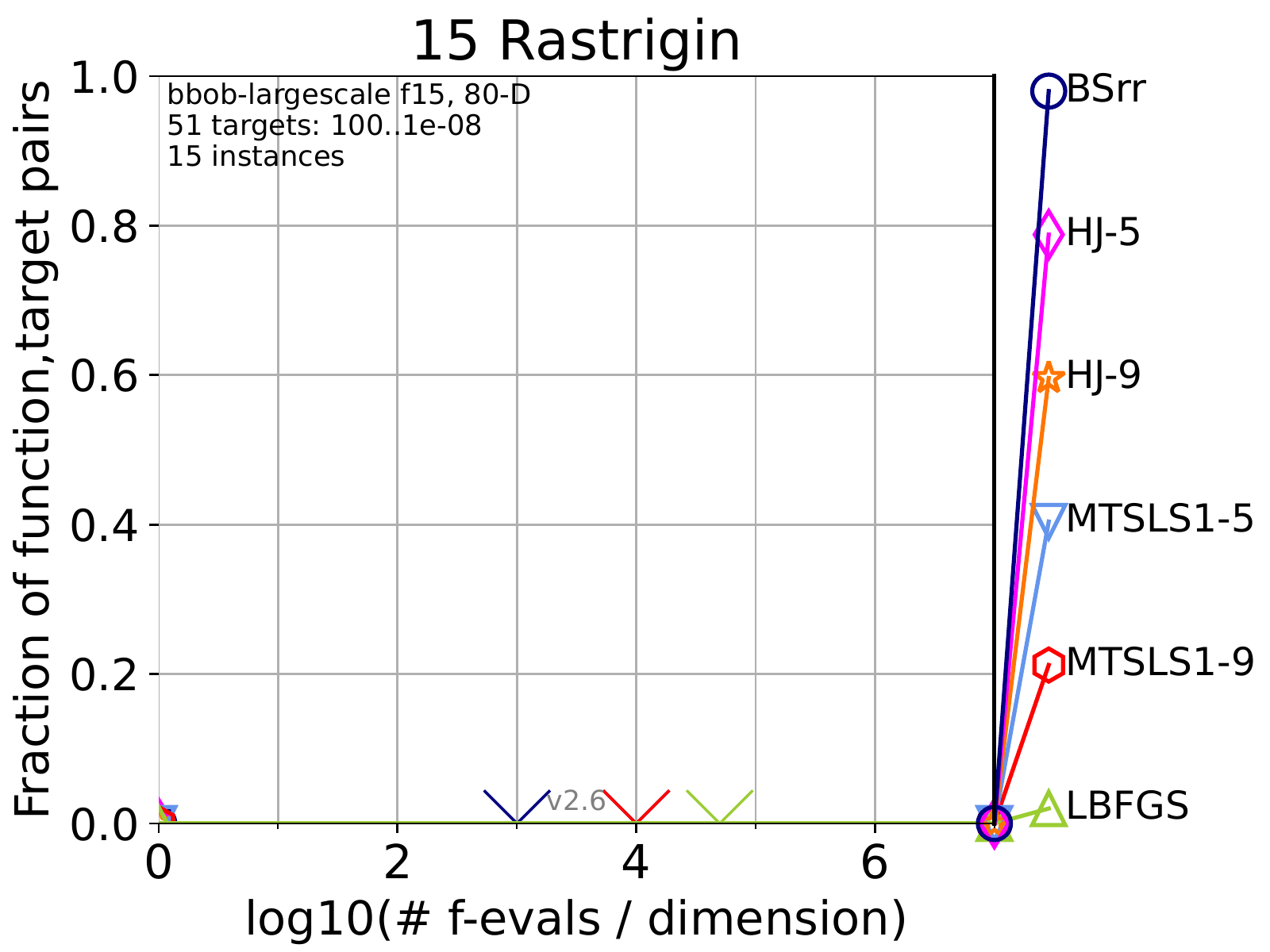}&
\includegraphics[width=\widthvar\textwidth]{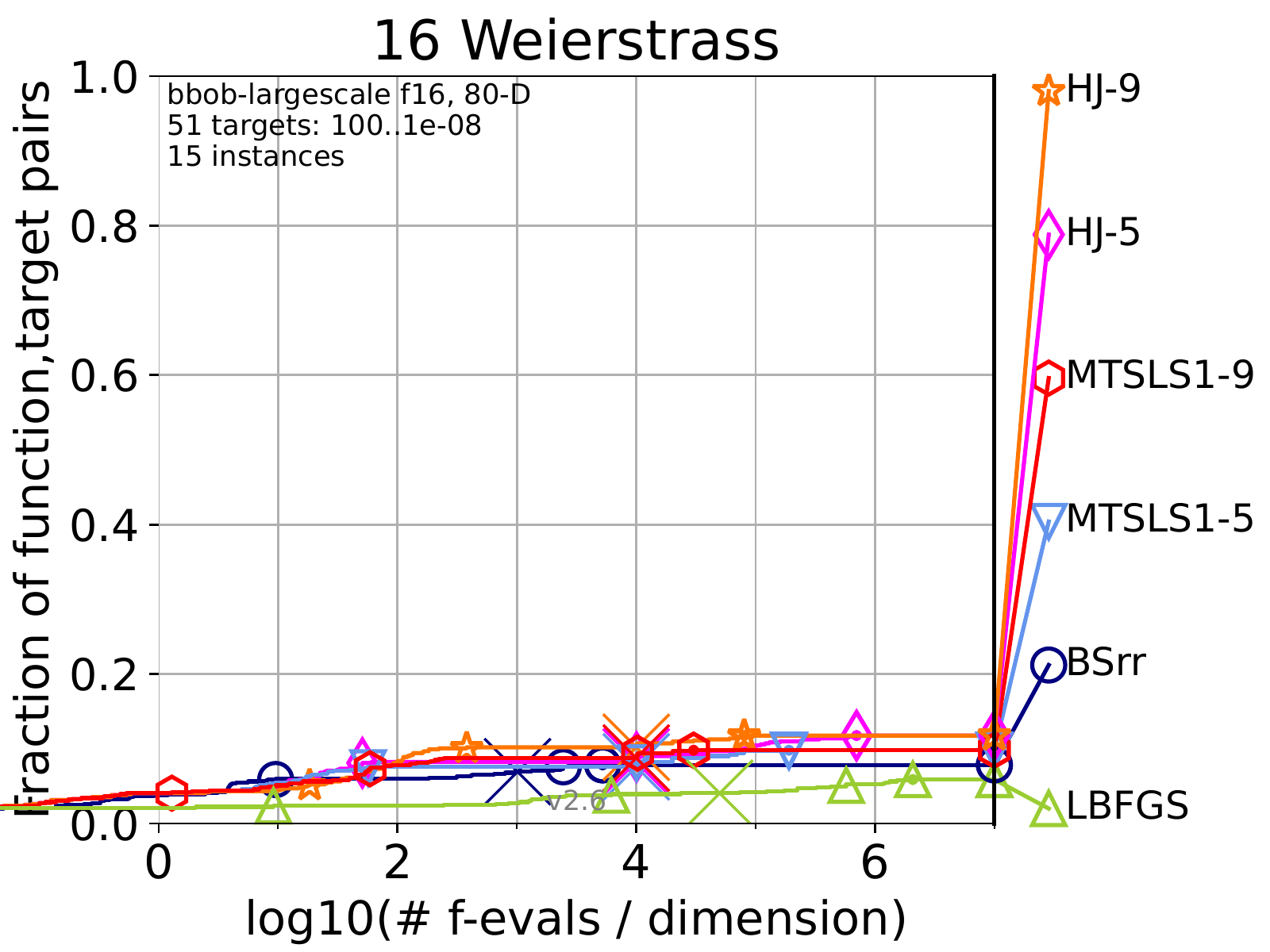}\\
\includegraphics[width=\widthvar\textwidth]{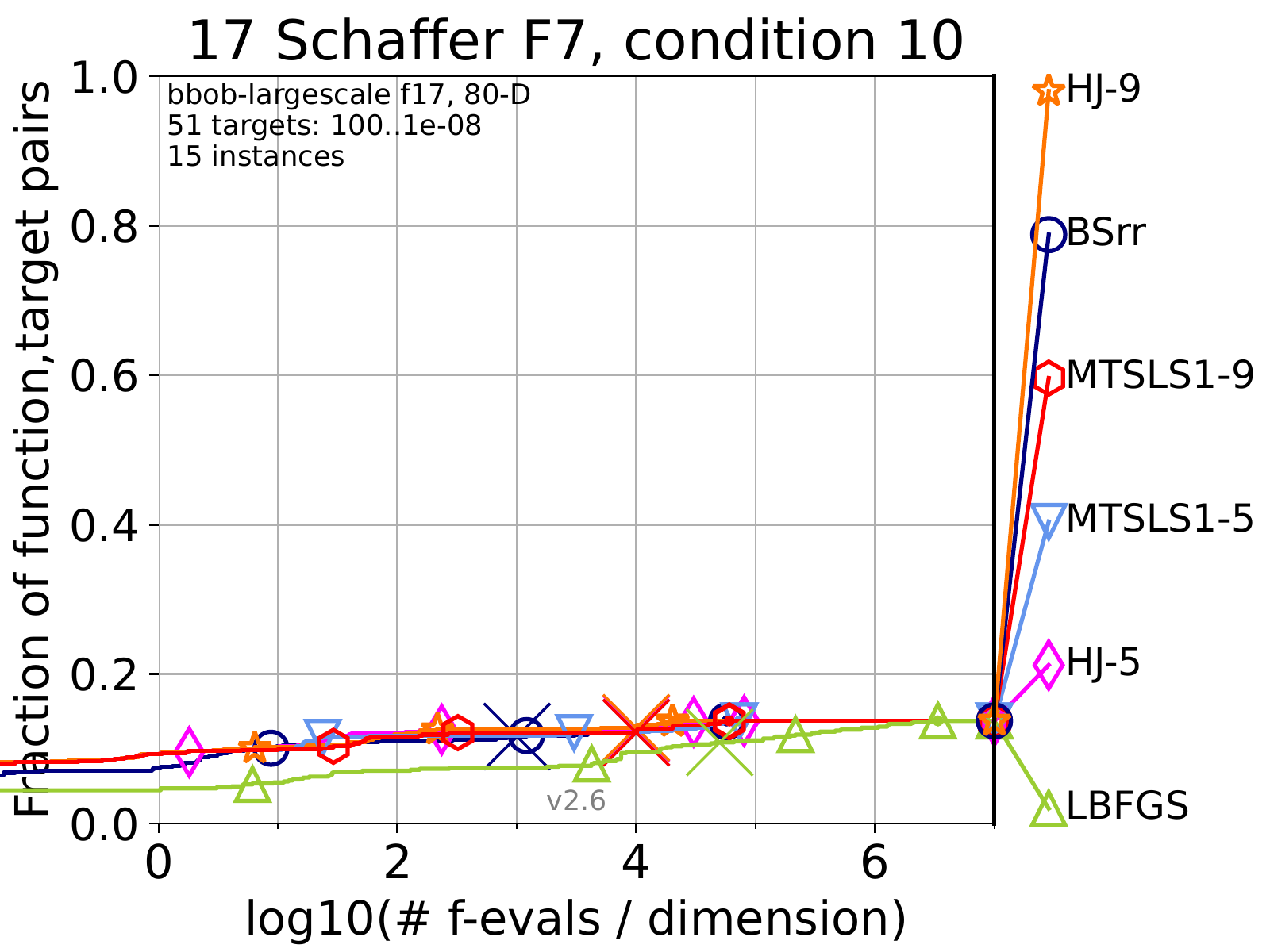}&
\includegraphics[width=\widthvar\textwidth]{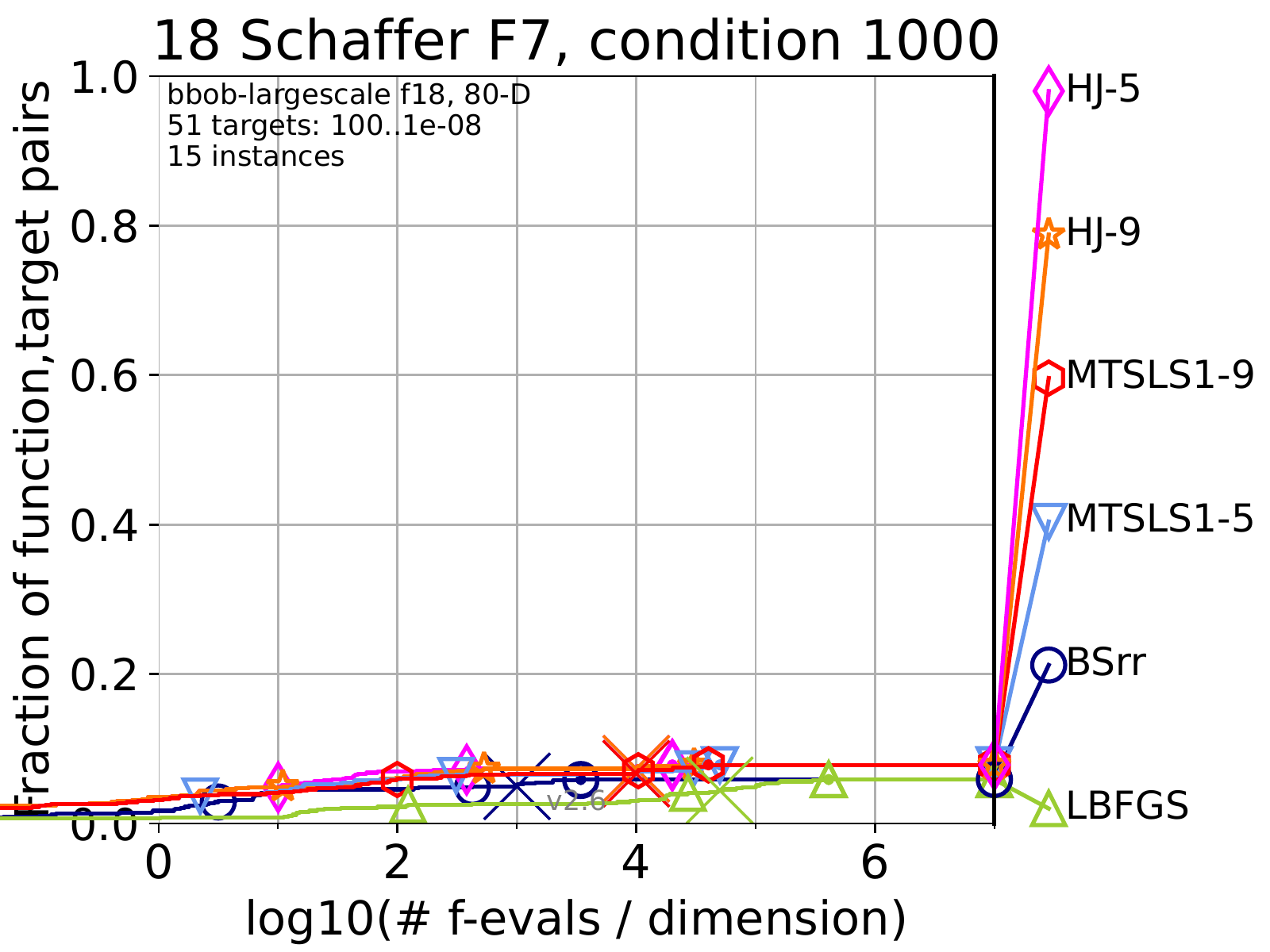}&
\includegraphics[width=\widthvar\textwidth]{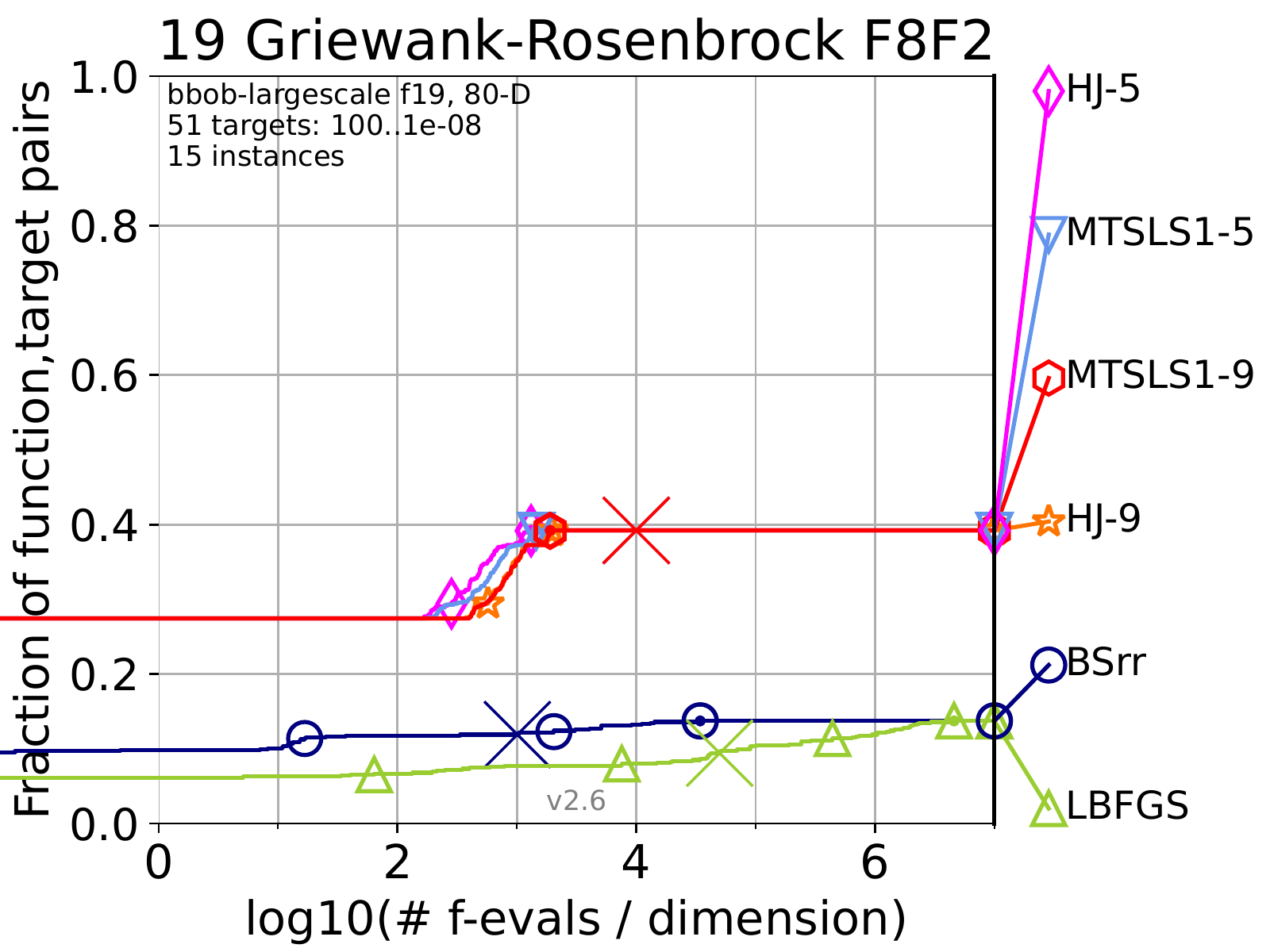}&
\includegraphics[width=\widthvar\textwidth]{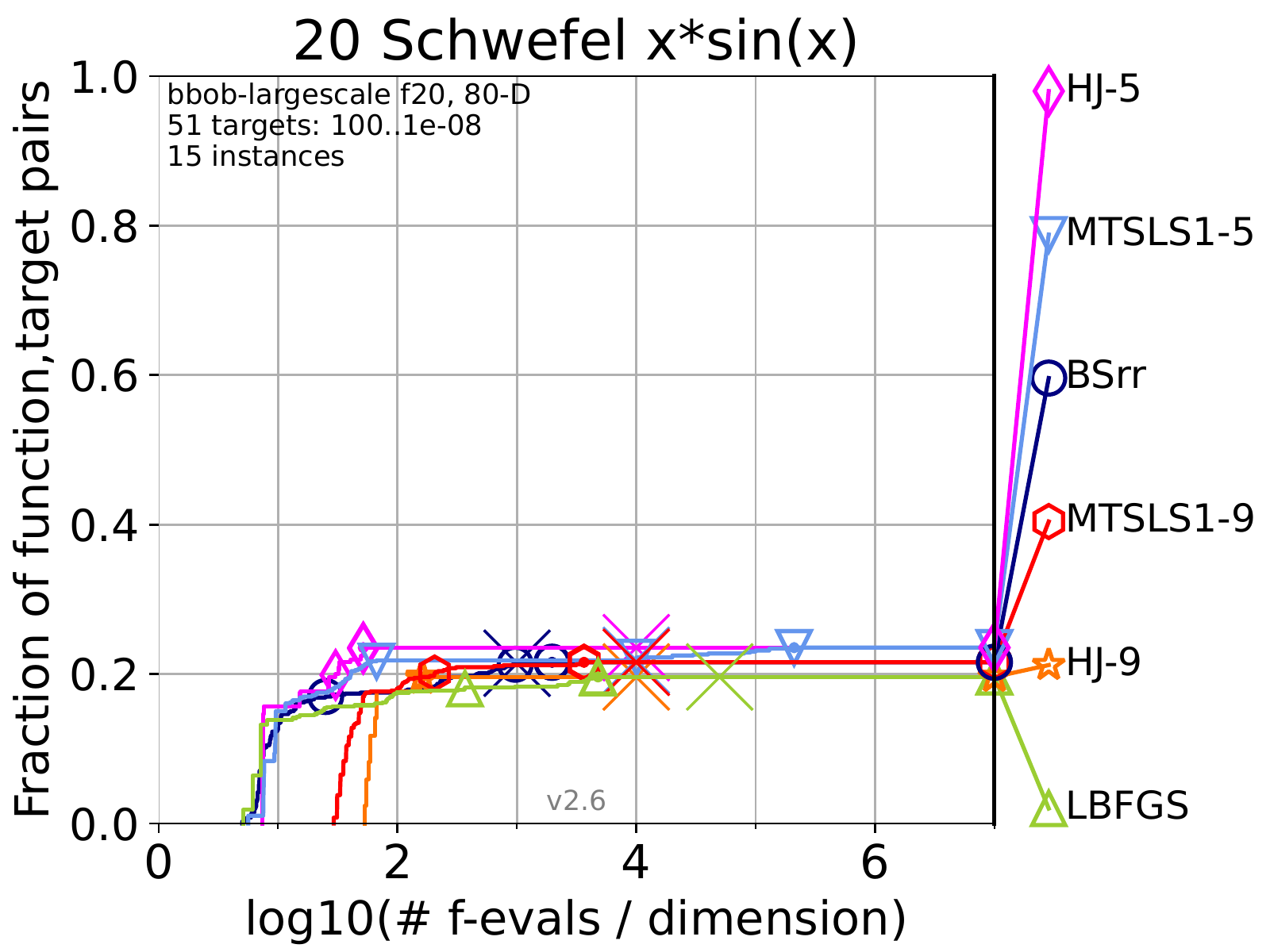}\\
\includegraphics[width=\widthvar\textwidth]{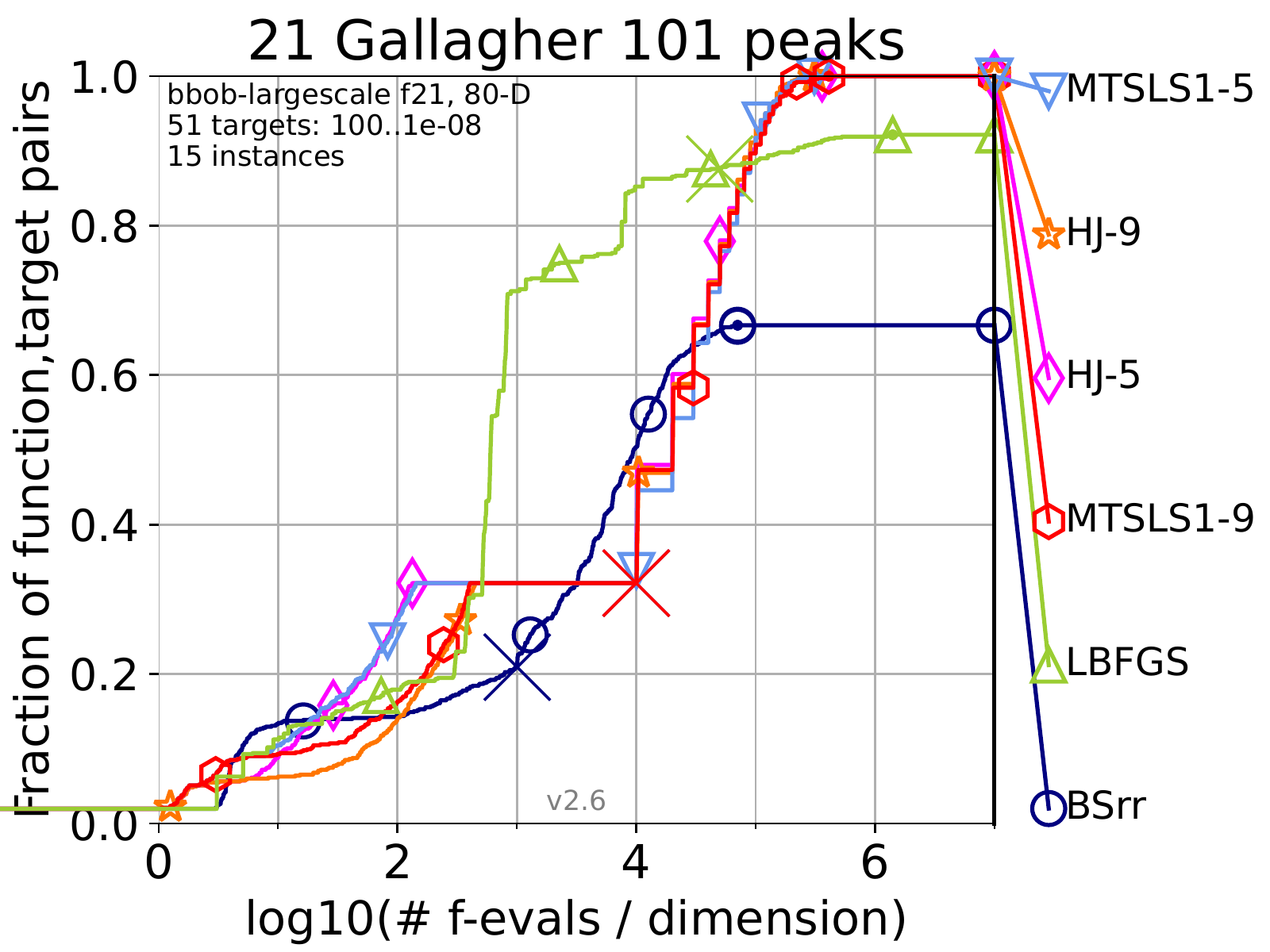}&
\includegraphics[width=\widthvar\textwidth]{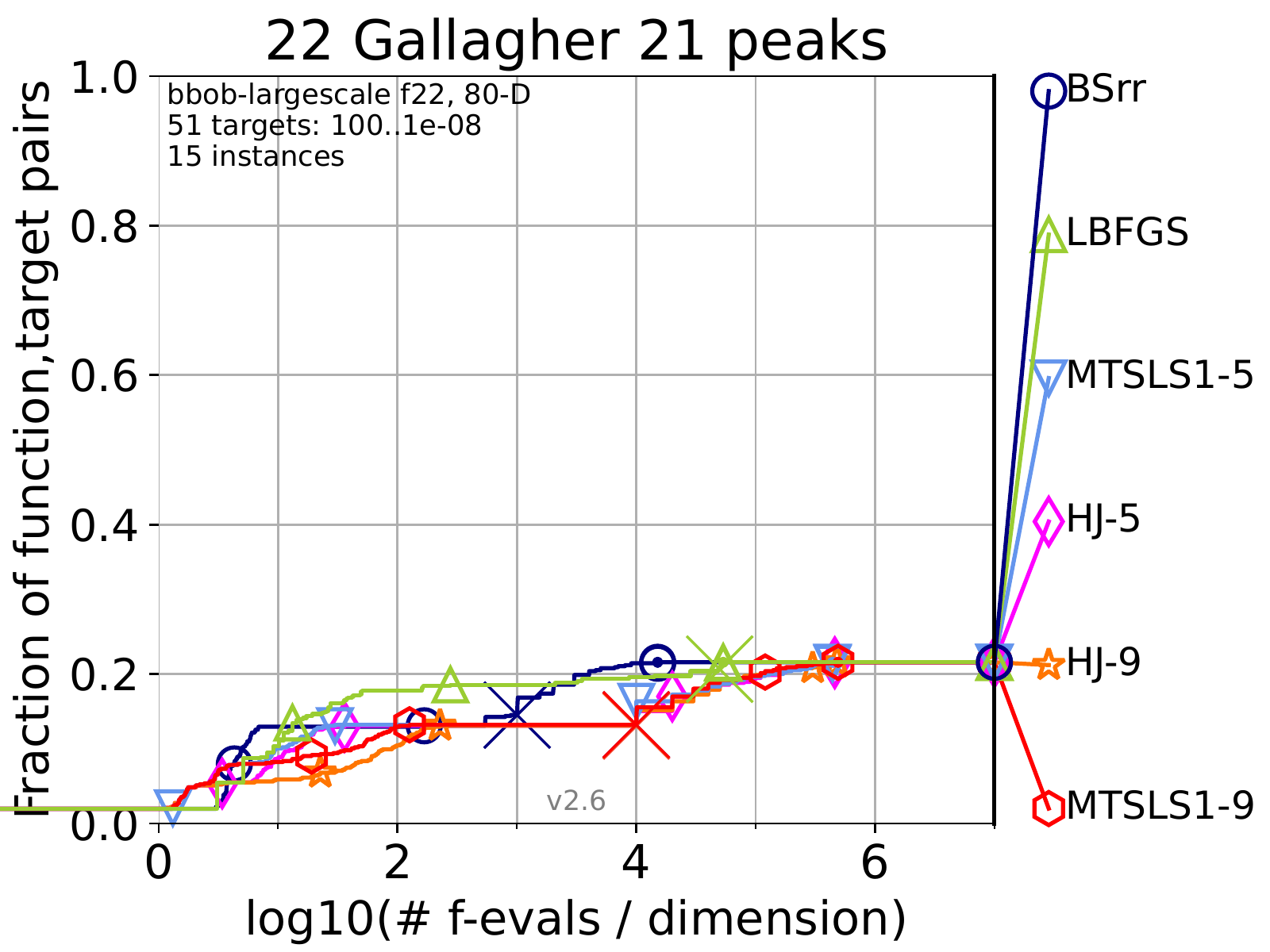}&
\includegraphics[width=\widthvar\textwidth]{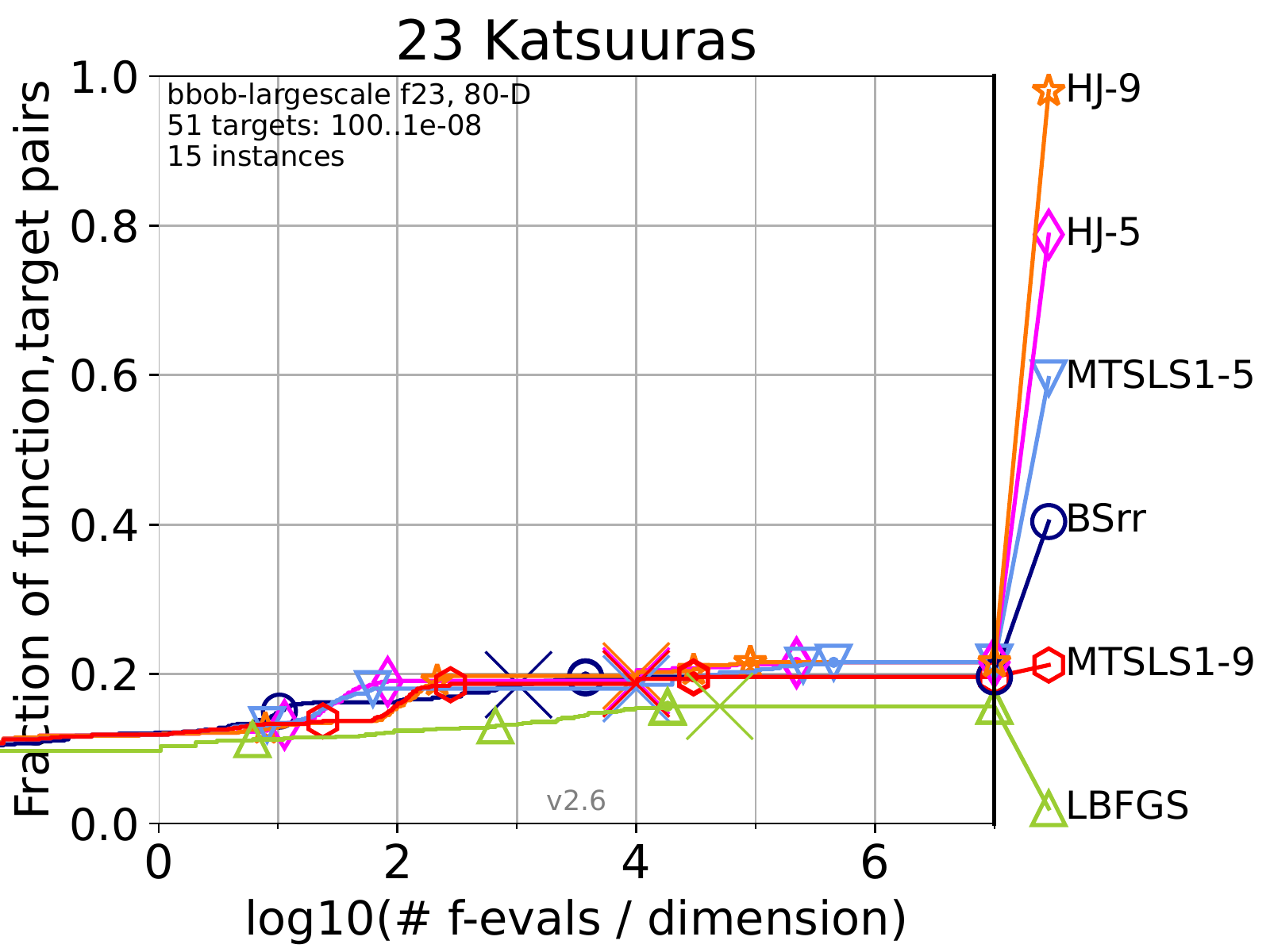}&
\includegraphics[width=\widthvar\textwidth]{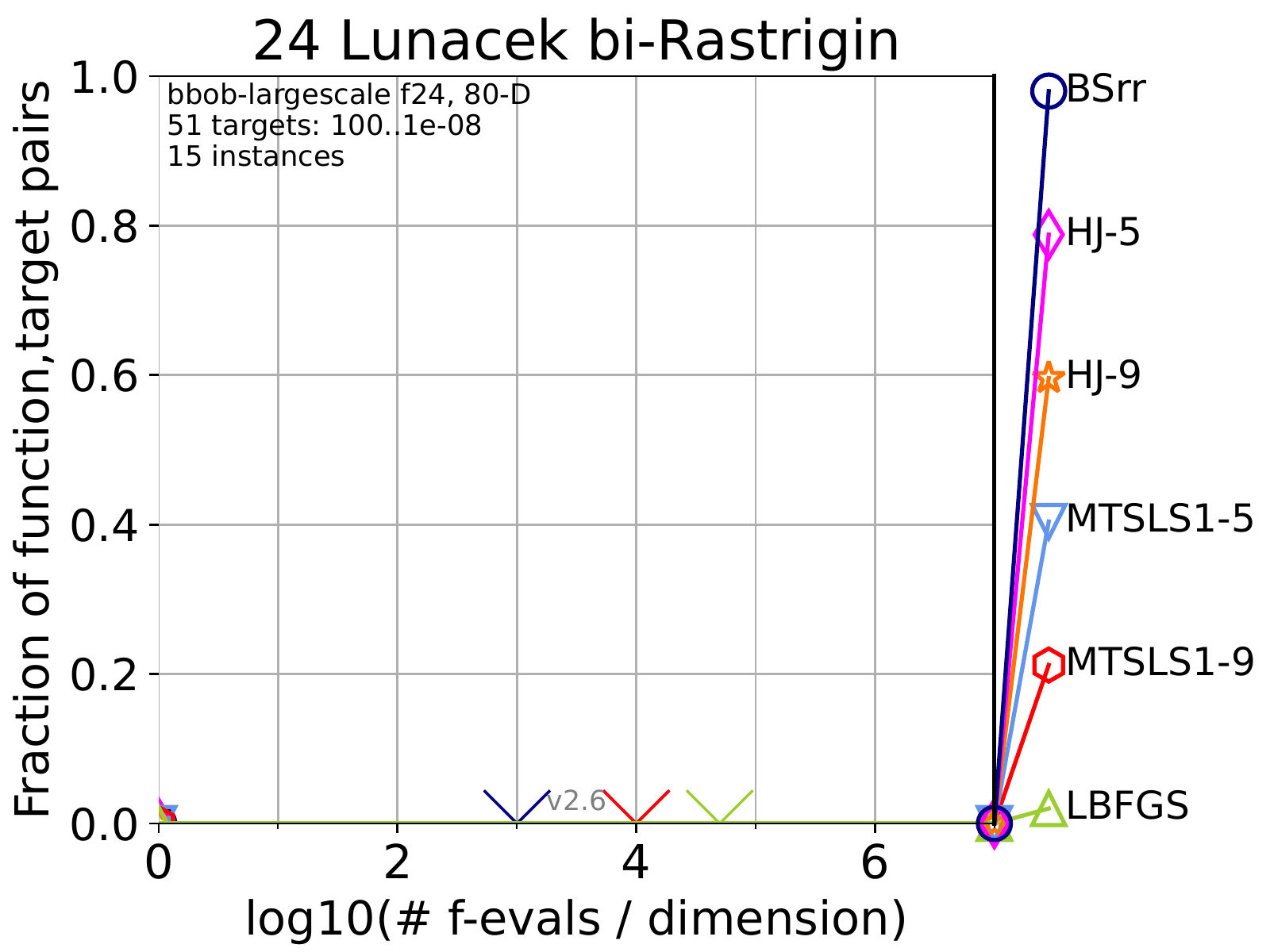}
\end{tabular}
 \caption{\label{fig:ECDFsingleOne80D}
	\bbobecdfcaptionsinglefunctionssingledim{80}
}
\end{figure*}

\begin{figure*}
  \newcommand{\widthvar}{0.2}    
  \centering
\begin{tabular}{@{}l@{}l@{}l@{}l@{}l@{}}
\includegraphics[width=\widthvar\textwidth]{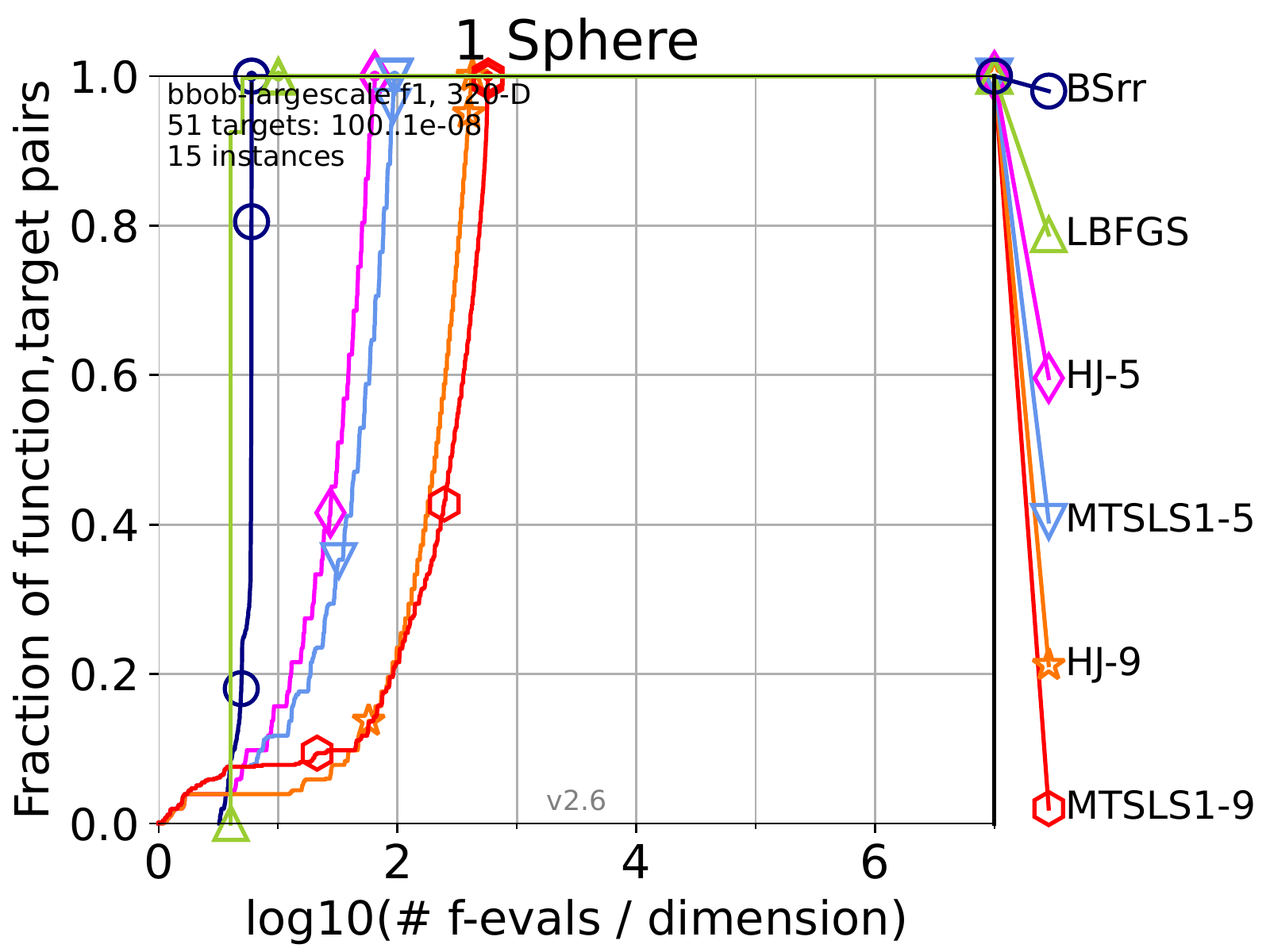}&
\includegraphics[width=\widthvar\textwidth]{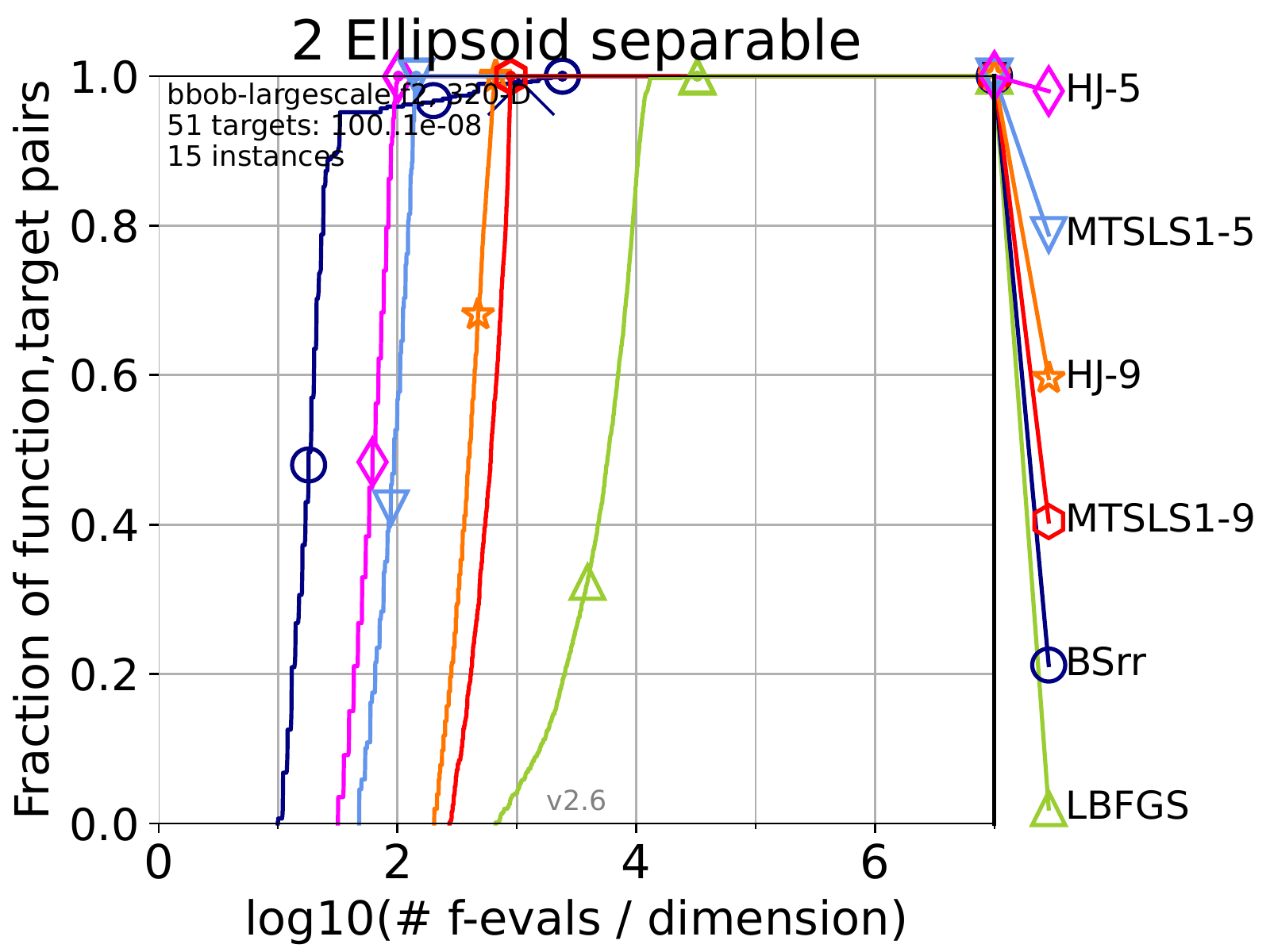}&
\includegraphics[width=\widthvar\textwidth]{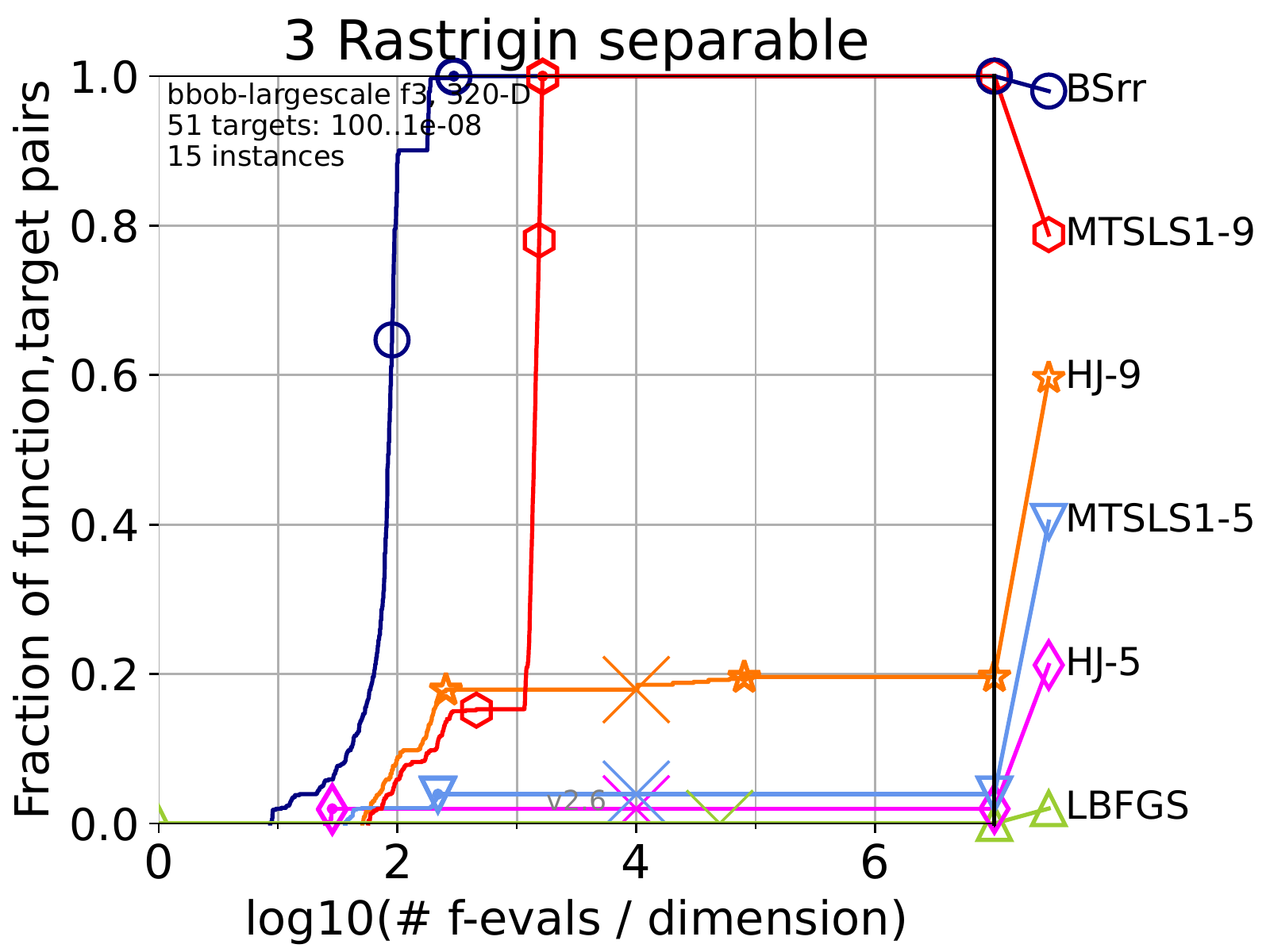}&
\includegraphics[width=\widthvar\textwidth]{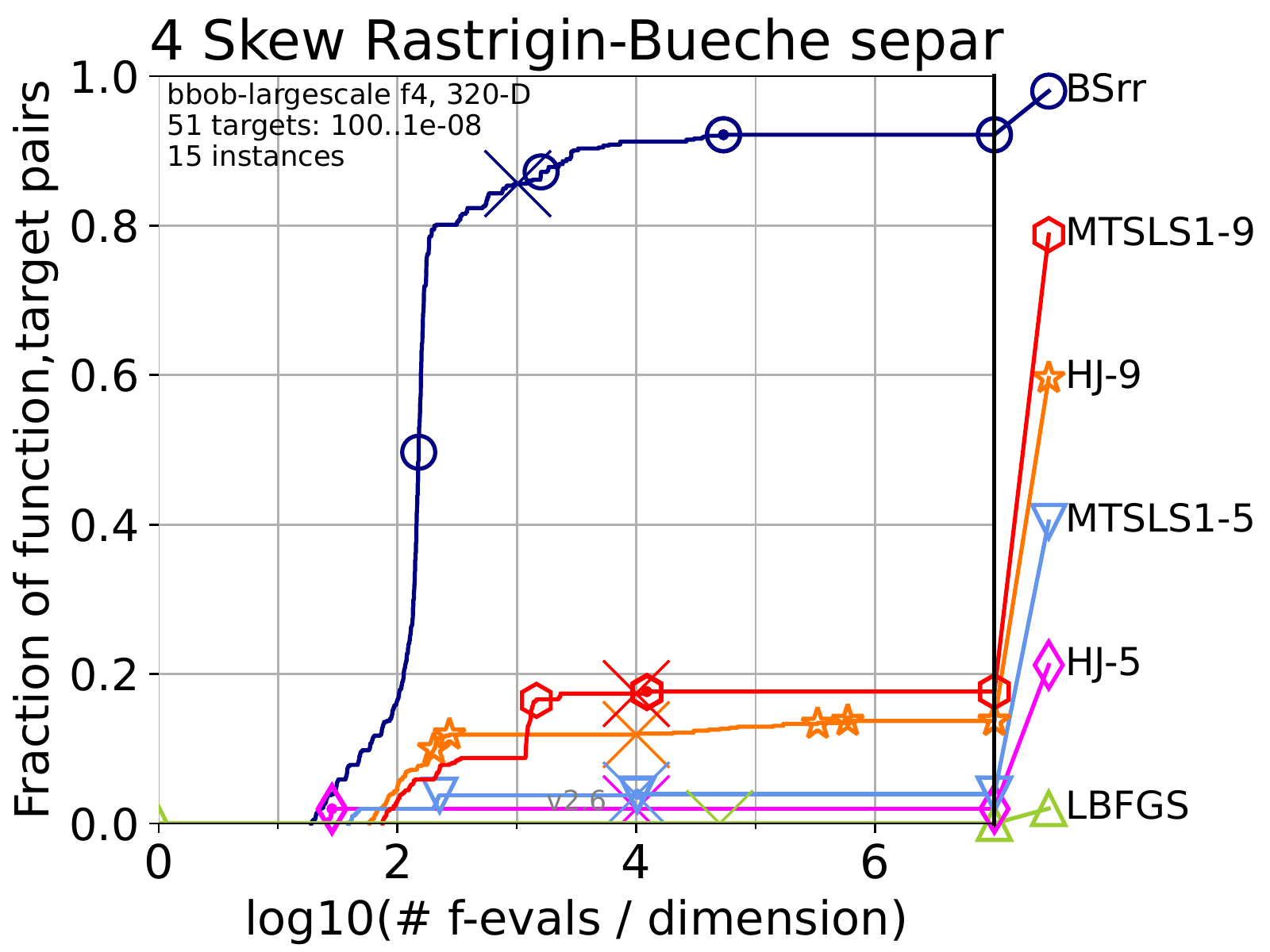}\\
\includegraphics[width=\widthvar\textwidth]{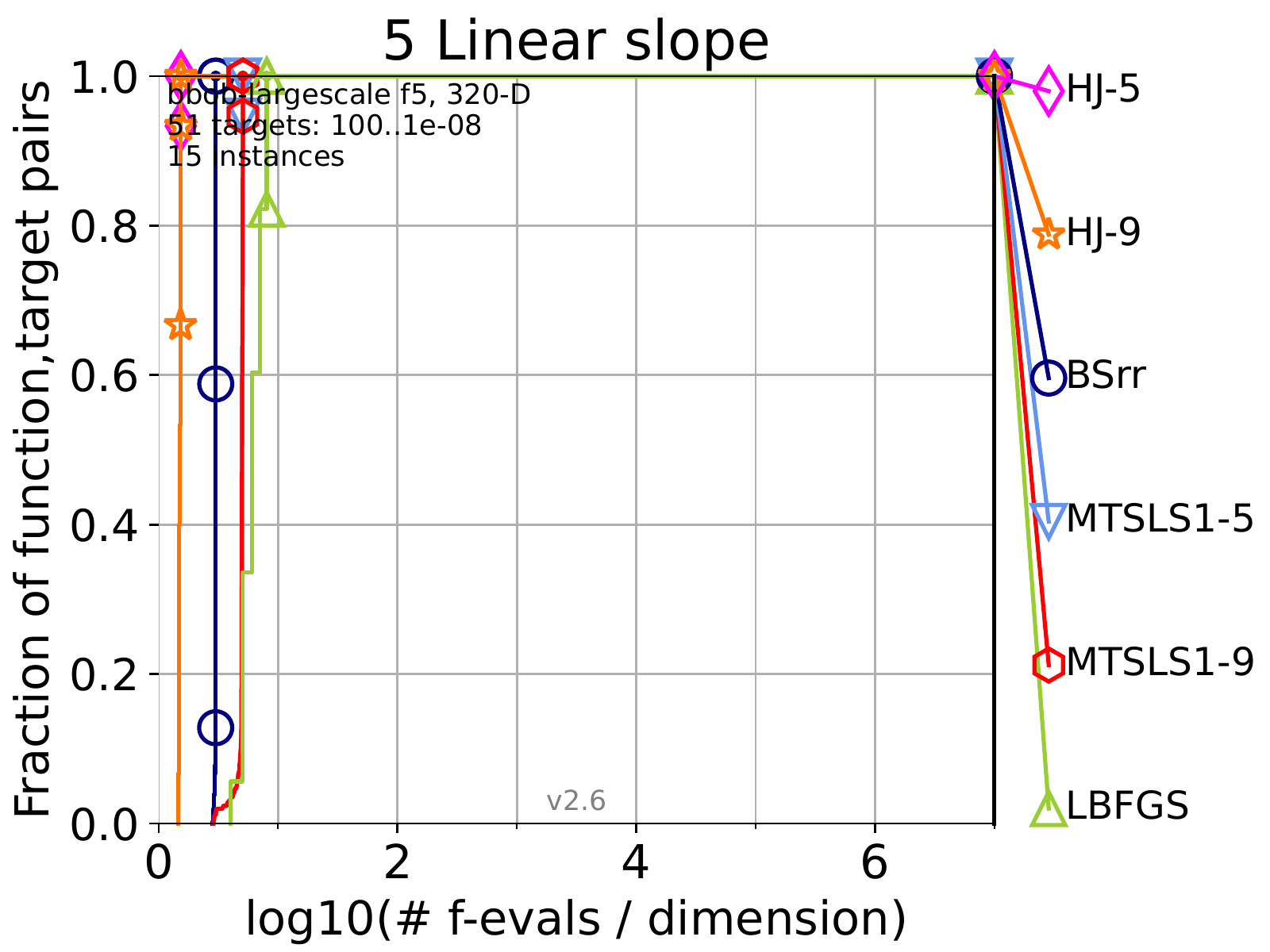}&
\includegraphics[width=\widthvar\textwidth]{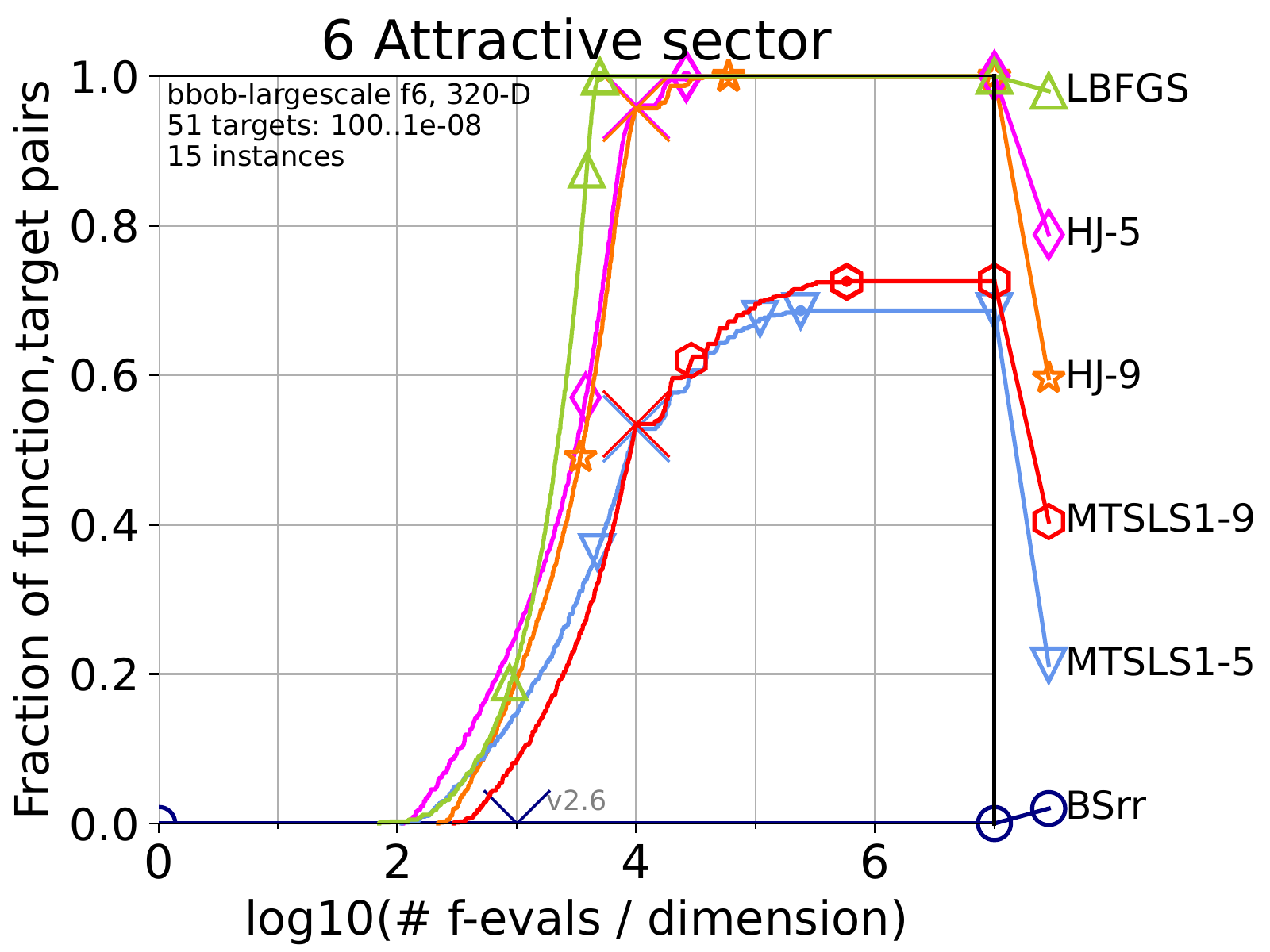}&
\includegraphics[width=\widthvar\textwidth]{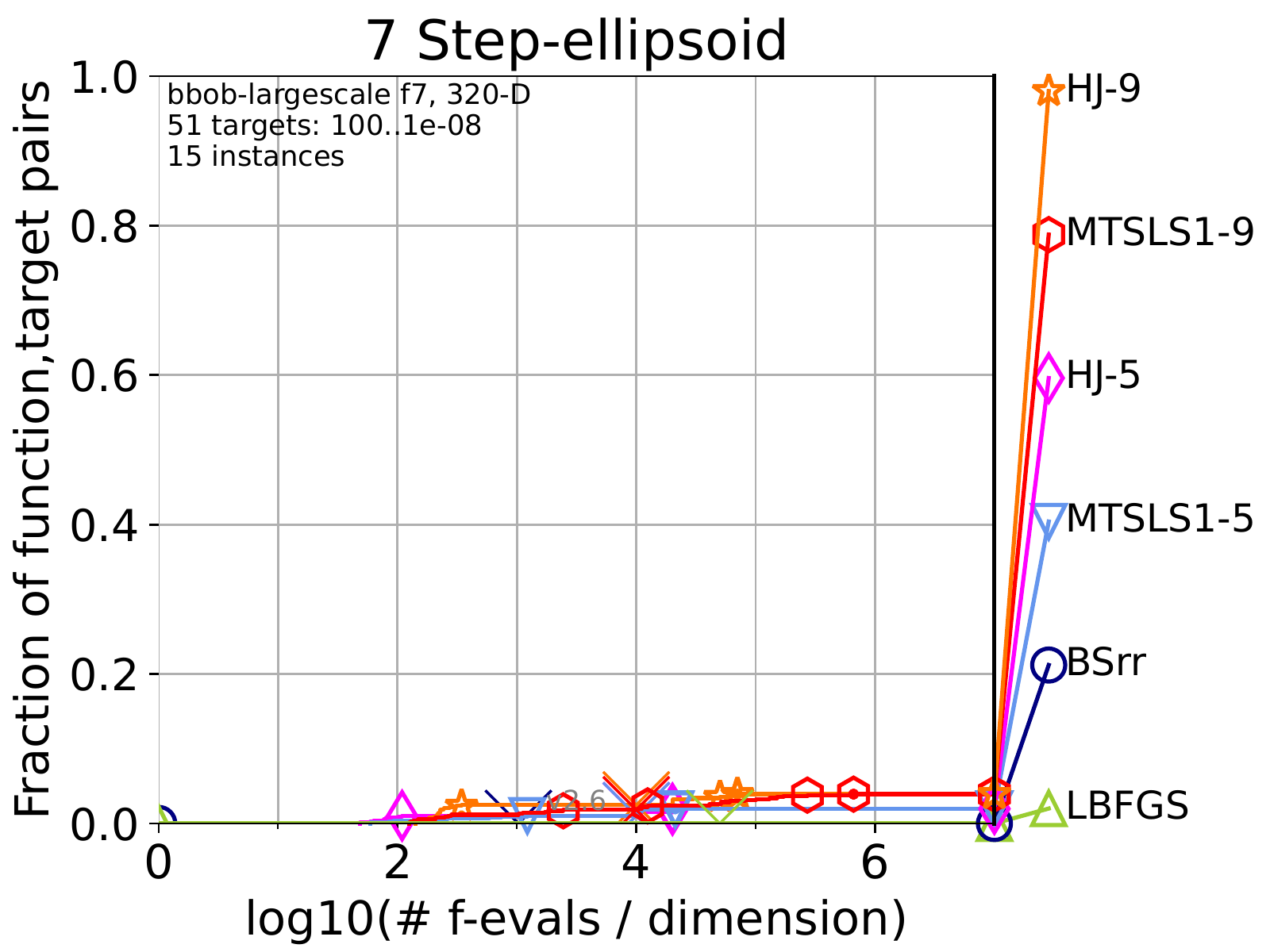}&
\includegraphics[width=\widthvar\textwidth]{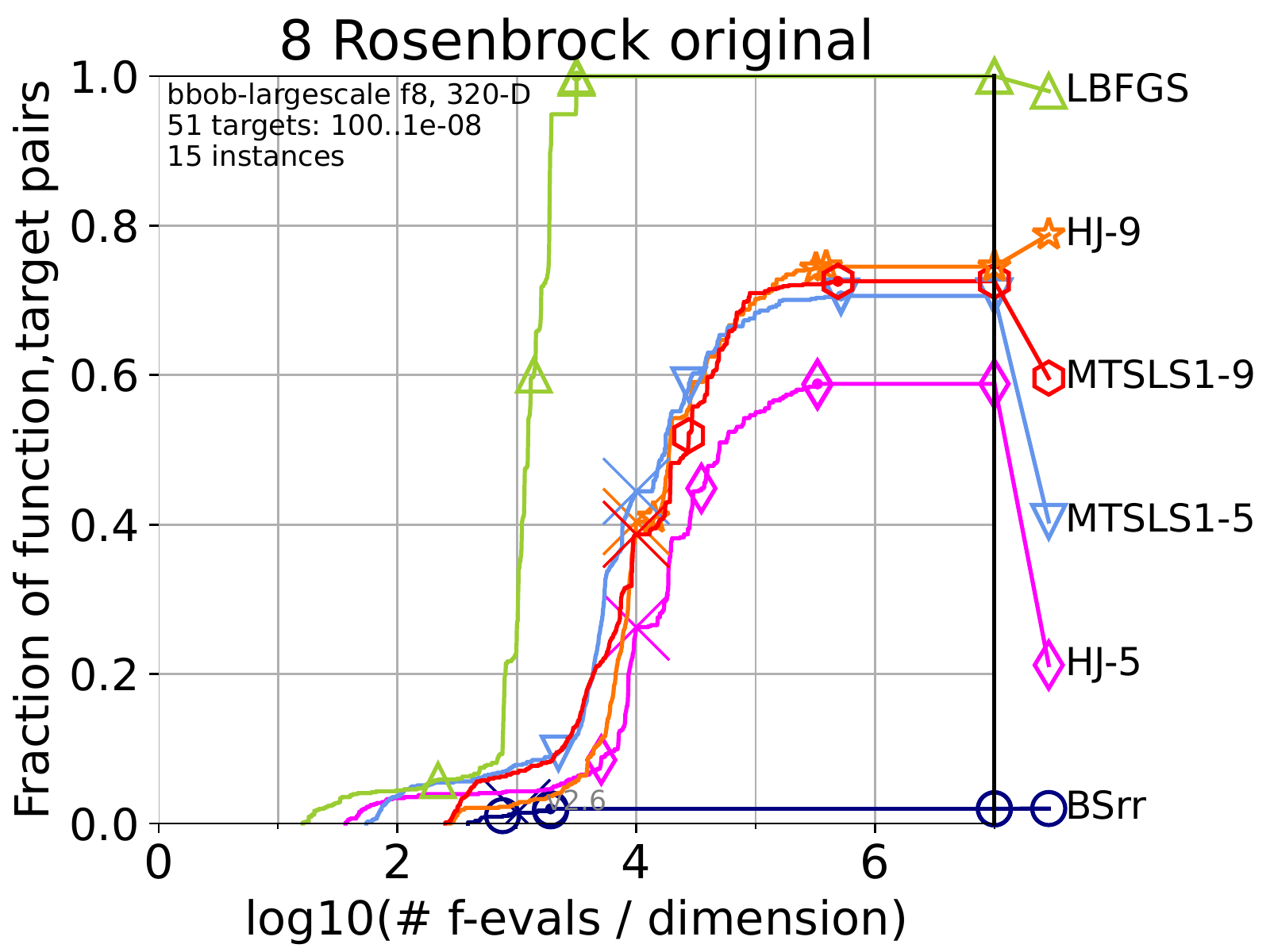}\\
\includegraphics[width=\widthvar\textwidth]{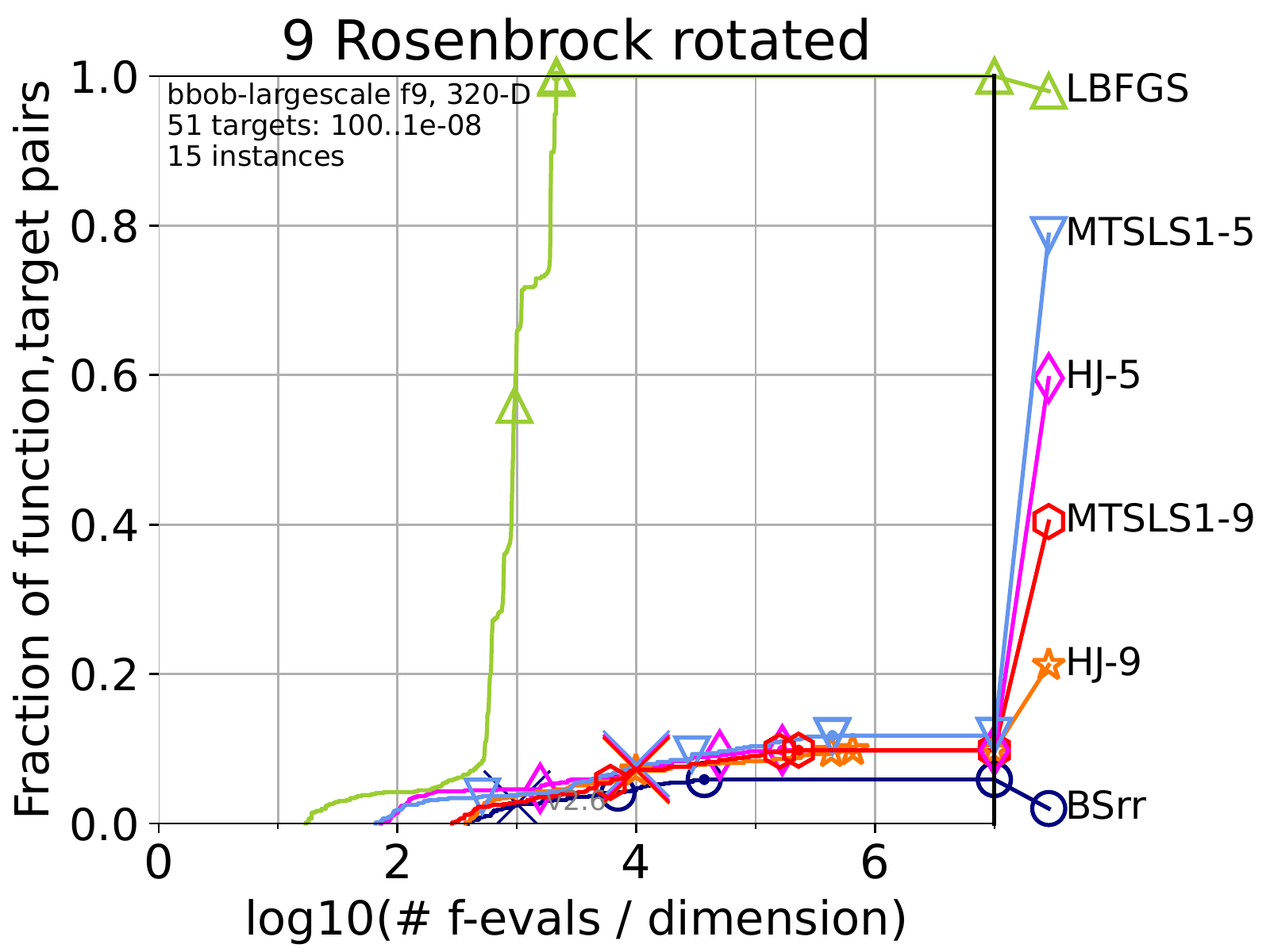}&
\includegraphics[width=\widthvar\textwidth]{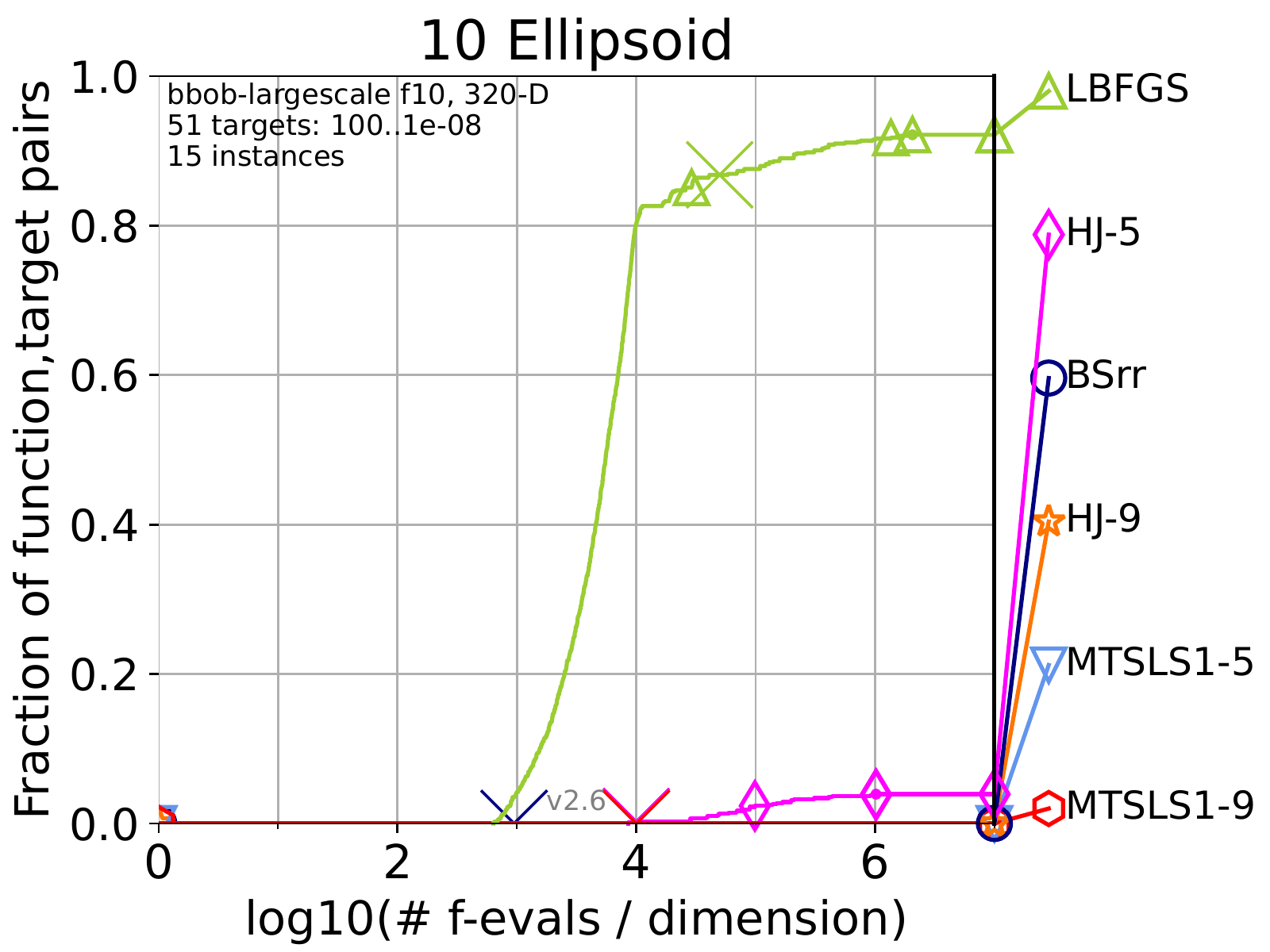}&
\includegraphics[width=\widthvar\textwidth]{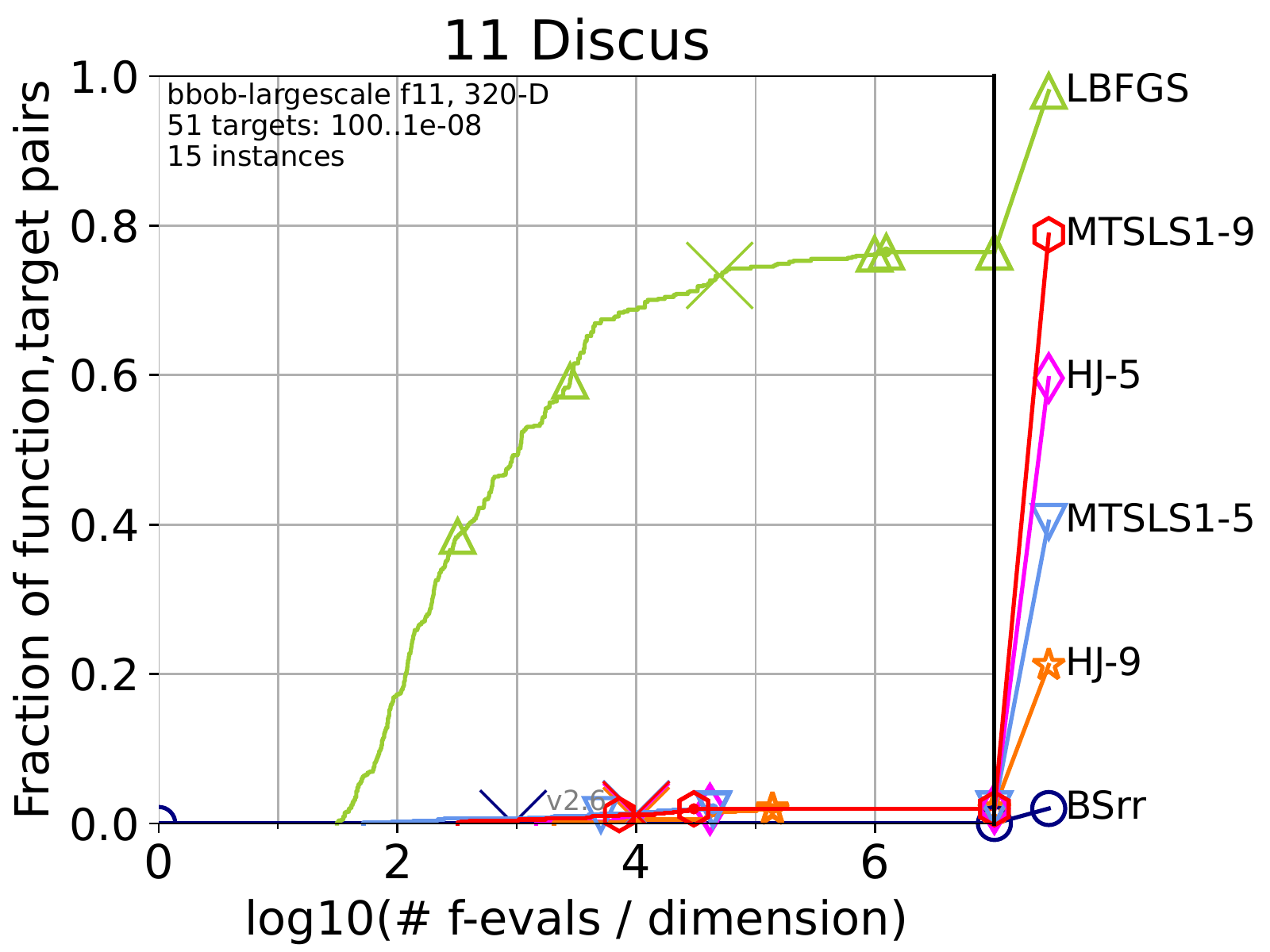}&
\includegraphics[width=\widthvar\textwidth]{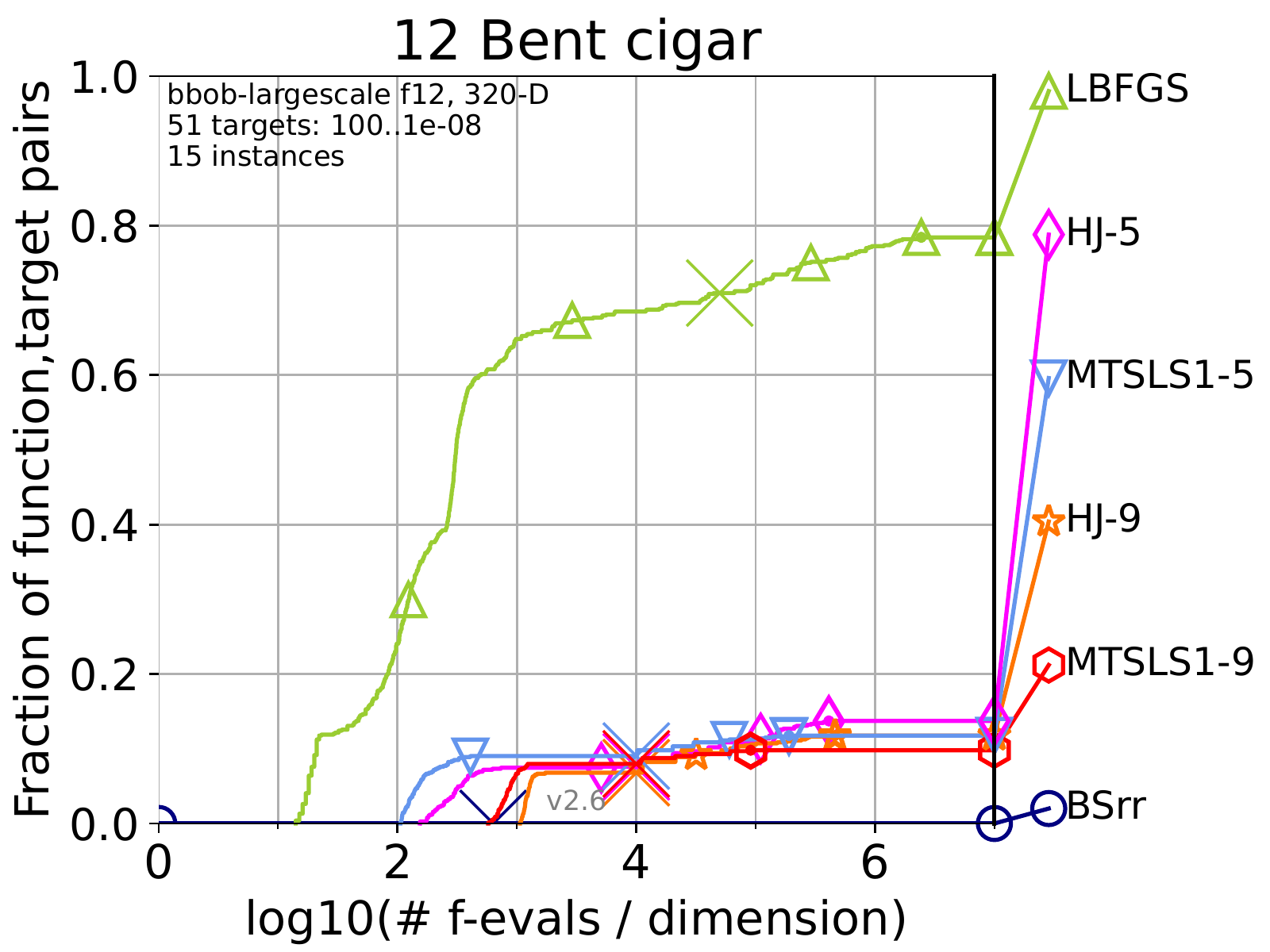}\\
\includegraphics[width=\widthvar\textwidth]{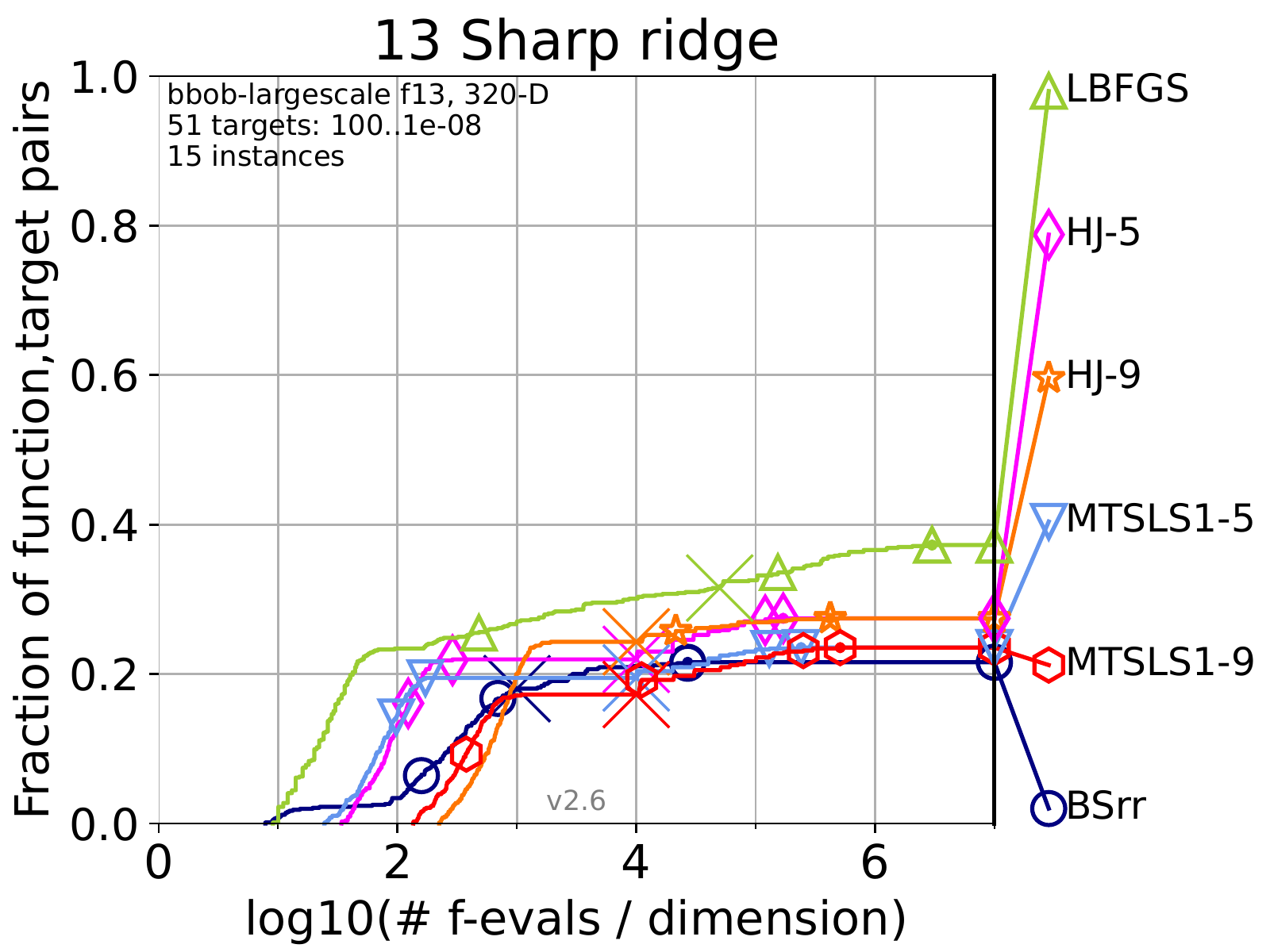}&
\includegraphics[width=\widthvar\textwidth]{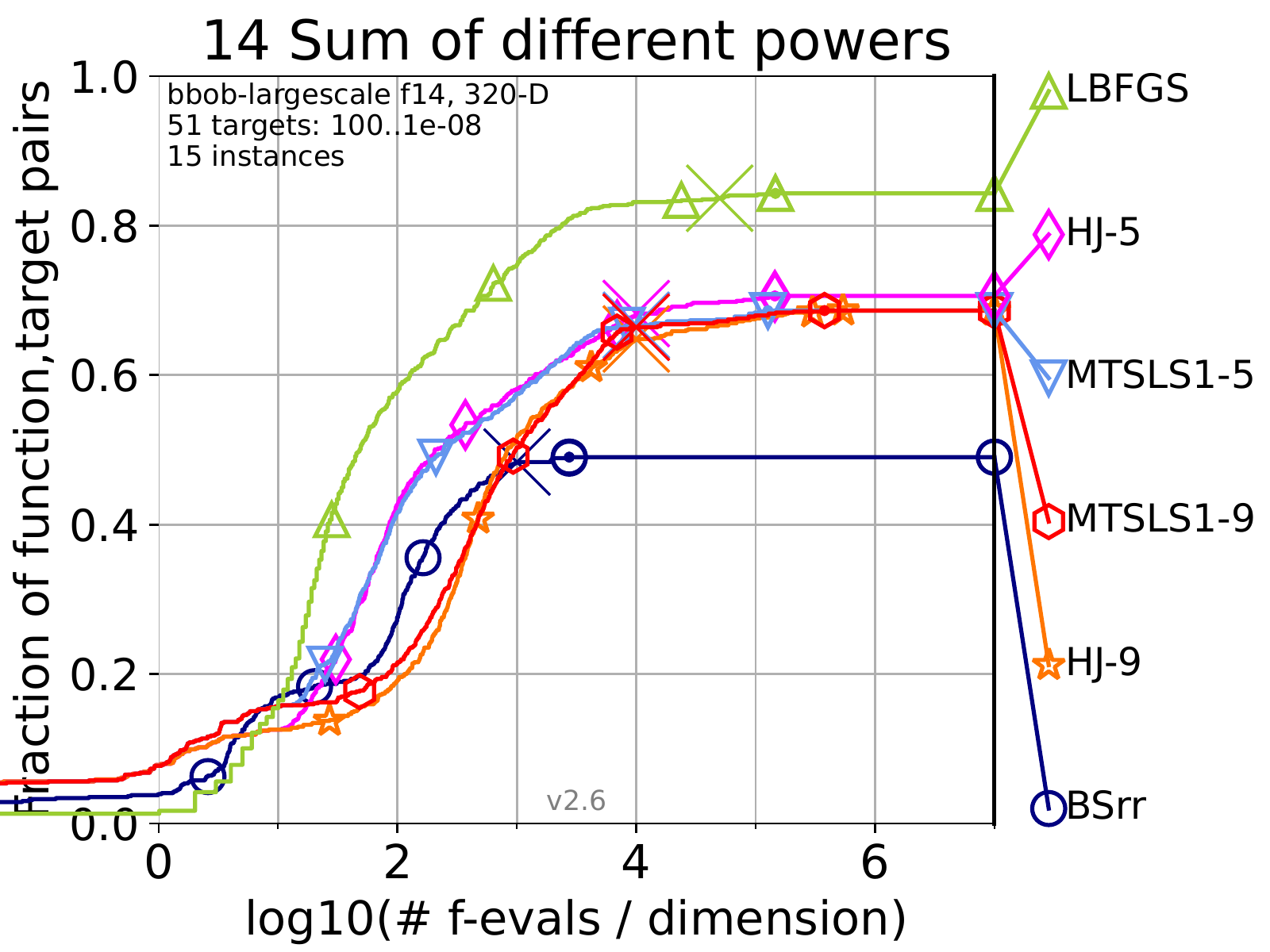}&
\includegraphics[width=\widthvar\textwidth]{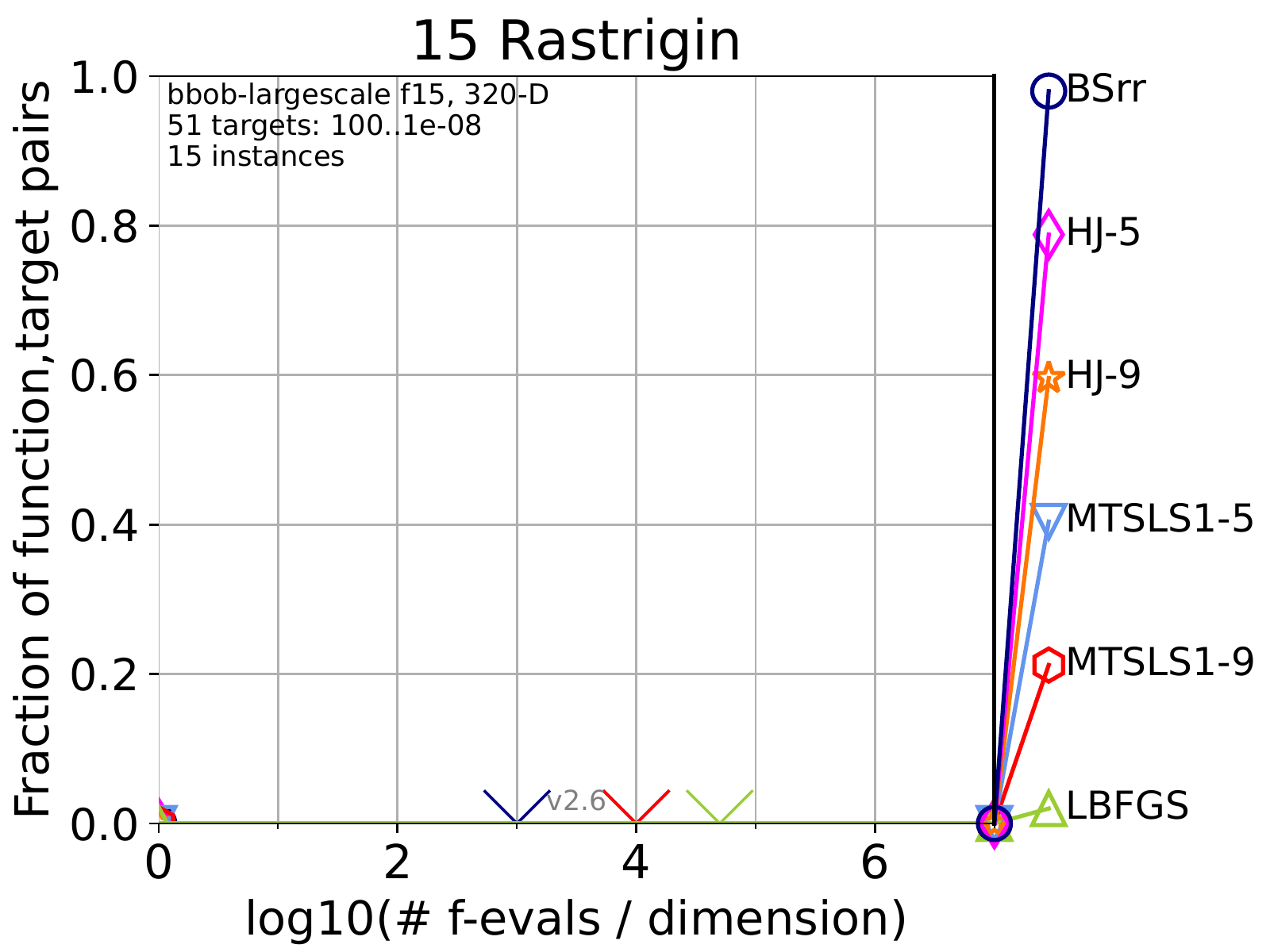}&
\includegraphics[width=\widthvar\textwidth]{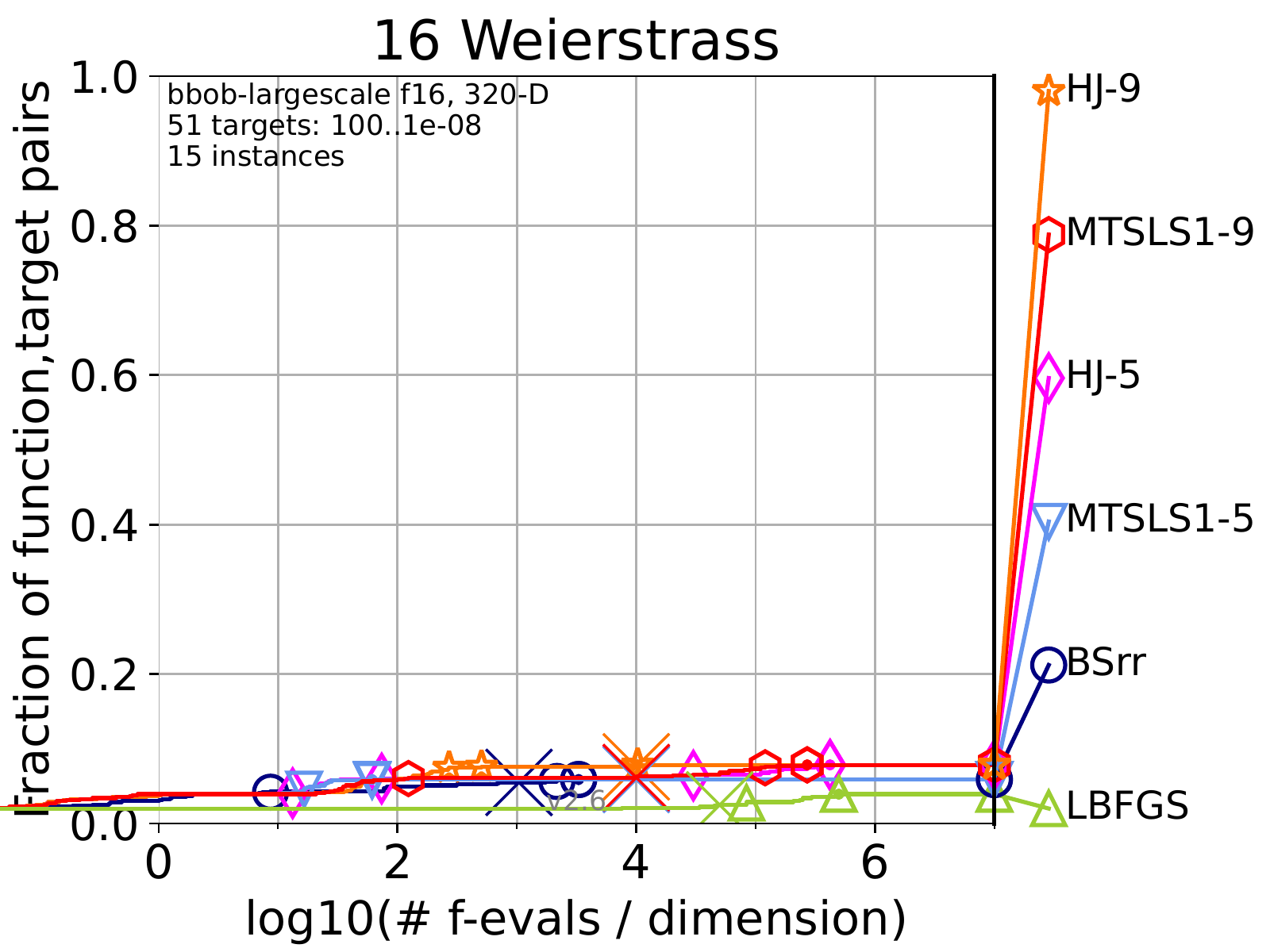}\\
\includegraphics[width=\widthvar\textwidth]{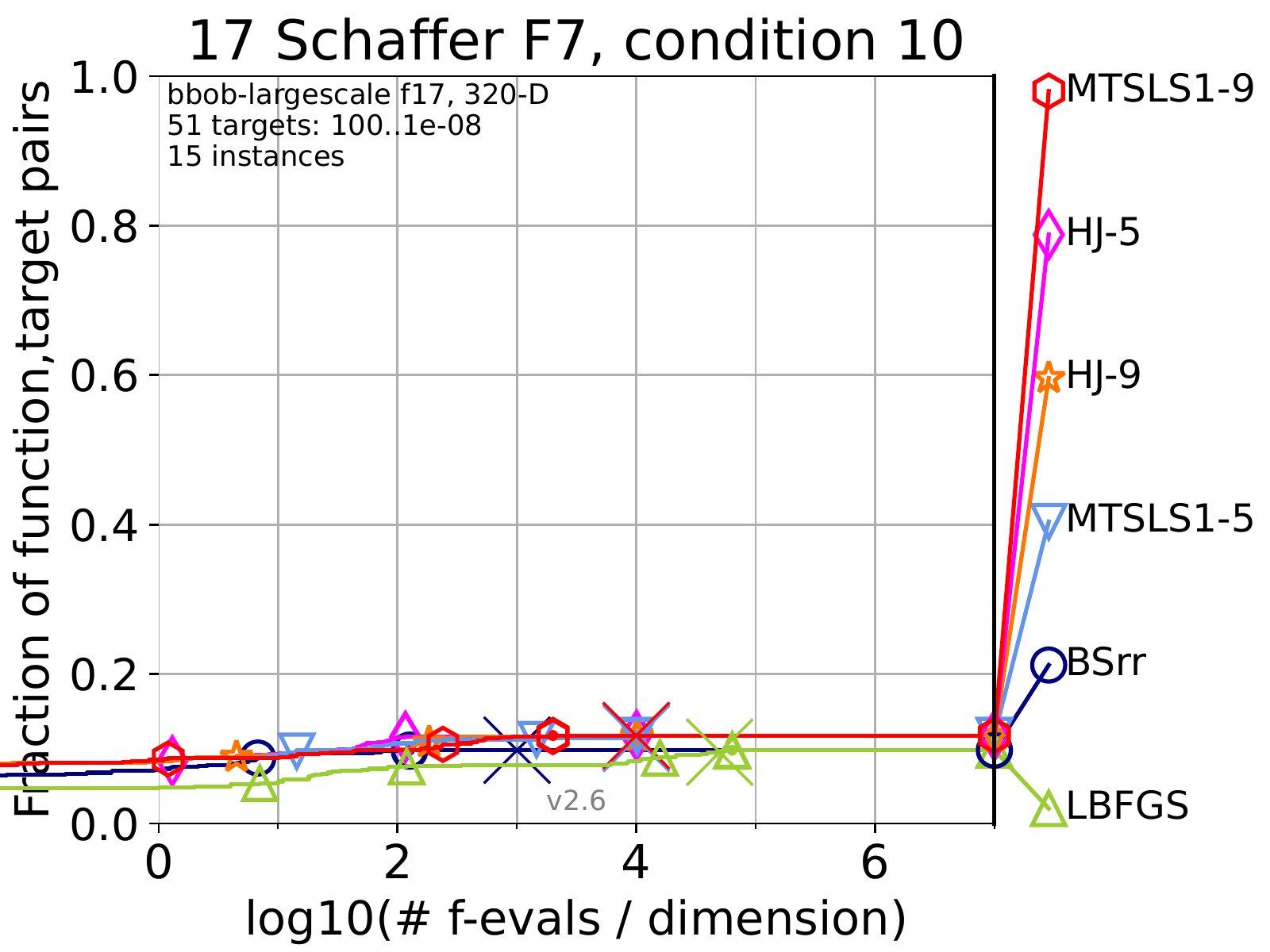}&
\includegraphics[width=\widthvar\textwidth]{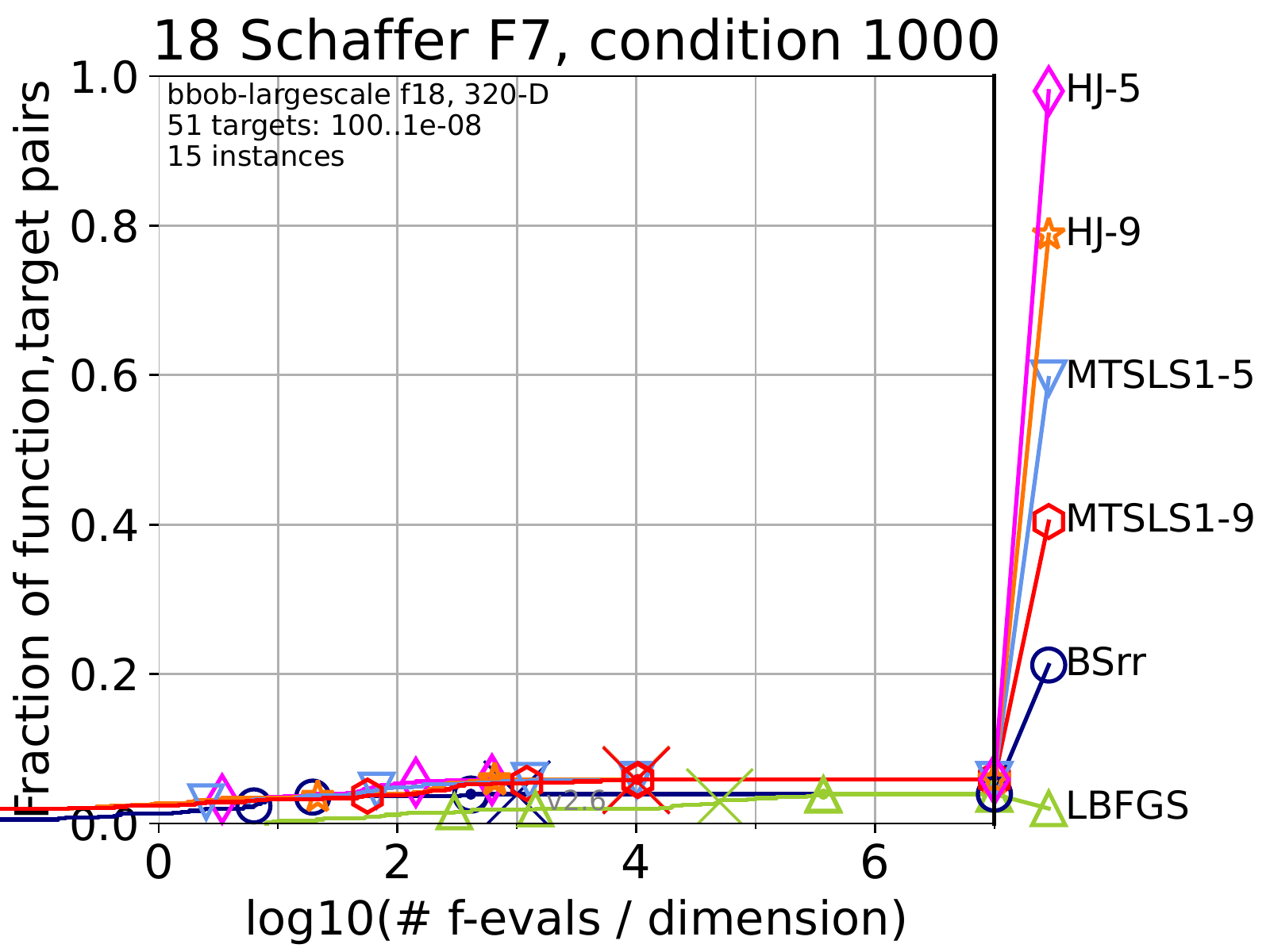}&
\includegraphics[width=\widthvar\textwidth]{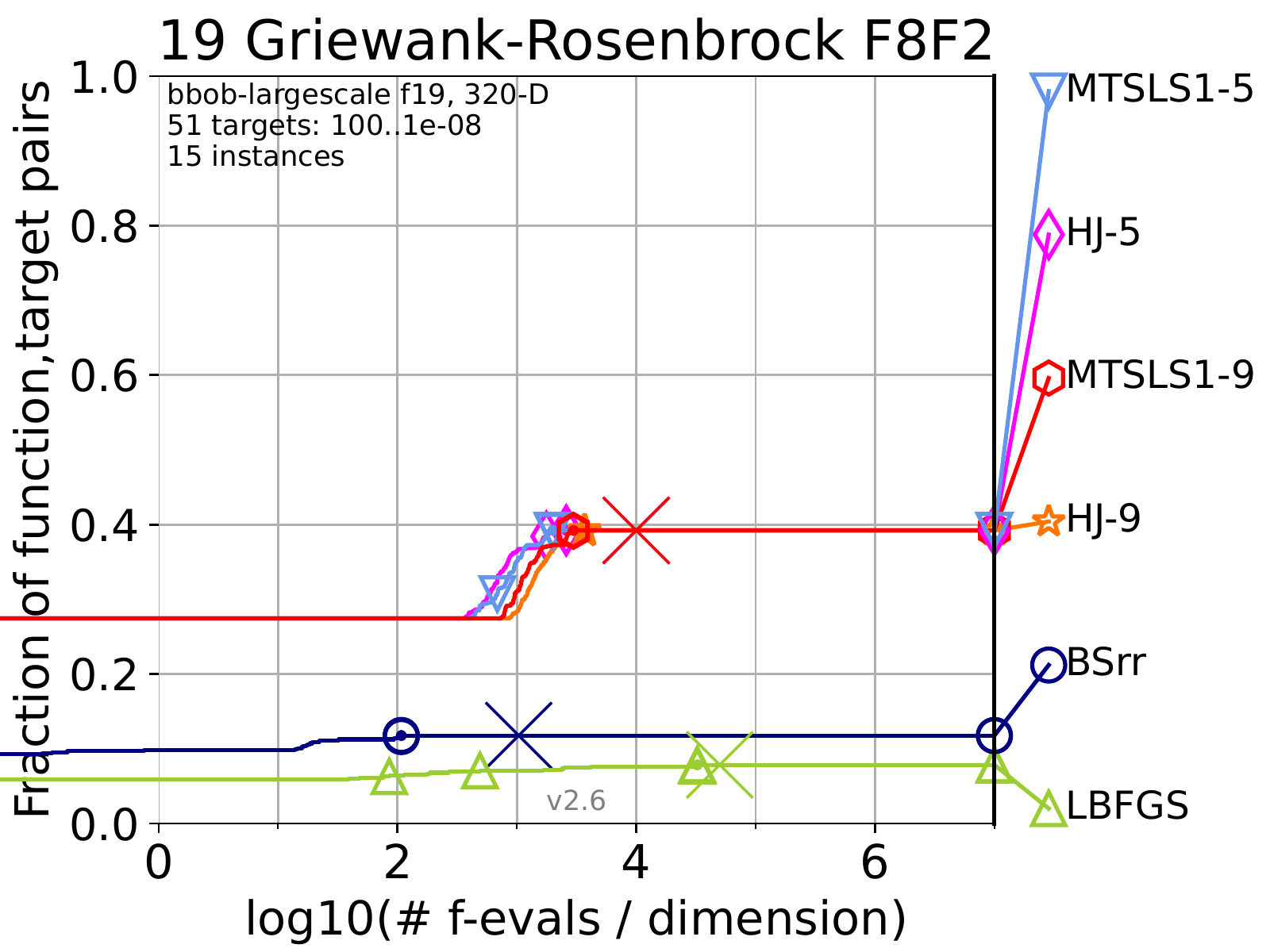}&
\includegraphics[width=\widthvar\textwidth]{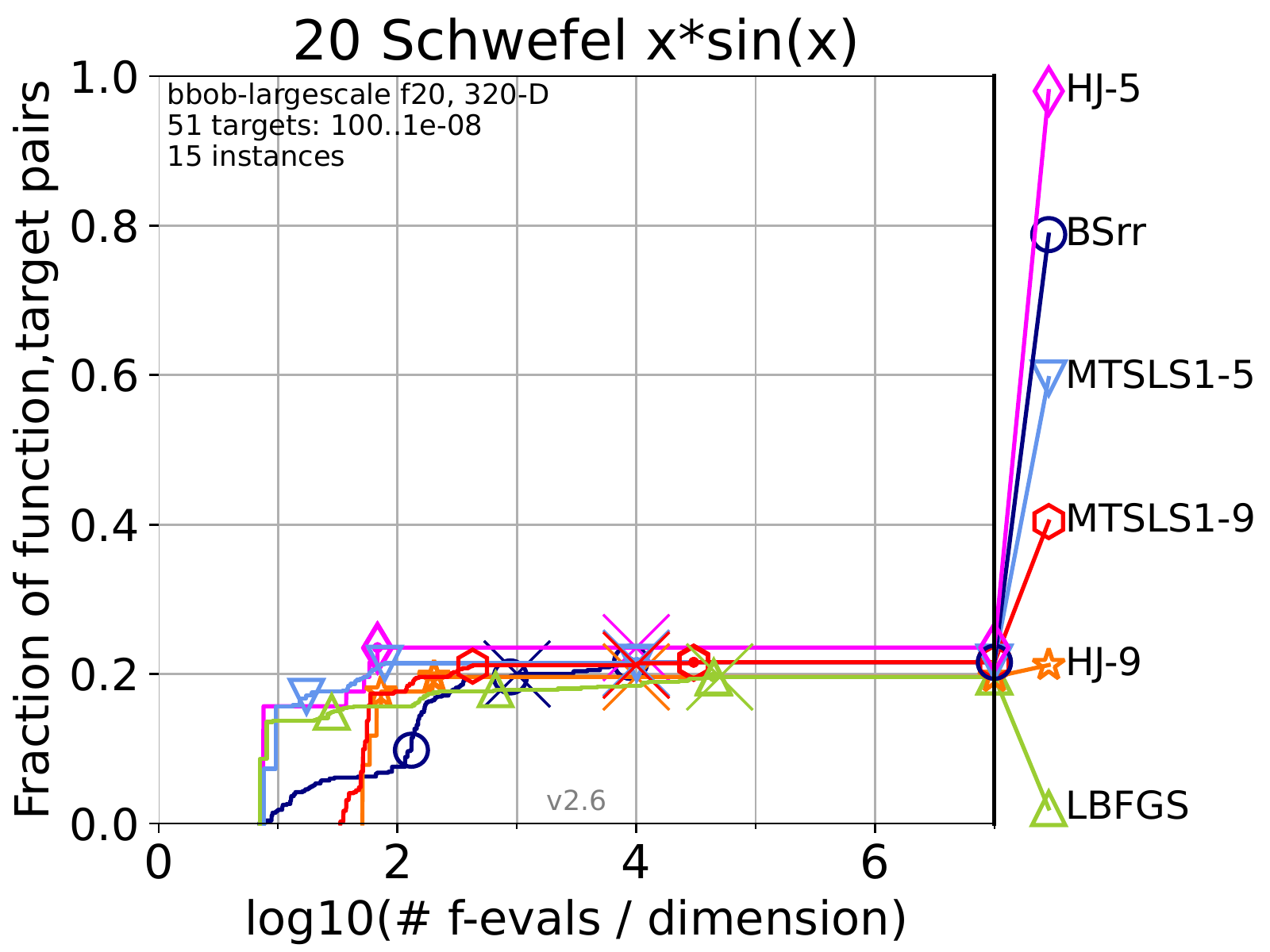}\\
\includegraphics[width=\widthvar\textwidth]{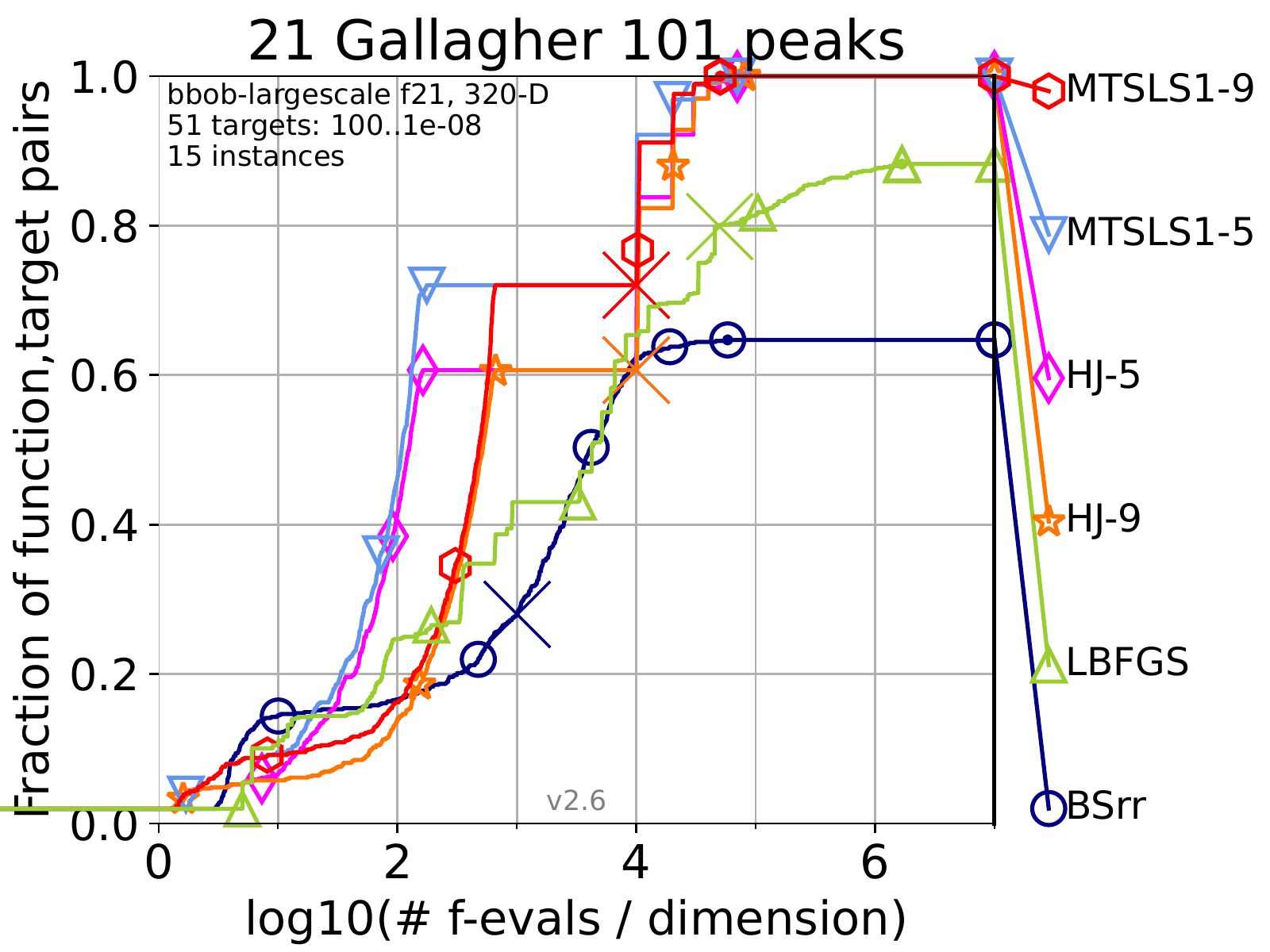}&
\includegraphics[width=\widthvar\textwidth]{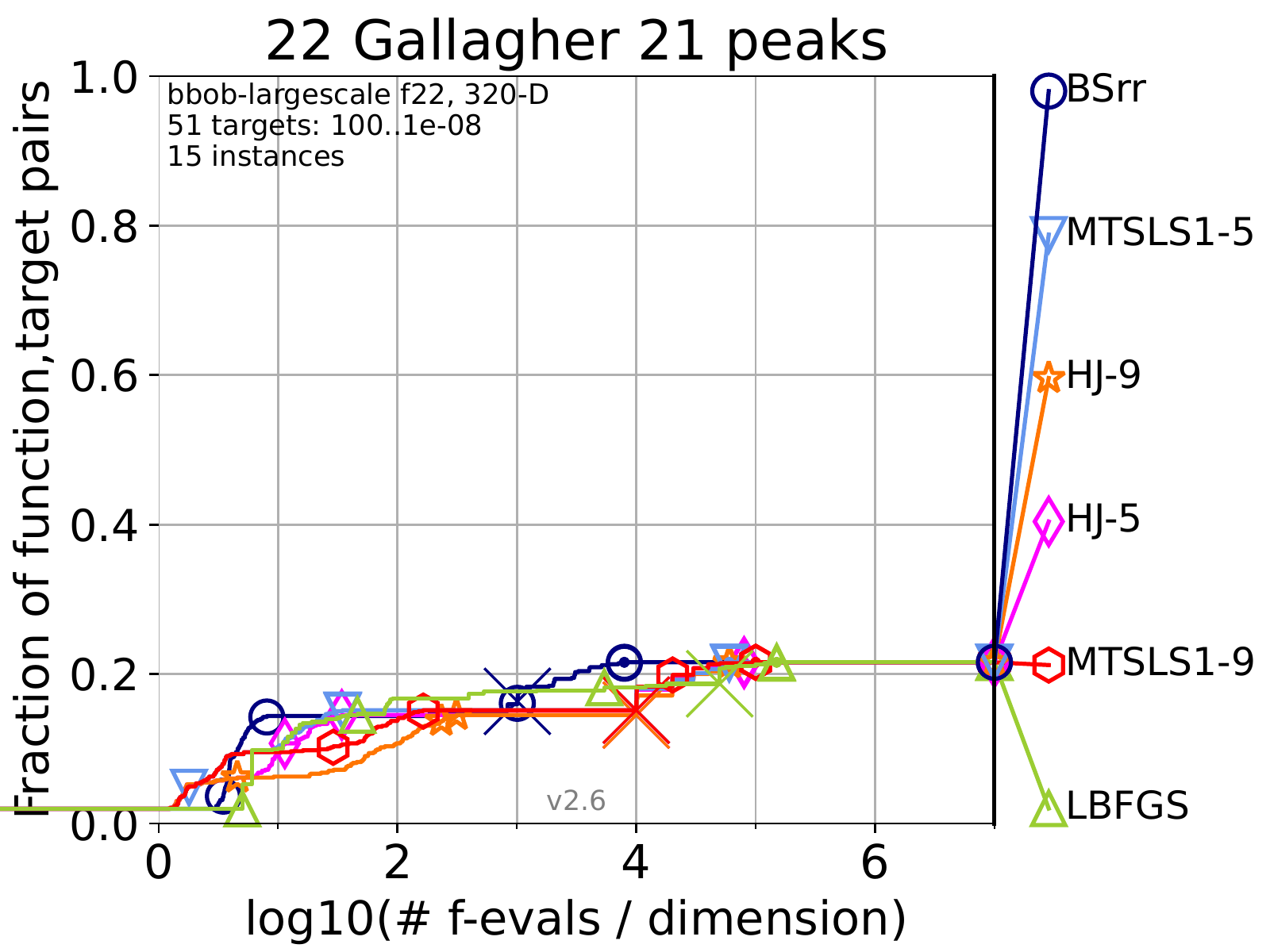}&
\includegraphics[width=\widthvar\textwidth]{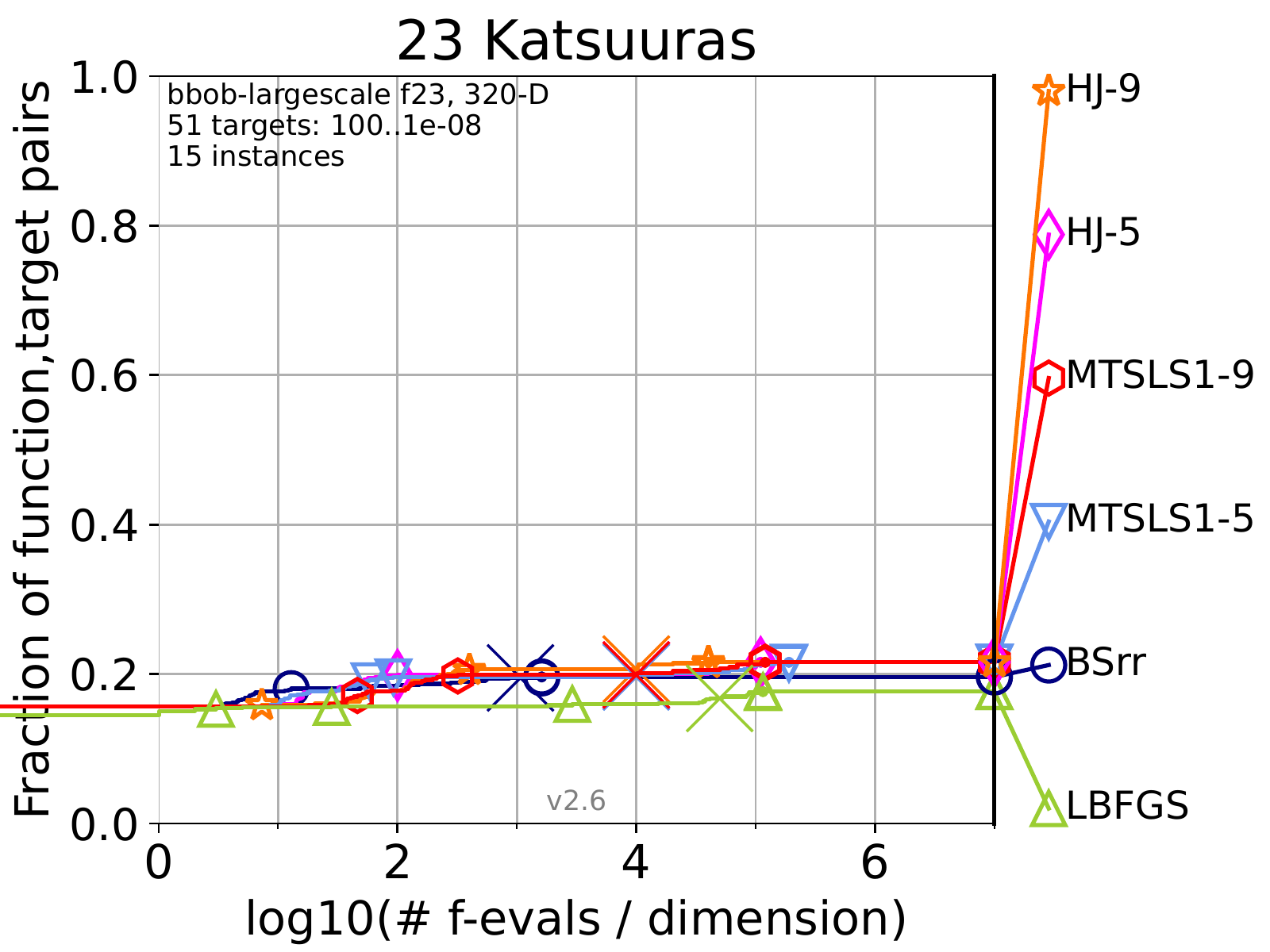}&
\includegraphics[width=\widthvar\textwidth]{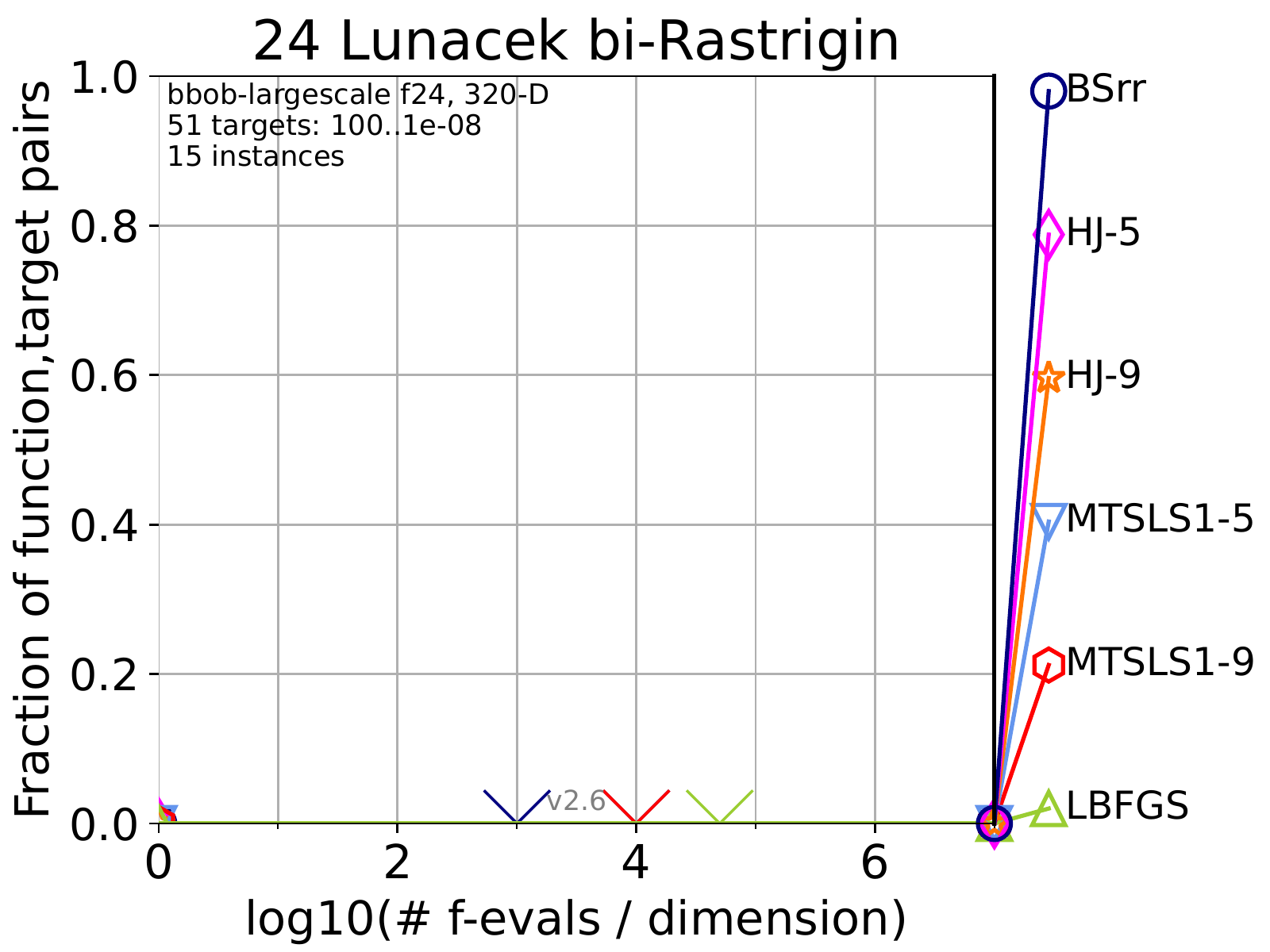}
\end{tabular}
 \caption{\label{fig:ECDFsingleOne}
	\bbobecdfcaptionsinglefunctionssingledim{320}
}
\end{figure*}

Results from experiments according to \cite{hansen2016exp} and \cite{hansen2016perfass} on the
benchmark functions given in \cite{varelas2020benchmarking} are
presented in
\ifthenelse{\equal{\numofalgs}{1}}{
Figures~\ref{fig:ERTgraphs}, \ref{fig:ECDFs}, and \ref{fig:ECDFsingleOne} and Tables~\ref{tab:ERTs80} and \ref{tab:ERTs320}.
}{\ifthenelse{\equal{\numofalgs}{2}}{
Figures~\ref{fig:scaling}, \ref{fig:scatterplots}, \ref{fig:ECDFs80D}, \ref{fig:ECDFs320D} and \ref{fig:ECDFsingleOne}, and Tables~\ref{tab:ERTs80} and \ref{tab:ERTs320}.
}{\ifthenelse{\(\numofalgs > 2\)}{
Figures~\ref{fig:scaling}, \ref{fig:ECDFs80D}, \ref{fig:ECDFs320D}, \ref{fig:ECDFsingleOne80D}, and \ref{fig:ECDFsingleOne} and Tables~\ref{tab:ERTs80} and \ref{tab:ERTs320}.
}{}}}
The experiments were performed with COCO \cite{hansen2020cocoplat}, version
2.5, the plots were produced with version 2.6.

The \textbf{expected runtime (ERT)} 
depends on a given target function value, $\ftarget=\fopt+\Df$, and is
computed over all relevant trials as the number of function
evaluations executed during each trial while the best function value
did not reach \ftarget, summed over all trials and divided by the
number of trials that actually reached \ftarget\
\cite{hansen2012exp,price1997dev}. 
\textbf{Statistical significance} is tested with the rank-sum test for a given
target $\Delta\ftarget$ using, for each trial,
either the number of needed function evaluations to reach
$\Delta\ftarget$ (inverted and multiplied by $-1$), or, if the target
was not reached, the best $\Df$-value achieved, measured only up to
the smallest number of overall function evaluations for any
unsuccessful trial under consideration.




\begin{table*}\tiny
{\normalsize \color{red}
\ifthenelse{\isundefined{\algorithmG}}{}{more than 6 algorithms: please split the tables below by hand until it fits to the page limits}
}
\mbox{\begin{minipage}[t]{0.495\textwidth}
\centering
\pptablesheader
\input{\bbobdatapath\algsfolder pptables_f001_80D} 
\input{\bbobdatapath\algsfolder pptables_f002_80D}
\input{\bbobdatapath\algsfolder pptables_f003_80D}
\input{\bbobdatapath\algsfolder pptables_f004_80D}
\input{\bbobdatapath\algsfolder pptables_f005_80D}
\input{\bbobdatapath\algsfolder pptables_f006_80D}
\input{\bbobdatapath\algsfolder pptables_f007_80D}
\input{\bbobdatapath\algsfolder pptables_f008_80D}
\input{\bbobdatapath\algsfolder pptables_f009_80D}
\input{\bbobdatapath\algsfolder pptables_f010_80D}
\input{\bbobdatapath\algsfolder pptables_f011_80D}
\input{\bbobdatapath\algsfolder pptables_f012_80D}
\end{tabularx}
\end{minipage}
\hspace{0.002\textwidth}
\begin{minipage}[t]{0.499\textwidth}\tiny
\centering
\pptablesheader
\input{\bbobdatapath\algsfolder pptables_f013_80D}
\input{\bbobdatapath\algsfolder pptables_f014_80D}
\input{\bbobdatapath\algsfolder pptables_f015_80D}
\input{\bbobdatapath\algsfolder pptables_f016_80D}
\input{\bbobdatapath\algsfolder pptables_f017_80D}
\input{\bbobdatapath\algsfolder pptables_f018_80D}
\input{\bbobdatapath\algsfolder pptables_f019_80D}
\input{\bbobdatapath\algsfolder pptables_f020_80D}
\input{\bbobdatapath\algsfolder pptables_f021_80D}
\input{\bbobdatapath\algsfolder pptables_f022_80D}
\input{\bbobdatapath\algsfolder pptables_f023_80D}
\input{\bbobdatapath\algsfolder pptables_f024_80D}
\end{tabularx}
\end{minipage}}

\caption{\label{tab:ERTs80}
\bbobpptablesmanylegend{dimension $80$} \cocoversion
}
\end{table*}

 

\begin{table*}\tiny
{\normalsize \color{red}
\ifthenelse{\isundefined{\algorithmG}}{}{more than 6 algorithms: please split the tables below by hand until it fits to the page limits}
}
\mbox{\begin{minipage}[t]{0.495\textwidth}
\centering
\pptablesheader
\input{\bbobdatapath\algsfolder pptables_f001_320D} 
\input{\bbobdatapath\algsfolder pptables_f002_320D}
\input{\bbobdatapath\algsfolder pptables_f003_320D}
\input{\bbobdatapath\algsfolder pptables_f004_320D}
\input{\bbobdatapath\algsfolder pptables_f005_320D}
\input{\bbobdatapath\algsfolder pptables_f006_320D}
\input{\bbobdatapath\algsfolder pptables_f007_320D}
\input{\bbobdatapath\algsfolder pptables_f008_320D}
\input{\bbobdatapath\algsfolder pptables_f009_320D}
\input{\bbobdatapath\algsfolder pptables_f010_320D}
\input{\bbobdatapath\algsfolder pptables_f011_320D}
\input{\bbobdatapath\algsfolder pptables_f012_320D}
\end{tabularx}
\end{minipage}
\hspace{0.002\textwidth}
\begin{minipage}[t]{0.499\textwidth}\tiny
\centering
\pptablesheader
\input{\bbobdatapath\algsfolder pptables_f013_320D}
\input{\bbobdatapath\algsfolder pptables_f014_320D}
\input{\bbobdatapath\algsfolder pptables_f015_320D}
\input{\bbobdatapath\algsfolder pptables_f016_320D}
\input{\bbobdatapath\algsfolder pptables_f017_320D}
\input{\bbobdatapath\algsfolder pptables_f018_320D}
\input{\bbobdatapath\algsfolder pptables_f019_320D}
\input{\bbobdatapath\algsfolder pptables_f020_320D}
\input{\bbobdatapath\algsfolder pptables_f021_320D}
\input{\bbobdatapath\algsfolder pptables_f022_320D}
\input{\bbobdatapath\algsfolder pptables_f023_320D}
\input{\bbobdatapath\algsfolder pptables_f024_320D}
\end{tabularx}
\end{minipage}}

\caption{\label{tab:ERTs320}
\bbobpptablesmanylegend{dimension $320$} \cocoversion
}
\end{table*}

\noindent \textbf{Some observations}
For the sake of reference, we show the results of L-BFGS \cite{LiuN89} reported in \cite{bbob2019:Varelas:2019:BLS:3319619.3326893}.
The comparison results show that the five optimizers perform significantly worse than L-BFGS on ill-conditioned nonseparable functions ($f_6, \dots, f_{14}$), except for $f_7$.
Our results and the results in \cite{bbob2019:Varelas:2019:BLS:3319619.3326893} indicate that any optimizer performs poorly on a high-dimensional $f_7$ with plateaus.
However, as shown in the results on $f_6$ for 80 and 320 dimensions in Tables \ref{tab:ERTs80} and \ref{tab:ERTs320}, the ERT values of HJ-9 are just approximately 4.9 and 2.2 times larger than those of L-BFGS, respectively.
These results indicate that the exploratory move in HJ is effective on $f_6$.
As seen from Figures \ref{fig:ECDFsingleOne80D}, and \ref{fig:ECDFsingleOne}, the position of the initial search point significantly influences the performance of optimizers on $f_{19}$.
This unexpected observation may relate to a known issue in $f_{19}$ (\url{https://github.com/numbbo/coco/issues/1851}).
Here, for HJ and MTS-LS1, we initialized the first search point $\vector{x}$ to the center of the search space $[-5, 5]^D$, i.e., $\vector{x} = (0, ..., 0)^{\top}$.
In contrast, BSrr and L-BFGS select the first search point uniformly at random in $[-1, 3]^D$ and $[-4, 4]^D$, respectively.
Tables \ref{tab:ERTs80} and \ref{tab:ERTs320} show that HJ-5 and HJ-9 reach the smallest target value $f_{\mathrm{opt}} + 10^{-8}$ in $f_5$ for 80 and 320 dimensions within approximately 122 and 481 function evaluations, respectively.
We believe that the excellent performance of HJ-5 and HJ-9 on $f_5$ is due to the replacement-based bound handling operation in our study.
Although the optimal solution in $f_5$ is on the bounds, the replacement operation moves infeasible solutions to the bounds.




\noindent \textbf{Answers to RQ1}:
Figure \ref{fig:scaling} shows that BSrr performs the best on $f_1$ with $D=640$, $f_2$ with $D \leq 160$, $f_3$ with any $D$, and $f_4$.
Especially, the performance of BSrr on $f_3$ and $f_4$ is notable.
Although BSrr cannot reach the smallest target value $f_{\mathrm{opt}} + 10^{-8}$ on $f_4$ for 320 and 640 dimensions, BSrr reaches other target values, e.g.,  $f_{\mathrm{opt}} + 10^{-6}$.
The performance of BSrr unexpectedly deteriorates on $f_2$ as the dimension increases.
This may be because we set the maximum number of function evaluations in BSrr to $10^3 \times D$.
The round-robin dimension selection strategy may also not be suitable for large dimensions.
Benchmarking of other variants of BSrr \cite{PosikB15} (e.g., BSqi) with a large budget of function evaluations is a future research topic.

\noindent \textbf{Answers to RQ2} 
As shown in Figures \ref{fig:ECDFsingleOne80D} and \ref{fig:ECDFsingleOne}, all 15 runs of MTS-LS1-9 on $f_3$ are successful while those of MTS-LS1-5 are unsuccessful.
These observations suggest that the default value of $c$ is not suitable for the BBOB functions.
However, as seen from the ERT values for 320 dimensions in Table \ref{tab:ERTs320}, MTS-LS1-9 performs approximately 6.1 and 6.3 times worse than MTS-LS1-5 on $f_1$ and $f_2$, respectively.
Interestingly, MTS-LS1 with any $c$ performs poorly on $f_4$.
Since MTS-LS1 uses (almost) the symmetric operation, MTS-LS1 can perform poorly on multimodal functions with a highly asymmetric landscape structure like $f_4$.

\noindent \textbf{Answers to RQ3}
Due to the exploratory move, HJ performs better than MTS-LS1 on the unimodal separable $f_1$ and $f_2$.
The unexpected high performance of HJ on $f_6$ may also be due to the exploratory move.
In contrast, HJ performs significantly worse than MTS-LS1 on $f_3$.
The reinitialization strategy for $\sigma$ in MTS-LS1 may make the difference.
We believe that the performance of HJ on $f_3$ can be improved by using the reinitialization strategy in MTS-LS1.

%

\section*{Acknowledgment}

This work was supported by Leading Initiative for Excellent Young Researchers, MEXT, Japan.


\clearpage

\bibliographystyle{ACM-Reference-Format}
\bibliography{reference}

\end{document}